\documentclass[sigconf, nonacm]{acmart}

\pdfoutput=1

\usepackage{adjustbox}
\usepackage[a-1b]{pdfx} 
\usepackage{caption}
\usepackage{multirow}
\usepackage{multicol}
\usepackage{enumitem}
\usepackage{rotating}
\usepackage{siunitx}
\usepackage{extra_commands}
\usepackage{algorithm}
\usepackage{colortbl}
\usepackage{listings}
\usepackage{xcolor}

\definecolor{codegreen}{rgb}{0,0.6,0}
\definecolor{codegray}{rgb}{0.5,0.5,0.5}
\definecolor{codepurple}{rgb}{0.58,0,0.82}
\definecolor{backcolour}{rgb}{0.95,0.95,0.92}

\lstdefinestyle{mystyle}{
    backgroundcolor=\color{backcolour},   
    commentstyle=\color{codegreen},
    keywordstyle=\color{magenta},
    numberstyle=\tiny\color{codegray},
    stringstyle=\color{codepurple},
    basicstyle=\ttfamily\footnotesize,
    breakatwhitespace=false,         
    breaklines=true,                 
    captionpos=b,                    
    keepspaces=true,                 
    numbers=left,                    
    numbersep=5pt,                  
    showspaces=false,                
    showstringspaces=false,
    showtabs=false,                  
    tabsize=2
}

\usepackage{subfig}
\begin{document}
\title{Similarity search in the blink of an eye with compressed indices}

\author{Cecilia Aguerrebere}
\email{cecilia.aguerrebere@intel.com}
\affiliation{%
  \institution{Intel Labs}
  \streetaddress{2200 Mission College Blvd}
  \city{Santa Clara}
  \state{California}
  \postcode{95054}
}

\author{Ishwar Singh Bhati}
\email{ishwar.s.bhati@intel.com}
\affiliation{%
  \institution{Intel Labs}
  \streetaddress{2111 NE 25th Ave}
  \city{Hillsboro}
  \state{Oregon}
  \postcode{97124}
}

\author{Mark Hildebrand}
\email{mark.hildebrand@intel.com}
\affiliation{%
  \institution{Intel Labs}
  \streetaddress{2111 NE 25th Ave}
  \city{Hillsboro}
  \state{Oregon}
  \postcode{97124}
}
\author{Mariano Tepper}
\email{mariano.tepper@intel.com}
\affiliation{%
  \institution{Intel Labs}
  \streetaddress{2111 NE 25th Ave}
  \city{Hillsboro}
  \state{Oregon}
  \postcode{97124}
}

\author{Theodore Willke}
\email{ted.willke@intel.com}
\affiliation{%
  \institution{Intel Labs}
  \streetaddress{2111 NE 25th Ave}
  \city{Hillsboro}
  \state{Oregon}
  \postcode{97124}
}

\thanks{Authors are listed in alphabetical order and have contributed equally to this work.}

\begin{abstract}
Nowadays, data is represented by vectors. Retrieving those vectors, among millions and billions, that are similar to a given query is a ubiquitous problem, known as similarity search, of relevance for a wide range of applications. 
Graph-based indices are currently the best performing techniques for billion-scale similarity search. However, their random-access memory pattern presents challenges to realize their full potential.
In this work, we present new techniques and systems for creating faster and smaller graph-based indices. To this end, we introduce a novel vector compression method, Locally-adaptive Vector Quantization (LVQ), that uses per-vector scaling and scalar quantization to improve search performance with fast similarity computations and a reduced effective bandwidth, while decreasing memory footprint and barely impacting accuracy. LVQ, when combined with a new high-performance computing system for graph-based similarity search, establishes the new state of the art in terms of performance and memory footprint. For billions of vectors, LVQ outcompetes the second-best alternatives:
(1) in the low-memory regime, by up to 20.7x in throughput with up to a 3x memory footprint reduction, and
(2) in the high-throughput regime by 5.8x with 1.4x less memory.
\end{abstract}

\maketitle

\section{Introduction}

In the deep learning era, high-dimensional vectors have become the quintessential data representation for unstructured data, e.g., for images, audio, video, text, genomics, and computer code \cite[e.g.,][]{devlin2018bert,radford2021learning,shvetsova2022everything,ji2021dnabert,li2022competition}. The representations are generated so that semantically related vectors are close to each other according to a chosen similarity function. A common procedure is to search over these vectors for the nearest neighbors to a given query vector. This enables a wide range of applications, such as image generation~\cite{blattmann2022retrieval}, NLP~\cite{borgeaud2022improving}, question answering~\cite{karpukhin2020dense}, recommender systems~\cite{lian2020lightrec}, and ad matching~\cite{grbovic2016scalable}. 
 Datasets with billions of vectors, each with hundreds of dimensions, are increasingly common ~\cite{simhadri2022results}. Because of the scale and the curse of dimensionality, exact nearest neighbor search is impractical and the literature is focused on approximate methods.  Graph-based approximate nearest neighbor methods ~\cite{arya1993approximate,malkov2018efficient,jayaram2019diskann} have been empirically found to offer a better latency-accuracy trade-off than other types of algorithms~\cite{wang_graph_survey_2021}.

\begin{figure}
  \centering
  \includegraphics[width=0.95\linewidth]{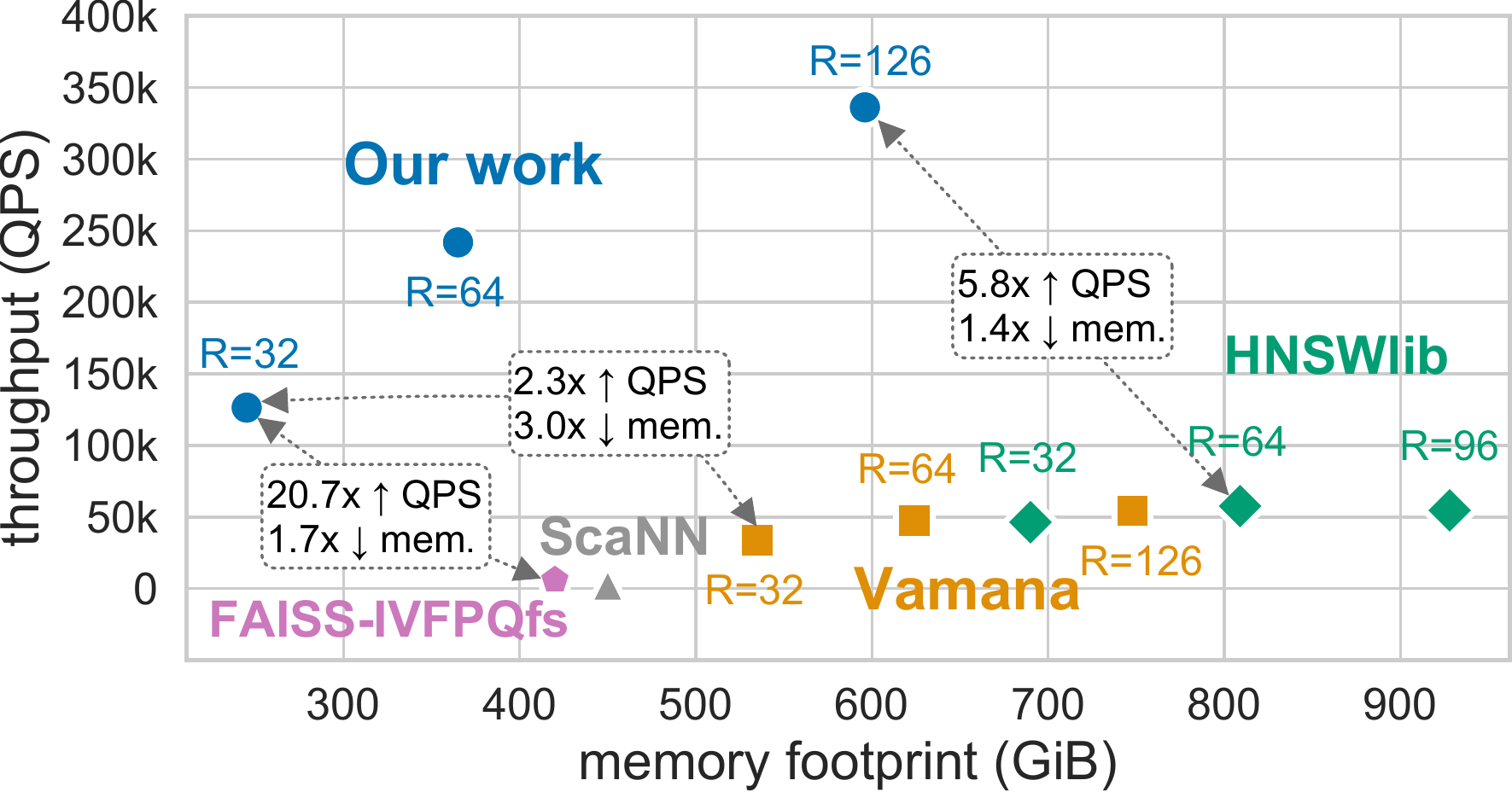}
   \caption{Our contributions enable high-throughput and high-accuracy similarity search with a small memory footprint (results with a 10-recall$@$10 of 0.9 for deep-96-1B). For graph methods, the memory footprint is a function of the graph out-degree $\maxOutDeg{}$ and the vectors' footprint.
   Our low-memory configuration (LVQ-8 and $\maxOutDeg{}=32$) outperforms the current leaders, Vamana~\cite{jayaram2019diskann}, HNSWlib~\cite{malkov2018efficient}, FAISS-IVFPQfs~\cite{johnson2019billion}, and ScaNN~\cite{guo2020accelerating}, by 2.3x, 2.2x, 20.7x, and 43.6x with 3.0x, 3.3x, 1.7x, and 1.8x less memory, respectively.
   Our highest-throughput configuration (LVQ-8 and $\maxOutDeg{}=126$) outperforms the second-highest by 5.8x while using 1.4x less memory. 
   }
   \label{fig:search_with_reduced_memfootprint_1B}
\end{figure}

Despite requiring fewer memory accesses per query, graph-based search algorithms continue to offer \textbf{limited throughput} and a \textbf{substantial memory footprint} at very large database sizes making single-machine deployment challenging.  Many works have focused on reducing the number of distance computations per query further by changing the parameters of the graph to lower the number of points visited per query.  Although this lowers latency and increases query throughput, it does so at the expense of recall and with little reduction in memory consumption.  Since applications typically demand a hard and fast lower bound on recall, there are limits to this approach.

The random memory access patterns that come with graph algorithms present further challenges to the system's throughput. The \textbf{inability to effectively prefetch} vectors with a hardware prefetcher means that the latency of accessing random vectors cannot be easily hidden and may quickly become a throughput bottleneck.  Furthermore, vectors are \textbf{difficult to cache} due to the size of the index. Although most of the literature on large-scale similarity search puts more emphasis on the computational intensity of this workload, the simplicity of the distance computation kernel and the aforementioned fetching issues ultimately make the workload memory-bandwidth limited.  However, many billion-scale similarity search systems lack the high-performance computing block-and-tackling necessary to wring out enough distance computation performance to put pressure on the memory subsystem.

Gains may be made by performing distance computations on compressed vectors, thereby lowering both computation and memory footprint.  However, compression introduces new challenges, including \textbf{lowering of recall}. Product Quantization and other lossy compression methods are often used to reduce the memory footprint~\cite{jayaram2019diskann} but incur more \textbf{expensive similarity computations} and require the auxiliary \textbf{storage of uncompressed vectors} anyway to boost recall during a final re-ranking step.  Other methods introduce too much distortion in the distances (e.g., dimensionality reduction and standard scalar quantization), leading to unacceptable accuracy.

In this work, we propose LVQ, a locally-adaptive vector quantizer, that uses a \textbf{simple and efficient compression} method to reduce memory pressure and a built-in two-level quantization remainder system that \textbf{avoids keeping full precision vectors}.  After centering the data, LVQ scales each vector individually (i.e., the local adaptation) and then performs uniform scalar quantization. Its per-vector compression introduces a \textbf{negligible accuracy degradation} thanks to its effective usage of all quantization levels. LVQ reduces the bandwidth by up to \textasciitilde8x compared to a float32-valued vector, greatly accelerating the search. When needed, the second-level quantization remainder is used for a final re-ranking to further boost search recall. Moreover, we show that we can build accurate graphs directly from LVQ-compressed vectors.

We also introduce a new open-source performance library for billion-scale similarity search that removes the barriers limiting the throughput of most graph-based search algorithms and, in our case, allows LVQ to shine. We \textbf{streamline memory accesses}, by flattening the memory layout, avoiding any memory indirections, and using a \textbf{new custom software prefetcher}. We also accelerate \textbf{similarity computations} using AVX instructions, which in the case of LVQ are \textbf{blazingly fast} (up to 2.12x faster than with float16-valued vectors). In summary, we present the following contributions:
\begin{itemize}[topsep=0.ex,leftmargin=2ex]
    \item \textbf{Novel Compression Algorithm}.  We present Locally-adaptive Vector Quantization (LVQ), a technique that strikes a balance between effective bandwidth reduction and introducing a minimal decoding overhead for similarity computations (\cref{sec:lvq}).
    \item \textbf{Fast Implementation}. The combination of LVQ and our optimized graph search (presented in \cref{sec:systems} with hyperparameter recommendations) establishes the new state-of-the-art for large-scale similarity search in terms of performance and memory footprint. We present a preview in \cref{fig:search_with_reduced_memfootprint_1B}. We backup these claims with an extensive set of experimental results (\cref{sec:results}).
    \item \textbf{Index construction with LVQ}. LVQ enables building graph indices directly from compressed vectors, releasing memory pressure in this time-consuming step while minimally affecting index quality (\cref{sec:lightweight_indices}).
    \item \textbf{Open Source Framework}.  We open-source a similarity search library\footnote{\url{https://github.com/IntelLabs/ScalableVectorSearch}} to allow the research community to experiment with our algorithms and billion-scale search framework.
    \item \textbf{New Dataset and Generator}.  To promote similarity search research in line with modern applications that use deep learning embeddings, we introduce a new dataset with $d=768$ dimensions, produced using large language models~\cite{karpukhin2020dense}. We open source the code\footnote{\url{https://github.com/IntelLabs/DPR-dataset-generator}} to generate this dataset from a standard corpus. In this paper we use an instance with 10 million vectors.
\end{itemize}

\section{Preliminaries}
\label{sec:preliminaries}

The similarity search problem (also known as nearest-neighbor search) is defined as follows. Given a vector database $\X = \{ \vect{x}_i \in \Real^d \}_{i=1,\dots,n}$, containing $n$ vectors with $d$ dimensions each, a similarity function, and a query $\q \in \Real^d$, we seek the $k$ vectors in $\X$ with maximum similarity to $\q$. Given the size of modern databases, guaranteeing an exact retrieval becomes challenging and this definition is relaxed to allow for a certain degree of error: some retrieved elements may not belong to the ground-truth top $k$. This relaxation avoids a full linear scan of the database.

Similarity is determined using a similarity function $\simfun : \Real^d \times \Real^d \mapsto \Real$, where a higher value indicates a higher degree of similarity. This function is symmetric, i.e., $\simfun(\vect{x}, \vect{x}') = \simfun(\vect{x}', \vect{x})$.

\textbf{Metrics.} Search accuracy is measured by $k$-recall$@k$, defined by $| S \cap G_t | / k$, where $S$ are the ids of the $k$ retrieved neighbors and $G_t$ is the ground-truth. Unless otherwise specified, we use $k=10$ in all experiments and 0.9 as the default accuracy value. Search performance is measured by queries per second (QPS). 

\textbf{Scalar quantization.}
Scalar quantization is a classical technique in signal processing mapping floating-point values to integer values within a specified range.
We define the uniform scalar quantization function as
\begin{equation}
    Q(x; B, \ell, u) = \Delta \left\lfloor \frac{x - \ell}{\Delta} + \frac{1}{2} \right\rfloor + \ell ,
    \quad\text{where}\quad
    \Delta = \tfrac{u - \ell}{2^B - 1},
\label{eq:scalar_quant}
\end{equation}
$B$ is the number of bits used for the code, and $u$ and $\ell$ are upper and lower bounds, commonly chosen as the maximum and minimum of the values to quantize.

\subsection{Graph-based similarity search}

Among other similarity search approaches, graph-based methods~\cite{arya1993approximate,malkov2018efficient,jayaram2019diskann} stand out with their high accuracy and performance for high-dimensional data. They are the state of the art at billion-scale~\cite{simhadri2022results}. The key idea is that a fast search algorithm is guaranteed to converge to the nearest neighbor by a best-first traversal of the Delaunay graph. However, building a Delaunay graph is too computationally expensive and approximations are used~\cite{wang_graph_survey_2021}. Many variations exist, e.g., using different edge selection strategies~\cite{fu_fast_2019,iwasaki_optimization_2018} or adding a hierarchy~\cite{malkov2018efficient,ChenW18}. In this work, we use the graph building introduced by Subramanya et al.~\cite{jayaram2019diskann} for its strong search performance, but our results apply to other graphs-based methods.

In the following discussion, let $\Gr=(\V,\E)$ be a directed graph with vertices $\V$ corresponding to elements in a dataset $\X$ and edges $\E$ representing neighbor-relationships between vectors. 
We denote with $\set{N}(\vect{x})$ the set of out-neighbors of $\vect{x}$ in $\Gr$.

\textbf{Graph search.}
Graph search involves retrieving the $k$ nearest vectors to query $\q \in \Real^d$ with respect to the similarity function $\simfun$ by using a modified greedy search over $\Gr$ (see pseudo-code in \cref{alg:search}).
The parameter \searchWin{} provides a knob for trading accuracy and performance as increasing \searchWin{} improves the accuracy of the $k$ nearest neighbors at the cost of lower performance by exploring more of the graph.
Practical implementations of \cref{alg:search} provide optimization opportunities discussed in Section~\ref{sec:systems}.

\begin{algorithm2e}[t]
\small
\DontPrintSemicolon
    \KwIn{graph $\Gr = (\V, \E)$, query $\q$, number of neighbors $k \in \Nat$, priority queue capacity $W \geq k$, initial candidates $\set{S} \subset \set{V}$, similarity function $\simfun$}
    \KwResult{$k$ approximate nearest neighbors to $\q$ in $\Gr$}
    
    $\queue = \set{S}$
    \tcp*[l]{Initialize candidate set $\queue$.}
    
    \tcp{Initially, no nodes are marked as explored.}
    \While{there exists an unexplored node in $\queue$}{
        $\vect{x} = $ closest unexplored node to $\q$ in $\queue$ w.r.t. $\simfun$\;
        Mark $\vect{x}$ as explored\;
        \lFor{$\vect{x}^{\prime} \in \set{N}(\vect{x})$}{
            $\queue \gets \queue \cup \vect{x}^{\prime}$
        }

        \tcp{Limit the size of $\queue$ to at most $W$:}
        $\queue = $ the (at most) $W$ closest nodes to $\q$ in $\queue$ w.r.t $\simfun$\;
    }
    \nllabel{alg:search_postprocessing}
    \Return{$k$ nearest nodes to $\q$ in $\queue$ w.r.t. $\simfun$}
    
    \caption{Greedy graph search.}
    \label{alg:search}
\end{algorithm2e}

\textbf{Graph construction.}
To build the graph we follow the approach by Subramanya et al.~\cite{jayaram2019diskann}.
Starting from an uninitialized graph $\Gr = (V, \emptyset)$ and target maximum degree $R$, we iteratively perform an update routine for each node $\vect{x} \in \V$.
For this, we first run Algorithm~\ref{alg:search} using the node $\vect{x}$ as the query with $\searchWin{} > \maxOutDeg{}$ on the current graph $\Gr$ to obtain $\Cl$: the $k = \searchWin{}$ approximate nearest neighbors to $\vect{x}$.
The pruning algorithm~\cite{jayaram2019diskann} shown in \cref{alg:pruning} is run on $\Cl$ to refine the candidate list.
The refined candidate list $\Cl$ is used to update the outward adjacency list for $\vect{x}$ in $\Gr$.
Finally, we add backward edges $(\vect{x},\vect{x}')$ for all $\vect{x}'$ in $\vect{x}$'s updated neighborhood and prune $\vect{x}'$'s edges using \cref{alg:pruning} to the maximum degree $\maxOutDeg{}$.

\begin{algorithm2e}[t]
\small
\DontPrintSemicolon
    \KwIn{graph $\Gr$, $\vect{x} \in \V$, set $\Cl$ of out-neighbor candidates for $\vect{x}$, relaxation factor $\alpha \in \Real^+$, out-degree bound $\maxOutDeg \in \Nat$, similarity function $\simfun$}
    \KwResult{The new out-neighbors $\set{N}(\vect{x})$ of $\vect{x}$ in $\Gr$ s.~t.~$|\set{N}(\vect{x})| \leq \maxOutDeg{}$.}
    $\Cl \gets (\Cl \cup \set{N}(\vect{x})) \setminus \{ \vect{x} \}$
    \tcp*[l]{Add the current out-neighbors}
    
    $\set{N}(\vect{x}) \gets \emptyset$
    \tcp*[l]{Clear the out-neighbors of $\vect{x}$}
    
    \While{$\Cl \neq \emptyset$}{
        $\displaystyle \vect{x}^* \gets \argmax_{\vect{x}'' \in \Cl} \simfun(\vect{x}, \vect{x}'')$\;
        $\set{N}(\vect{x}) \gets \set{N}(\vect{x}) \cup \{ \vect{x}^* \}$\;
        \lIf{$| \set{N}(\vect{x})| = \maxOutDeg{}$}{
            \texttt{break}     
        }
        \For{$\vect{x}' \in \Cl$}{
            \lIf{$\alpha \cdot \simfun(\vect{x}^*,\vect{x}') \geq \simfun(\vect{x},\vect{x}')$}{
                \nllabel{alg:pruning_rule}
                $\Cl \gets \Cl \setminus \{\vect{x}'\}$
            }
        }
    }
    \caption{Neighborhood graph pruning~\cite{jayaram2019diskann}.}
    \label{alg:pruning}
\end{algorithm2e}

Two passes are done through the dataset~\cite{jayaram2019diskann}: one with the relaxation factor $\alpha=1.0$ and the other with a potentially different $\alpha$.
The optimal values for hyperparameters such as $\alpha$, $\maxOutDeg{}$, and $\searchWin{}$ depend on several factors, such as the dataset manifold, its scale and the accuracy-performance trade-off of choice.
Nevertheless, we find in practice that the same parameter values work very well for different datasets of similar scale (see \cref{sec:results}).

\section{Locally-adaptive Vector Quantization}
\label{sec:lvq}

The IEEE 754 format~\cite{IEEE754} is designed for flexibility, allowing to represent a wide range of very small and very large numbers. However, our empirical analysis of many standard datasets and deep learning embeddings informed us of regularities in the empirical distributions of their component values. In line with modern trends in AI~\cite{gholami2021survey}, we leverage these regularities for quantization. We explored several 8-bit floating point encodings~\cite{viswanathan2013intel}, but found the precision over the small dynamic range of numeric values present in our application to be insufficient for the required search accuracy.

\begin{figure}[t]
  \centering 
  \includegraphics[width=\linewidth]{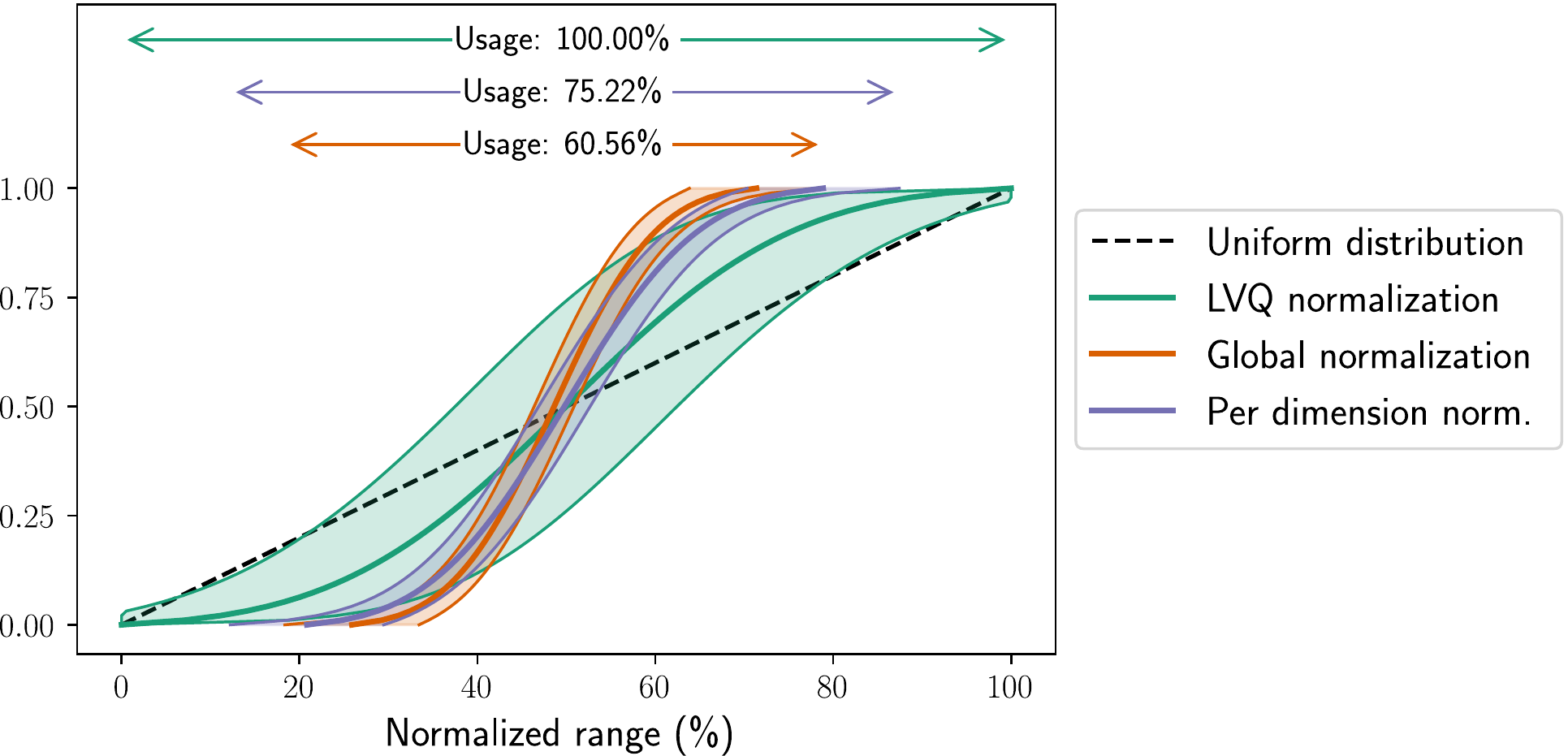}  
  \caption{Empirical distributions of the values in each vector for deep-96-1M (we show the mean across vectors $\pm 2 \sigma$). For 95\% of the vectors, global and per dimension normalization only utilize around 60\% and 75\% of the available range, respectively. LVQ normalization approximates the uniform distribution better, utilizing the whole range and yielding a more faithful encoding.}
  \label{fig:scalar_vs_lvq}
\end{figure}

We further found that scalar quantization, in \cref{eq:scalar_quant}, with global bounds for the whole dataset or with bounds computed individually for each dimension do not make a good use of the available bits, as demonstrated in \cref{fig:scalar_vs_lvq}. In the following, \textit{global quantization} refers to scalar quantization with global normalization.

We thus introduce Locally-adaptive Vector Quantization (LVQ) that fully utilizes the available range (\cref{fig:scalar_vs_lvq}) by changing the slicing direction for computing the quantization bounds. Retaining the simplicity of scalar quantization allows for fast similarity computations while reducing the effective bandwidth.

\begin{definition}
\label{def:lvq}
We define the Locally-adaptive Vector Quantization (LVQ-$B$) of vector $\vect{x} = [ x_1, \dots, x_d ]$ with $B$ bits as 
\begin{equation}
    Q(\vect{x}) = [Q(x_1 - \mu_1; B, \ell, u), \dots, Q(x_d  - \mu_d; B, \ell, u)] ,
\label{eq:quant_def}
\end{equation}
where the scalar quantization function $Q$ is defined in \cref{eq:scalar_quant}, $\vect{\mu} = [\mu_1, \dots, \mu_d]$ is the mean of all vectors in $\mat{X}$ and the constants $u$ and $\ell$ are individually defined for each vector $\vect{x} = [ x_1, \dots, x_d ]$ by
\begin{align}
    u &= \max_{j} x_j - \mu_j,
    &
    \ell &= \min_{j} x_j - \mu_j.
\end{align}
\end{definition}

\begin{figure}[t]
  \centering
  \includegraphics[width=0.48\linewidth]{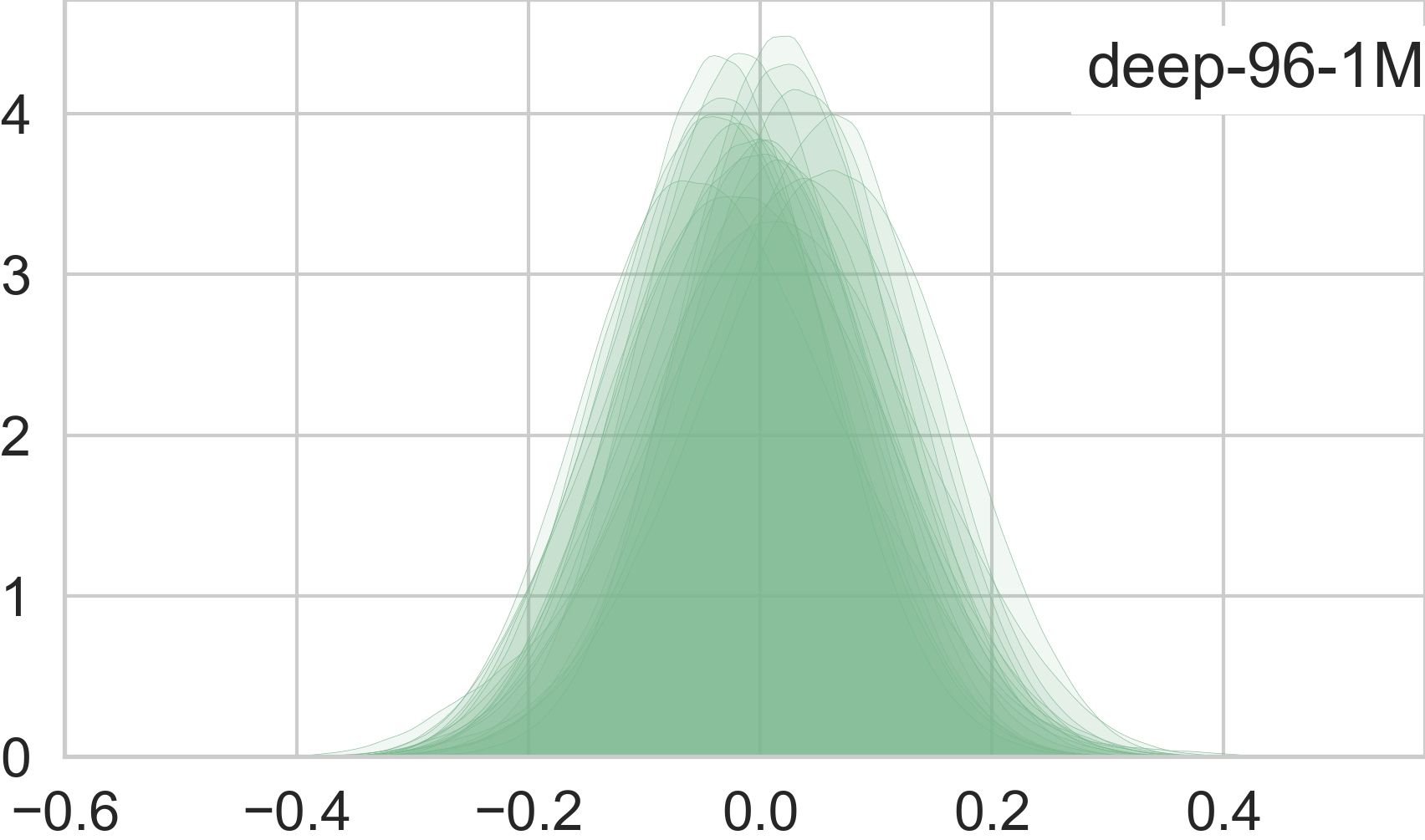}  
  \includegraphics[width=0.49\linewidth]{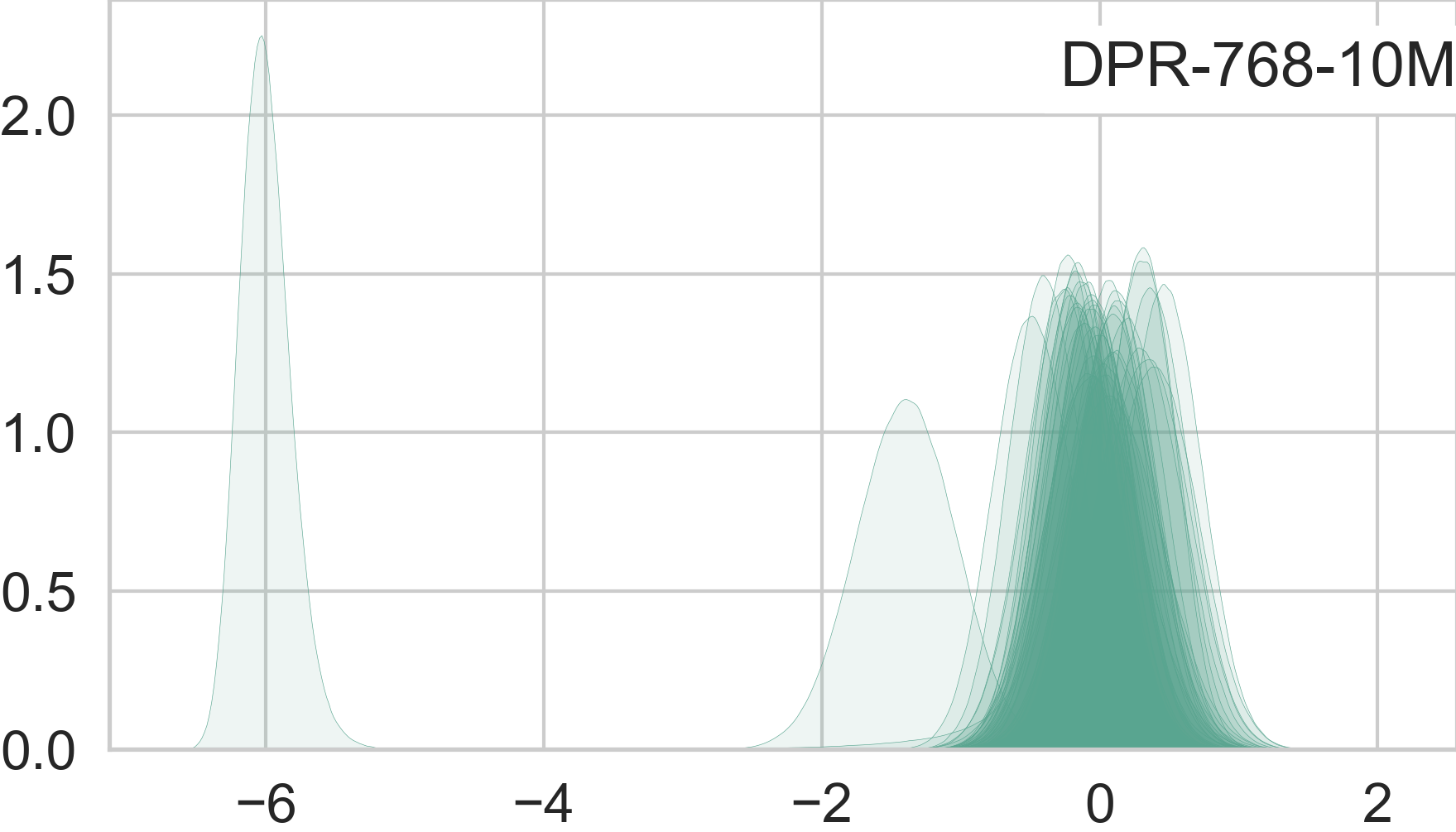}
  \caption{Empirical distributions of vector values in individual dimensions for two prototypical datasets. After de-meaning, the values become highly amenable to quantization, as the distributions will not contain regions within the dynamic range with either very high or very small density. Additional datasets are included in \cref{fig:vector_components_distro_supMat} in the supplementary material \cite{aguerrebere2023similarity}.}
  \label{fig:vector_components_distro}
\end{figure}

LVQ works with mean-centered vectors to homogenize the distributions across dimensions, see \cref{fig:vector_components_distro}.
The quantization bounds $u$ and $\ell$ are computed individually for each vector and, hence, are locally adaptive. This normalization ensures that the dynamic range is used efficiently, see \cref{fig:scalar_vs_lvq}.
Treating all dimensions equally could be problematic in the presence of large variance differences across vector dimensions. Although this scenario is not observed in practice (\cref{fig:vector_components_distro} and \cref{fig:vector_components_distro_supMat} in the supplementary material \cite{aguerrebere2023similarity}), either in standard datasets or in deep learning embeddings, we empirically show the robustness of LVQ in \cref{supp:large_variance_differences} \cite{aguerrebere2023similarity}.

For each $d$-dimensional vector compressed with LVQ-$B$, we need to store the quantized values and the constants $u$ and $\ell$. Moreover, to improve search performance, LVQ-compressed vectors can be padded to a multiple of $p=32$ bytes to be aligned with half cache lines. 
The footprint in bytes of a vector compressed with LVQ-$B$ is
\begin{equation}
    \label{eq:LVQ_footprint}
    \operatorname{footprint}(Q(\vect{x})) = \lceil (d \cdot B + 2B_{const}) / 8 / p \rceil \cdot p, 
\end{equation}
where $B_{const}$ is the number of bits used for $u$ and for $\ell$. Typically, we encode them in float16, in which case $B_{const} = 16$.

The compression ratio $\operatorname{CR}$ for LVQ is given by
\begin{equation}
    \operatorname{CR}(Q(\vect{x})) = d \cdot B_{orig} / (8 \cdot \operatorname{footprint}(Q(\vect{x}))) ,
    \label{eq:compression_ratio}
\end{equation}
where $B_{orig}$ is the number of bits per each dimension of $\vect{x}$. Typically, vectors are encoded in float32, thus $B_{orig} = 32$. For example, when using $B=8$ bits and no padding ($p=0$), the compression ratio for deep-96-1B ($d=96$) is 3.84 and 3.98 for DPR-768-10M ($d=768$).

\subsection{Two-level quantization}

 In graph search, most of the search time is spent (1) performing random memory accesses to retrieve the vectors associated with the out-neighbors of each node and (2) computing the similarity between the query and each vector.
 After optimizing the compute using AVX instructions, search is heavily dominated by the memory access time. This is exacerbated as the number $d$ of dimensions increases ($d$ is in the upper hundreds for deep learning embeddings).
 
To reduce the effective memory bandwidth during search, we compress each vector in two levels, each with a fraction of the available bits.
After using LVQ for the first level, we quantize the residual vector $\vect{r} = \vect{x} - \vect{\mu} - Q(\vect{x})$. The scalar random variable $Z = X - \vect{\mu} - Q(X)$, which models the first-level quantization error, follows a uniform distribution in $[-\Delta/2, \Delta/2)$ (see \cref{eq:scalar_quant}). Thus, we encode each component of $\vect{r}$ using the scalar quantization function
\begin{equation}
    Q_{\text{res}}(r; B') = Q(x; B', -\Delta/2, \Delta/2) ,
    \label{eq:scalar_quant2}
\end{equation}
where $B'$ is the number of bits used for the residual code. 

\begin{definition}
\label{def:lvq2}
We define the two-level Locally-adaptive Vector Quantization (LVQ-$B_1\mathsf{x}B_2$) of vector $\vect{x}$ as a pair of vectors $Q(\vect{x}), Q_{\text{res}}(\vect{r})$, such that
\begin{itemize}
    \item $Q(\vect{x})$ is the vector $\vect{x}$ compressed with LVQ-$B_1$,
    \item $Q_{\text{res}}(\vect{r}) = [Q_{\text{res}}(r_1; B_2), \dots, Q_{\text{res}}(r_d; B_2)]$,
\end{itemize}
where $\vect{r} = \vect{x} - \vect{\mu} - Q(\vect{x})$ and $Q_{\text{res}}$ is defined in \cref{eq:scalar_quant2}.
\end{definition}

No additional constants are needed for the second-level, as they can be deduced from the first-level ones. Given the first-level function in \cref{eq:LVQ_footprint}, the memory footprint of LVQ-$B_1\mathsf{x}B_2$ is
\begin{equation}
    \operatorname{footprint}(Q(\vect{x}),Q_{\text{res}}(\vect{r})) = \operatorname{footprint}(Q(\vect{x})) + d \cdot B_2 .
\end{equation}

\subsection{Integrating LVQ into graph-based indices}
We use first-level LVQ to search the graph. This improves the search performance by decreasing the effective bandwidth, determined by the number $B_1$ of bits transmitted from memory for each vector. The reduced number of bits might generate a loss in accuracy. When present, the second level, or compressed residuals, is used for a final re-ranking step, recovering part of the accuracy lost in the first level. Here, we replace \cref{alg:search_postprocessing} of \cref{alg:search} by a gather operation, that fetches $Q_{\text{res}}(\vect{r})$ for each vector $Q(\vect{x})$ in $\queue$, recomputes the similarity between the query $\q$ and each $Q(\vect{x}) + Q_{\text{res}}(\vect{r})$, and finally selects the top-$k$.
Moreover, we can safely build the graph from vectors compressed using LVQ, as we show in the next section.

\textbf{Adapting to shifts in the data distribution}
In the case of dynamic indices (supporting insertions, deletions and updates), a compression method should easily adapt to data distribution shifts. Search accuracy can highly degrade over time if the compression model and the index are not periodically updated. Such an update often involves running expensive algorithms (e.g., PQ~\cite{jegou2010product} involves running multiple instances of k-means). For LVQ, the model update is simpler, requiring recomputation of the dataset mean $\vect{\mu}$ and reencoding of the data vectors, operations that scale linearly with $n$, and do not require loading the full dataset in memory.

\section{Theoretical results on graph construction}
\label{sec:lightweight_indices}

\begin{figure}
  \centering
  \includegraphics[width=0.9\linewidth]{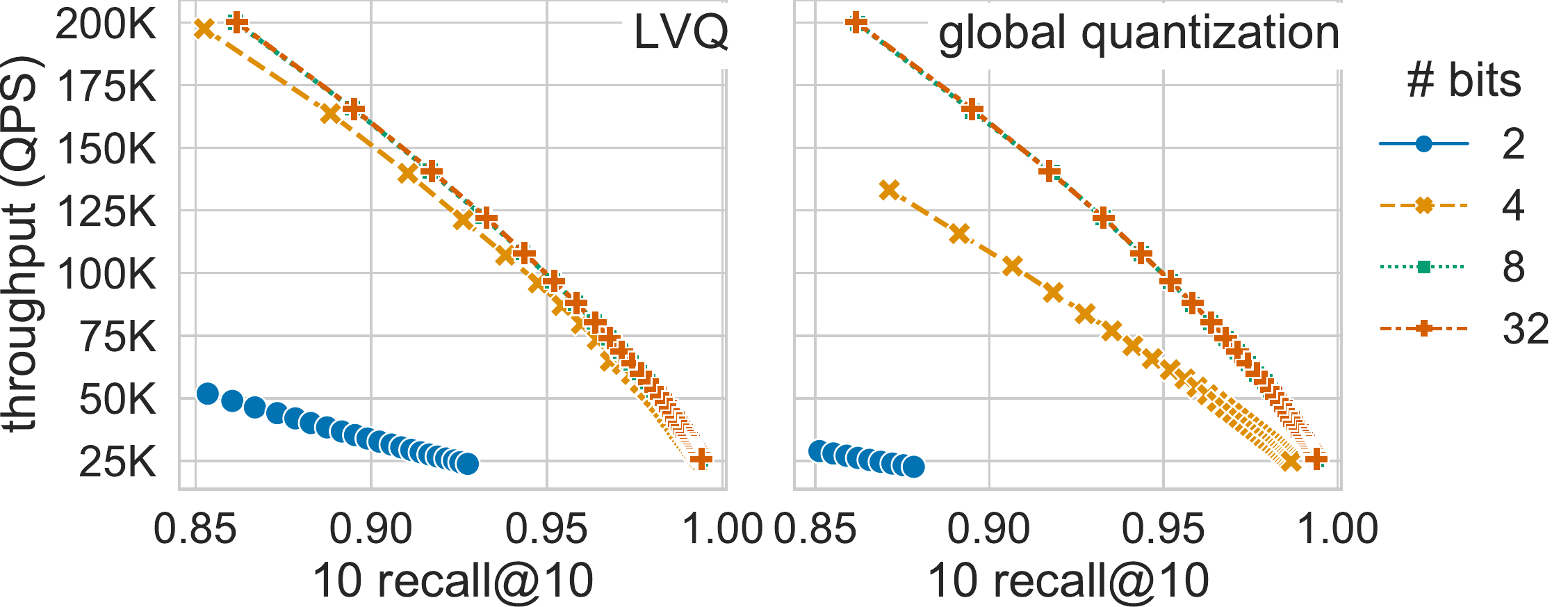}
  \caption{Search performance (conducted with float32-valued vectors to normalize for compute differences) on graphs built with vectors compressed using LVQ (left) and global quantization (right) with different number of bits for deep-96-100M. We observe almost no decrease in throughput in graphs built with 4 or more bits with LVQ (the curves with 8 and 32 bits overlap). In contrast, we observe a sharp drop in throughput for graphs built using global quantization with 4 bits. Thus, we can save memory by building graphs with LVQ-4 or LVQ-8 without affecting the search quality.
  }
  \label{fig:indexing_with_compr_searchAcc_deep1M}
\end{figure}

This section is devoted to showing that we can build a graph with LVQ-compressed vectors without impacting search accuracy, thus accelerating and reducing the footprint of the expensive index construction step. For example, for deep-96-1B (\cref{sec:datasets}), graph building requires at least 835GiB for a maximum out degree $R=128$. Notably, when graphs are built with LVQ-compressed vectors, the search accuracy is almost unchanged even when setting $B$ as low as 8 or 4 bits (see Figure~\ref{fig:indexing_with_compr_searchAcc_deep1M}). The minimum memory requirements (graph + dataset size) in GiB to construct a graph from full precision and from LVQ with $B=4$ bits are reported in \cref{table:index_build_compression}, where the memory reduction can be as high as 6.2x. In this section we explain why LVQ makes these savings possible, corroborating that this is indeed logical and even expected.

\begin{table}[t]
  \caption{Memory requirements (graph + vectors) for graph building with full-precision (FP) and with LVQ-$B$ vectors. Depending on the dataset and the graph maximum out-bound degree ($\maxOutDeg$=32,64,128), the memory reduction can reach 6.2x.}
  \label{table:index_build_compression}  
  \resizebox{\linewidth}{!}{
  \begin{tabular}{cccccccccc}
    \toprule
 & \multicolumn{3}{c}{deep-96-1B} & \multicolumn{3}{c}{text2Image-200-100M} & \multicolumn{3}{c}{DPR-768-10M} \\
    \cmidrule(lr){2-4} \cmidrule(lr){5-7} \cmidrule(lr){8-10}

    & \multicolumn{2}{c}{Size (GiB)} & \multirow{2}{*}{Ratio} & \multicolumn{2}{c}{Size (GiB)} & \multirow{2}{*}{Ratio} & \multicolumn{2}{c}{Size (GiB)} & \multirow{2}{*}{Ratio} \\
    \cmidrule{2-3} \cmidrule{5-6} \cmidrule{8-9}
    $\maxOutDeg$ &  FP & LVQ-4 &&  FP & LVQ-4 &&  FP & LVQ-4 \\
    \midrule
    32 & 477 & 168 & 2.84 & 864 & 216 & 4.00 & 298 & 48 & 6.20\\
    64 & 596 & 287 & 2.08 & 983 & 335 & 2.93 & 310 & 60 & 5.17\\
    128 & 834 & 525 & 1.59 & 1222 & 574	& 2.13 & 334 & 84 & 3.98\\
    \bottomrule
  \end{tabular}}  
\end{table}

There are three steps required to generate the adjacency list of each vertex in the graph: building, sorting and pruning the neighbors candidate list. We will begin with the latter, as it is the critical step to make the graph searchable.

\subsection{Graph pruning with LVQ} 

We will now characterize both theoretically and experimentally the errors introduced in the graph-pruning step when building the graph from vectors compressed with LVQ. These errors are mild and LVQ is fully compatible with the graph-pruning rule in \cref{alg:pruning_rule} of \cref{alg:pruning}.

Let us consider $\Cl$ the set of candidates for $\vect{x}$'s adjacency list in $\Gr$. The pruning process iterates through $\Cl$ and, at each step, adds to the set of out-neighbors of $\vect{x}$ its most similar vector $\vect{x}^*$, removing from $\Cl$ all the vector that are closer to $\vect{x}^*$ than to $\vect{x}$ (\cref{alg:pruning}).

When the similarity is Euclidean distance, i.e.,
$\simfun(\vect{x}, \vect{x}') = -\norm{\vect{x} - \vect{x}'}{2}$,
the pruning rule in \cref{alg:pruning_rule} of \cref{alg:pruning} becomes
\begin{equation}
    \alpha \norm{\vect{x}^* - \vect{x}'}{2} \leq \norm{\vect{x} - \vect{x}'}{2} .
    \label{eq:pruning_rule_euclidean}
\end{equation}

\begin{figure}
    \centering
    \begin{minipage}{0.45\linewidth}
         \centering
         \includegraphics[width=\textwidth]{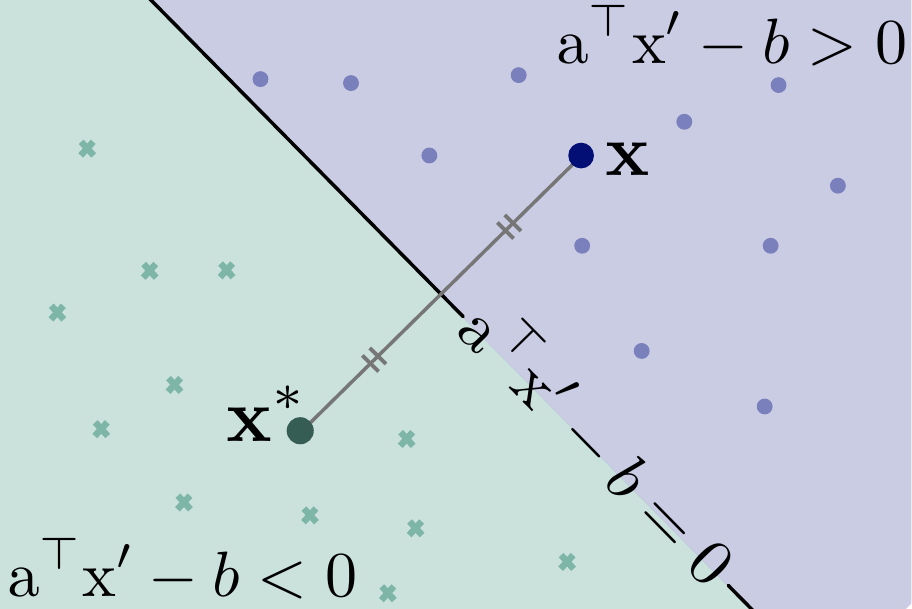}         
     \end{minipage}
     \begin{minipage}{0.5\linewidth}
         \centering
          \includegraphics[width=\textwidth]{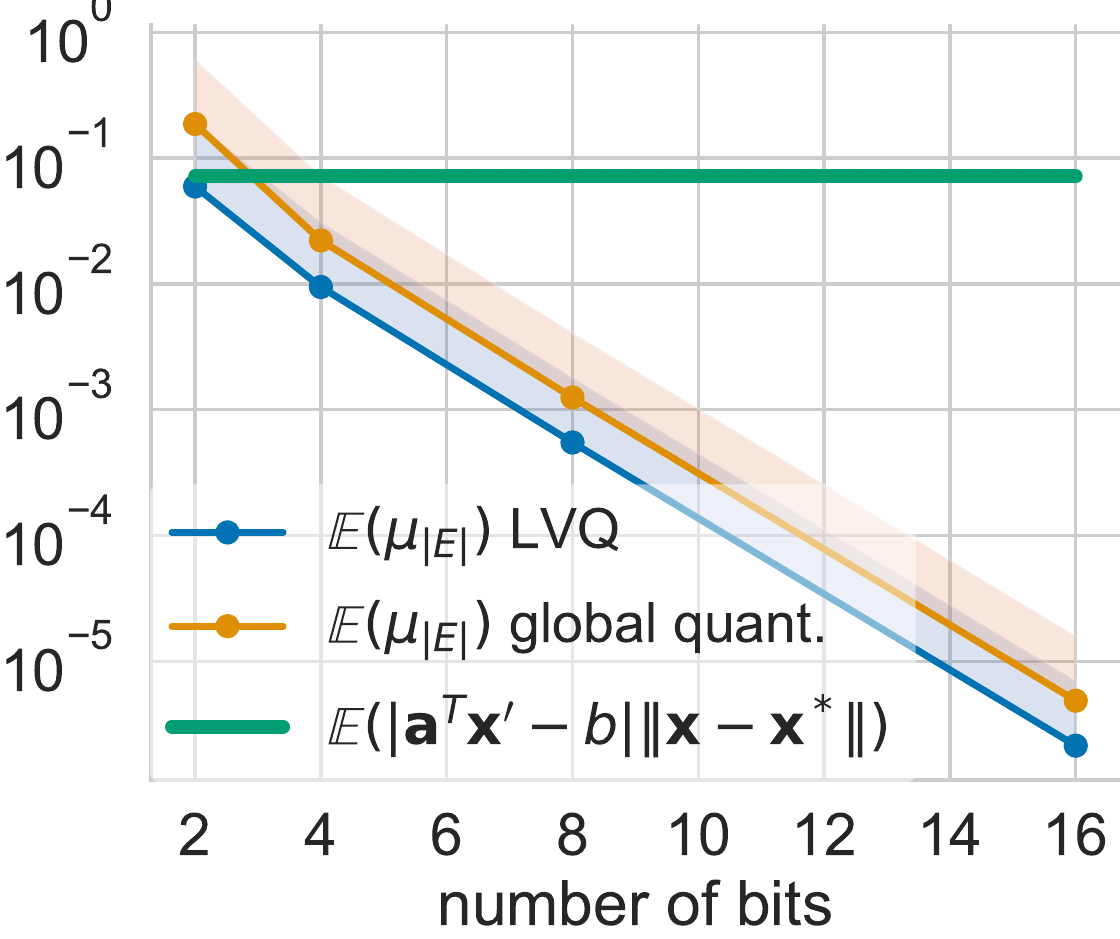}          
     \end{minipage}             

    \caption{(Left) For $\alpha=1$, the graph pruning rule (see \cref{alg:pruning}) can be interpreted as determining the perpendicular bisector hyperplane for $\vect{x}$ and $\vect{x}^*$, and eliminating the candidates (green crosses) that lie on the same (green) half-space as $\vect{x}^*$. This amounts to eliminating the vectors with $\sign(\transpose{\vect{a}}\vect{x}' - b) = -1$, see \cref{eq:sign_ax_b}.
    (Right) \cref{prop:pruning_rule_euclidean_equivalence} predicts that, under the green line, two graphs, one built with compressed vectors and one with full-precision, are equally accurate. We compare LVQ (in blue) and global quantization (in orange) with a varying number of bits (expectations computed across $10^6$ samples and error bands at $+3\sigma_{|\error|}$ for deep-96-100M). Empirically, graphs pruned using vectors compressed with LVQ-4 or LVQ-8 are well within the safe zone, while the errors bands for 4-bit global quantization are close to the threshold. This correlates with the search accuracy in \cref{fig:indexing_with_compr_searchAcc_deep1M} that is only slightly lower for LVQ-4 but severely degraded for 4-bit global quantization.}
    \label{fig:indexing_with_compression_pruning}
\end{figure}

Geometrically, as shown in \cref{fig:indexing_with_compression_pruning} (left), this is equivalent to determining the perpendicular bisector hyperplane for $\vect{x}$ and $\vect{x}^*$, and eliminating from $\Cl$ all vectors $\vect{x}'$ that lie on same half-space as $\vect{x}^*$. The pruning can be performed by computing
\begin{equation}
    \sign(\transpose{\vect{a}}\vect{x}' - b), \quad\text{with}\quad \vect{a}=\frac{\vect{x}-\vect{x}^*}{\|\vect{x} - \vect{x}^*\|_2}, 
    \quad b=\frac{\|\vect{x}\|_2^2 - \|\vect{x}^*\|_2^2}{2\|\vect{x} - \vect{x}^*\|_2} .
    \label{eq:sign_ax_b}
\end{equation}
and eliminating those vectors $\vect{x}'$ for which $\sign(a\vect{x}' - b) = -1$.

\begin{proposition}
\label{prop:pruning_rule_euclidean_equivalence}
When using Euclidean distance as the similarity function and $\alpha=1$, the graph pruning rule for full-precision vectors, \cref{eq:pruning_rule_euclidean}, and the one using vectors compressed with LVQ, i.e.,
\begin{equation}
    \norm{Q(\vect{x}^*) - Q(\vect{x}')}{2} \leq \norm{Q(\vect{x}) - Q(\vect{x}')}{2} ,
\end{equation}
are equivalent (in the sense of simultaneously holding) when
\begin{equation}
    |\transpose{\vect{a}}\vect{x}' - b| \cdot \|\vect{x} - \vect{x}^*\| \geq |\error| ,
    \label{eq:pruning_rule_euclidean_equivalence}
\end{equation}
where $\transpose{\vect{a}}$ and $b$ are defined in \cref{eq:sign_ax_b} and $\error \in \Real$ is an error that depends on the quantization error and the vectors $\vect{x}$, $\vect{x}^*$, and $\vect{x}'$.
\end{proposition}

The proof is in \cref{proof:compute_mu_sigma}~\cite{aguerrebere2023similarity}.

Classical signal processing theory dictates that under normal conditions the error introduced by a scalar quantization follows a uniform distribution. This uniformity is inherited by the quantization error in LVQ (see \cref{fig:uniform_quantization_error} in the supplementary material \cite{aguerrebere2023similarity}).

\begin{proposition}
\label{prop:pruning_rule_euclidean_equivalence_error}
Under the conditions in  \cref{prop:pruning_rule_euclidean_equivalence} and assuming a uniformly distributed quantization error, the error $\error$ is a normally distributed random variable with mean $\mu_{\error}$ and variance $\sigma^2_{\error}$, given by
\begin{align}
    \mu_{\error} = & \frac{d}{24}(\Delta_x^2-\Delta_{x^*}^2), \\
    \sigma^2_{\error} = &\frac{\Delta_{\vect{x}}^2}{12} \|\vect{x}'-\vect{x}\|^2 +  \frac{\Delta_{\vect{x}^*}^2}{12} \|\vect{x}'-\vect{x}^*\|^2 +  \frac{\Delta_{\vect{x}'}^2}{12} \|\vect{x}-\vect{x}^*\|^2 +  \nonumber \\
    & + \frac{d(\Delta_{\vect{x}}^4+\Delta_{\vect{x}^*}^4)}{720} + \frac{d \Delta_{\vect{x}'}^2 (\Delta_{\vect{x}}^2+\Delta_{{\vect{x}}^*}^2)}{144} .
    \label{eq:projection_error_mean_var}
\end{align}
where $\Delta_{\vect{x}}$, $\Delta_{\vect{x}}^*$, and $\Delta_{\vect{x}'}$ are the quantization steps for $\vect{x}$, $\vect{x}^*$, and $\vect{x}'$, respectively, given by \cref{eq:scalar_quant}.
\end{proposition}

The proof is in \cref{proof:compute_mu_sigma}~\cite{aguerrebere2023similarity}.

\begin{corollary}
Let $\Phi$ be the normal cumulative distribution function.
$|\error|$ follows a folded normal distribution parameterized by
\begin{align}
    \mu_{|\error|} &= \sigma_{\error} \sqrt{\frac{2}{\pi}} \exp{(-\mu_{\error}^2/2\sigma_{\error}^2) + \mu_{\error}(1 - 2\Phi(-\tfrac{\mu_{\error}}{\sigma_{\error}}))},
    \\
    \sigma^2_{|\error|} &= \mu_{\error}^2 + \sigma_{\error}^2 - \mu_{|\error|}^2 .
\end{align}
\end{corollary}

With these theoretical results in hand, we are now ready to characterize \cref{prop:pruning_rule_euclidean_equivalence} empirically, i.e., the number of bits $B$ (see \cref{def:lvq}) needed to run the pruning algorithm with minimal errors.
We generate triplets of vectors $\vect{x}$, $\vect{x}^*$ and $\vect{x}'$ that may be found during pruning from 1 million vectors taken from the deep-96-100M dataset presented in \cref{table:datasets}. For this, we select a vector $\vect{x}$ at random, find its (ground-truth) $\maxPrune$ nearest neighbors, and among those we first randomly sample $\vect{x}^*$, and then $\vect{x}'$ from those that are farther from $\vect{x}$ than $\vect{x}^*$. The results in \cref{fig:indexing_with_compression_pruning} (right) show that we can reduce $B$ in LVQ safely to 4 bits without affecting the pruning rule. With 2 bits, the error bars overlap and no guaranties can be given. These results are in agreement with the search accuracy we observe in Figure~\ref{fig:indexing_with_compr_searchAcc_deep1M}.

\subsection{Candidates selection and sorting with LVQ}

The candidates list consists of the $\maxPrune$ closest vectors among those visited in a search. If the compression error is small enough compared to the distance between $\vect{x}$ and its $\maxPrune$-th nearest neighbor, we could expect similar candidates lists when using compressed or full-precision vectors. To evaluate this, we analyze the $\maxPrune$ nearest neighbors of an exhaustive search with compressed vectors.

\begin{figure}[t]
  \centering
     \begin{minipage}[b]{0.46\linewidth}
         \centering
         \includegraphics[width=\textwidth]{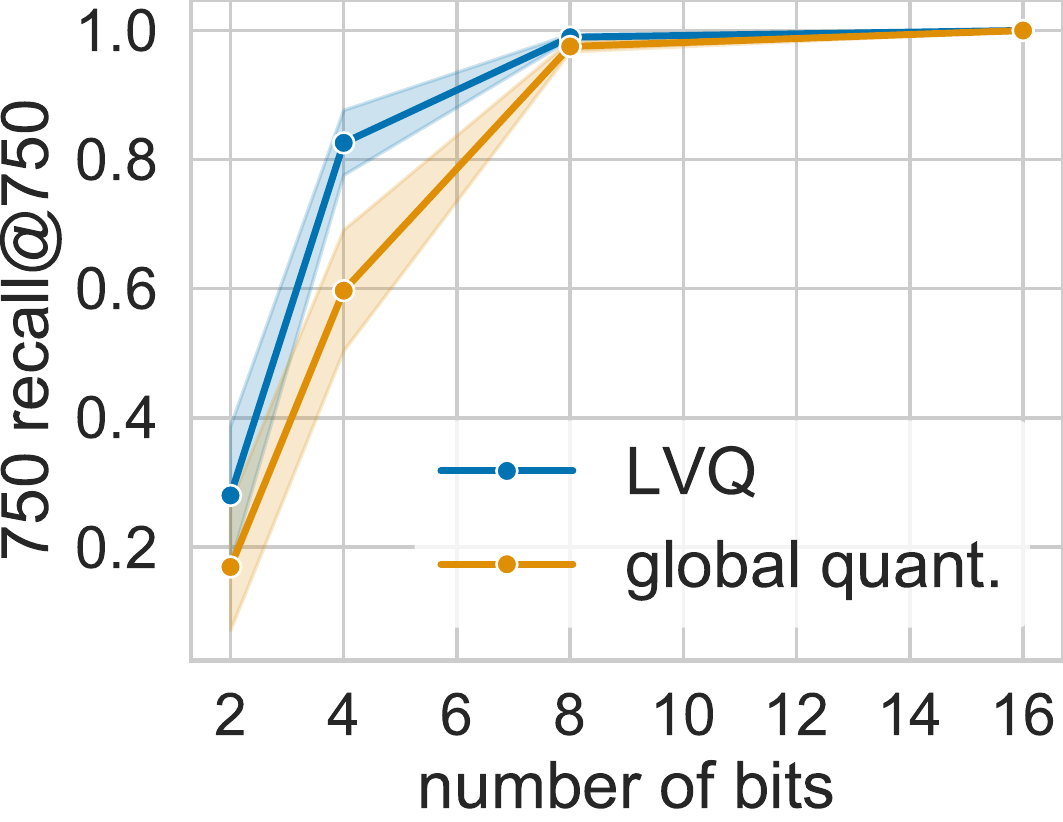}
     \end{minipage}
     \begin{minipage}[b]{0.46\linewidth}
         \centering
         \includegraphics[width=\textwidth]{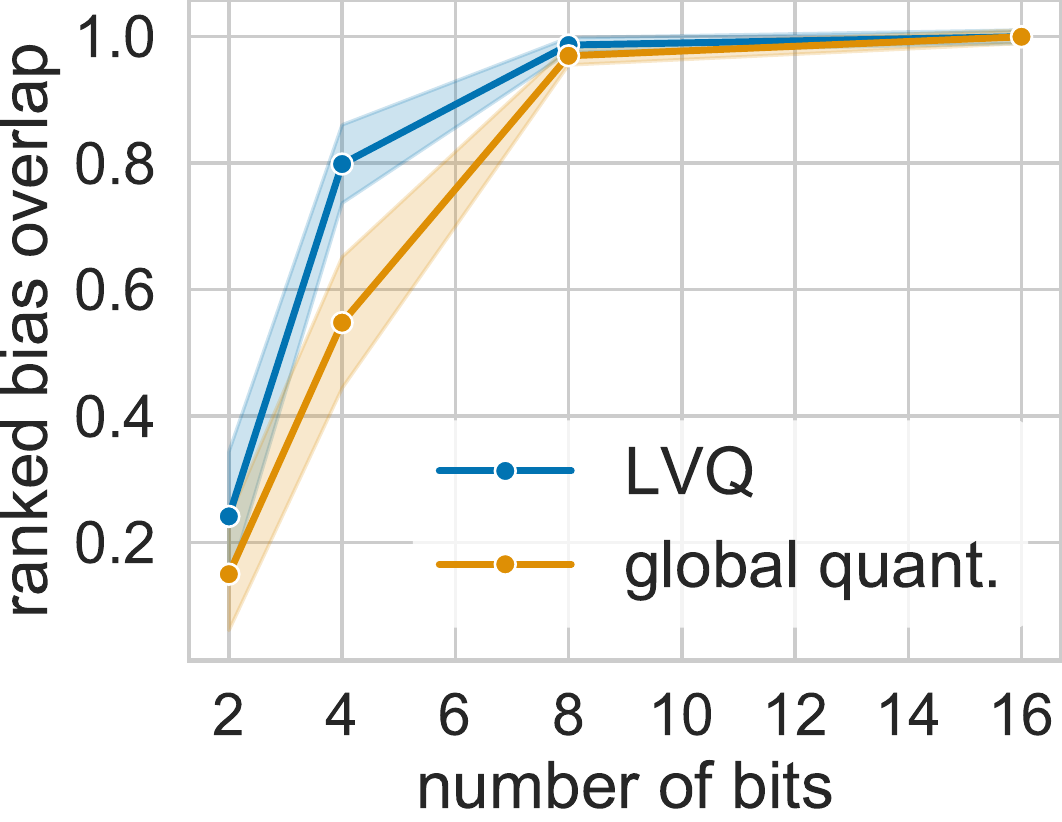}
     \end{minipage}        
  \caption{With four or more bits per value, LVQ does not introduce artifacts in the computation and sorting of the candidate lists (of length 750) used for graph pruning (bands at $\pm$ one standard deviation). To check the presence of the correct elements, we track 750-recall@750, confirming that LVQ stays above 0.8 for four bits, whereas the global quantization drops to 0.6. To check the order of the elements, we use Ranked Bias Overlap (RBO)~\cite{webber2010similarity}, obtaining similar numbers.}
  \label{fig:indexing_with_compression}
\end{figure}

In \cref{fig:indexing_with_compression} (left), we present average results for $10^5$ vectors chosen at random from the  dataset deep-96-100M ($\maxPrune{} = 750$). For LVQ with 8 and 16 bits, the recall is almost one. This suggests that there should be no difference between the candidates lists using full-precision or LVQ. At 4 bits, the recall for LVQ is 0.82, suggesting a high degree of agreement between the compressed and uncompressed lists of candidates. For global quantization, we observe a degraded recall of $0.6$, pointing to a loss of equivalence between the candidate lists. These results are in agreement with what we observe in \cref{fig:indexing_with_compr_searchAcc_deep1M}. The search throughput for graphs built with LVQ using 4 bits is almost unchanged from the baseline (32 bits). For graphs built with global vector quantization using 4 bits, the search throughput suffers significantly. 

To assess how LVQ affects the ordering of the candidate list elements, we use Ranked Bias Overlap (RBO)~\cite{webber2010similarity}, a standard metric to compare ranked lists. First, we find the two candidate lists, i.e., for LVQ-compressed and full-precision vectors, using exhaustive search. Next, we compute the RBO between the lists sorted according to the similarity between LVQ-compressed and full-precision vectors. Figure~\ref{fig:indexing_with_compression} shows the results  for $10^5$ vectors chosen at random in the dataset deep-96-100M. Again, for LVQ with 4, 8, and 16 bits, we observe a high-quality sorting. For global quantization at 4 bits, the sorting gets affected. 

\section{Implementation challenges}
\label{sec:systems}

As with most high-performance algorithms, a fine-tuned implementation is fundamental to realize the full potential of graph-based similarity search. We now describe a set of optimizations that are geared towards putting the system in its \emph{natural} memory bottlenecked regime and improving its performance. To illustrate the discussion with experimental results and ablation studies, we use the deep-96-100M dataset (see \cref{table:datasets} in \cref{sec:datasets}) and use the system described in \cref{sec:system_setup}.

\begin{figure*}[t]
    \centering
    \begin{minipage}[b]{0.19\textwidth}
        \subfloat[Impact of prefetching]{
            \includegraphics[width=\linewidth]{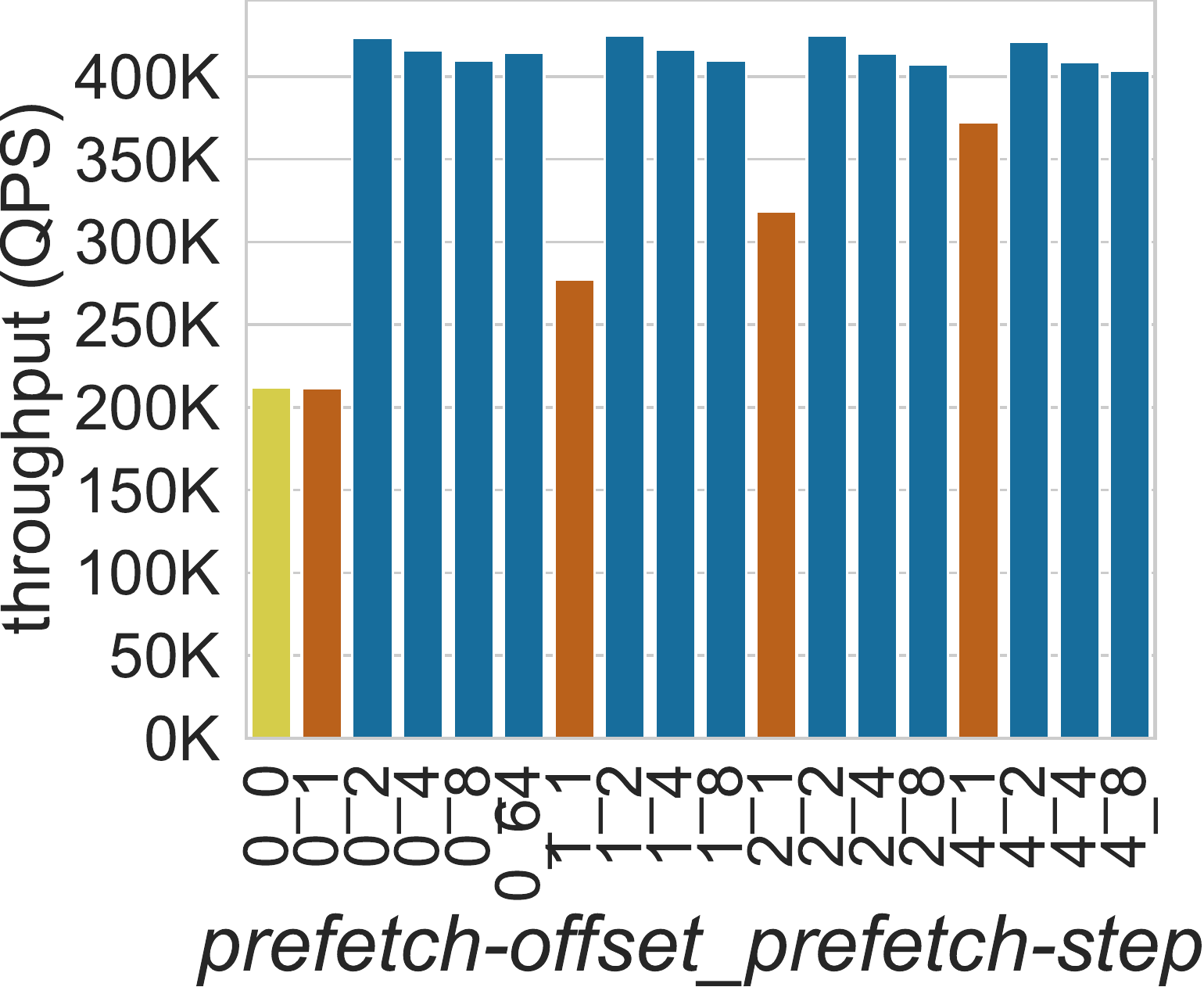}
            \label{fig:prefetching}
        }
     \end{minipage}%
     \begin{minipage}[b]{0.39\textwidth}
        \centering
        \subfloat[Effect of huge pages]{
            \includegraphics[width=0.505\linewidth]{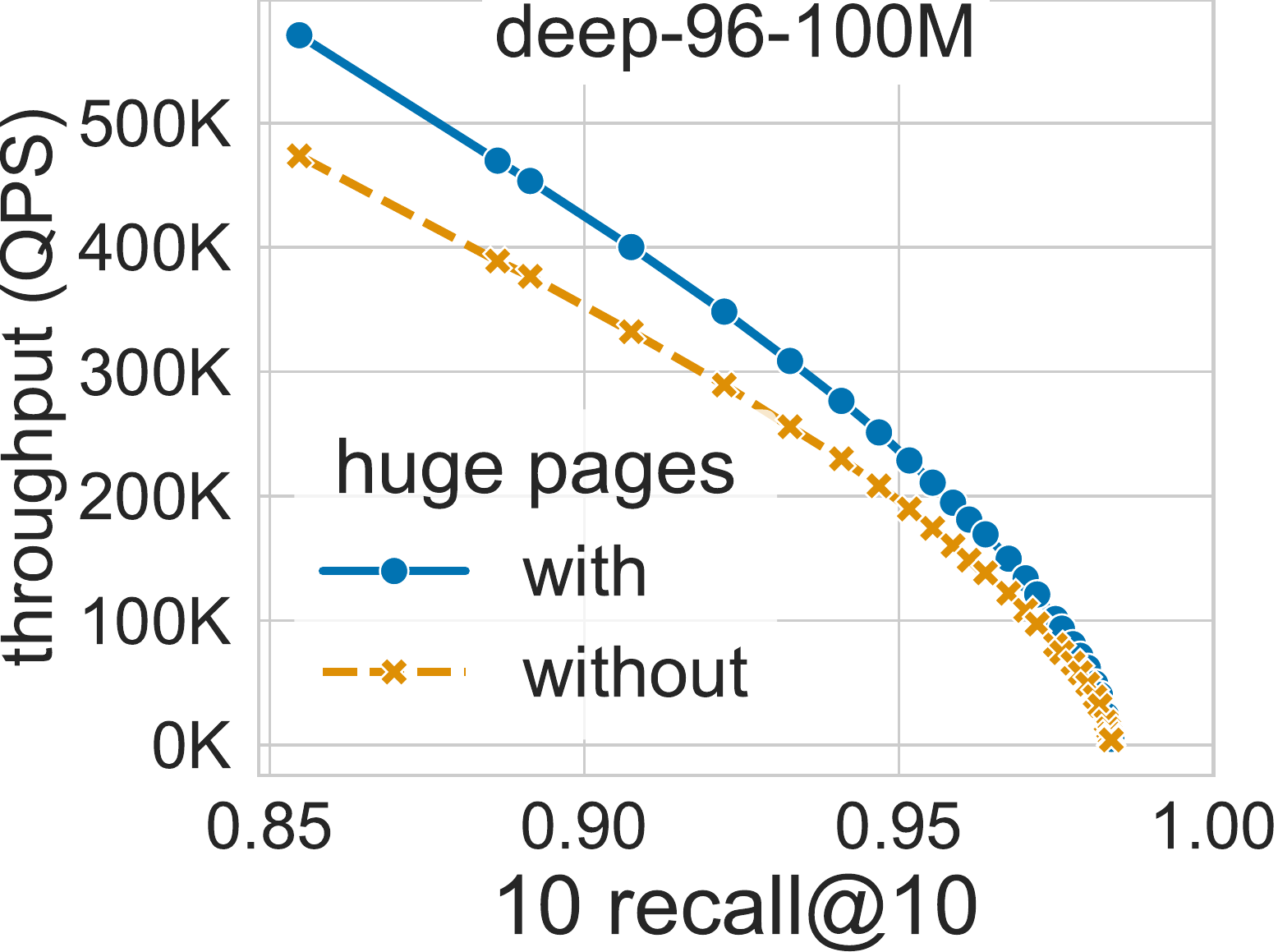}%
            \includegraphics[width=0.48\linewidth]{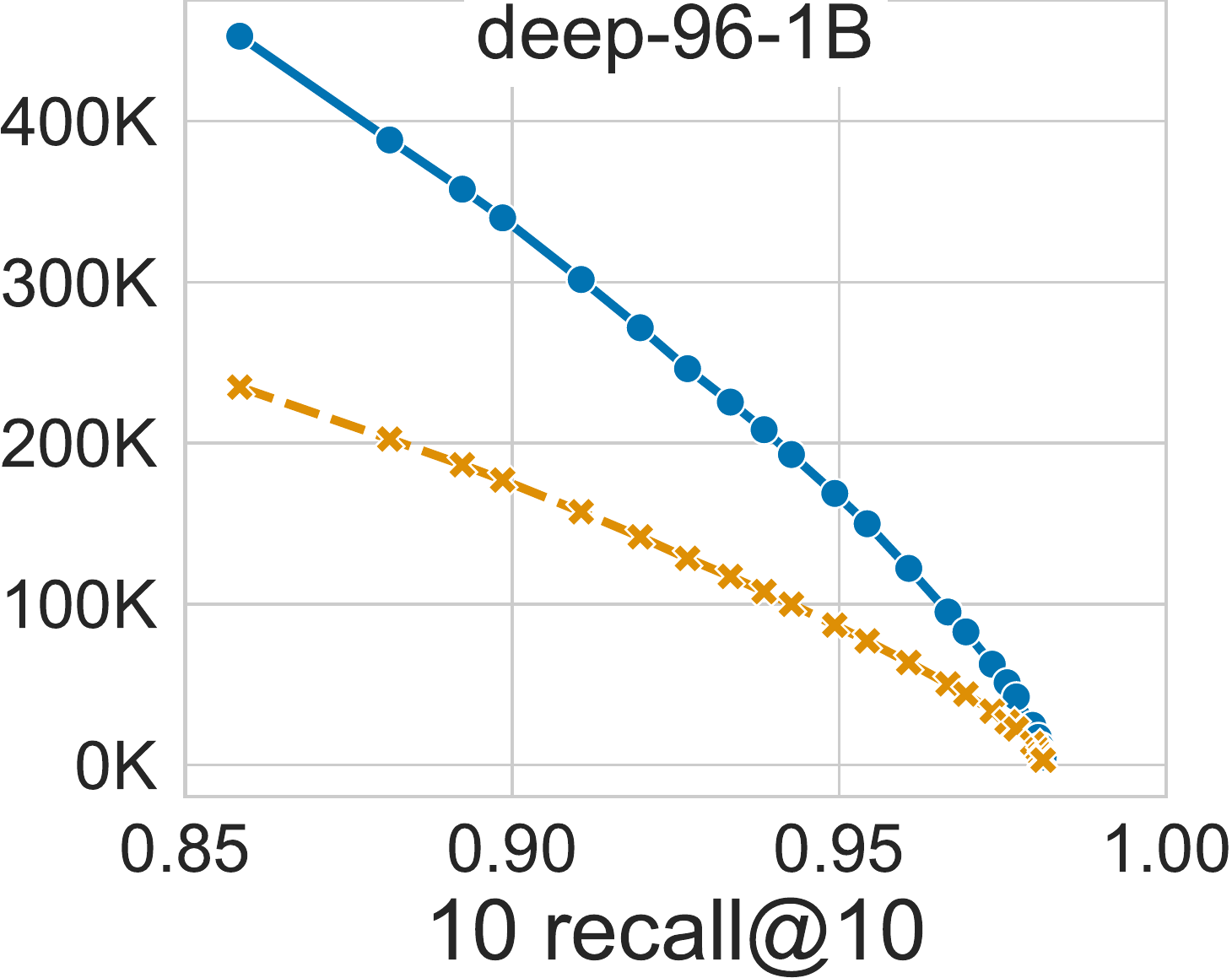}%
            \label{fig:huge_pages}
        }
     \end{minipage}%
     \begin{minipage}[b]{0.19\textwidth}
        \centering
        \subfloat[Bandwidth utilization]{
            \includegraphics[width=\linewidth]{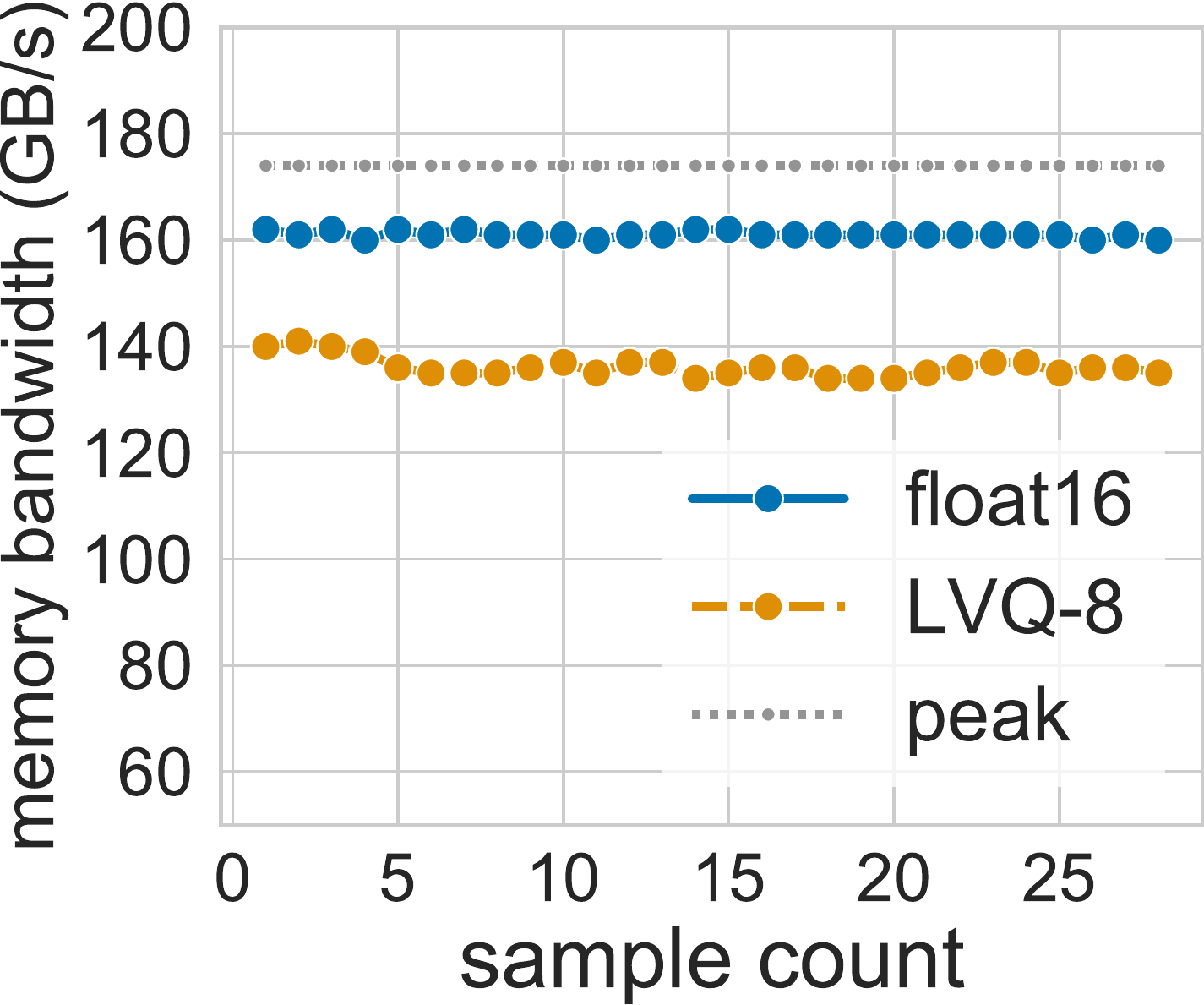}
            \label{fig:bandwidth}
        }
     \end{minipage}%
     \begin{minipage}[b]{0.19\textwidth}
        \centering
        \subfloat[Core scaling]{
            \includegraphics[width=\linewidth]{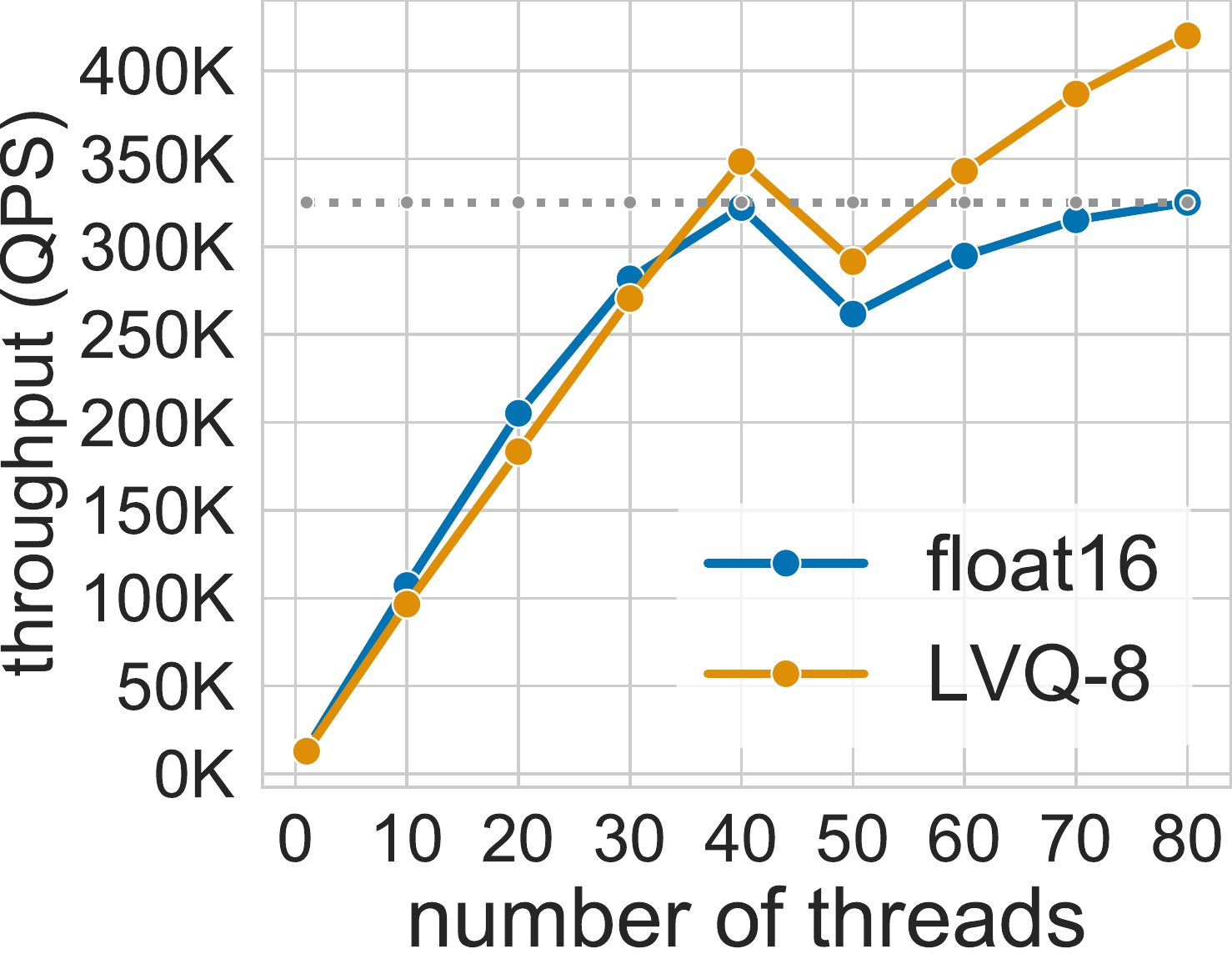}
            \label{fig:core_scaling}
        }
     \end{minipage}%
    \caption{
  \protect\subref{fig:prefetching} Our advanced prefetching, parameterized by \emph{prefetch-offset} and \emph{prefetch-step}, provides up to 2x performance gain over no-prefetch (yellow bar). Commonly used prefetch schemes using \emph{prefetch-step}=1 (red bars) show sub-optimal gains.
  \protect\subref{fig:huge_pages} By explicitly utilizing huge pages, we achieve significant performance gains as datasets grow: 20\% in deep-96-100M and 90\% in deep-96-1B.
  \protect\subref{fig:bandwidth} We reach 90\% and 78\% of the read-only peak bandwidth for float16 and LVQ-8 data types, respectively.
  \protect\subref{fig:core_scaling} We achieve good performance scaling with the number of threads, obtaining 23.5x and 33x gains over single-thread for float16 and LVQ-8 vectors, respectively. For float16 vectors, performance tops at 40 threads (on a system with 40 cores). Thanks to the reduced bandwidth, LVQ-8 performance continues to grow up to the maximum number (80) of hyperthreaded cores. Unless specified, experiments are done for deep-96-100M with a $10$-recall$@10$ of 0.9.}
  \label{fig:system_analysis}
\end{figure*}

\textbf{Efficient similarity calculations using LVQ with AVX.}
Computing the similarity between two vectors is a key kernel underpinning similarity search.
SIMD vector instructions can be used to efficiently implement distance computations for LVQ-$B$ and LVQ-$B_1\mathsf{x}B_2$.
We store compressed vectors as densely packed integers with scaling constants stored inline.
When 8-bits are used, native AVX instructions are used to load and convert the individual components into floating-point values which are combined with the scaling constants.
The case $B_1=B_2=4$ in LVQ-$B_1\mathsf{x}B_2$ requires a little more work, involving vectorized integer shifts and masking.
We fuse the decompression with the distance computation against the query vector.
This fusion, combined with loop unrolling and masked operations to tail elements, creates an efficient distance computation implementation that makes no function calls, decompresses the quantized vectors on-the-fly and accumulates partial results in AVX registers.

\begin{figure}
  \centering
  \includegraphics[width=0.85\linewidth]{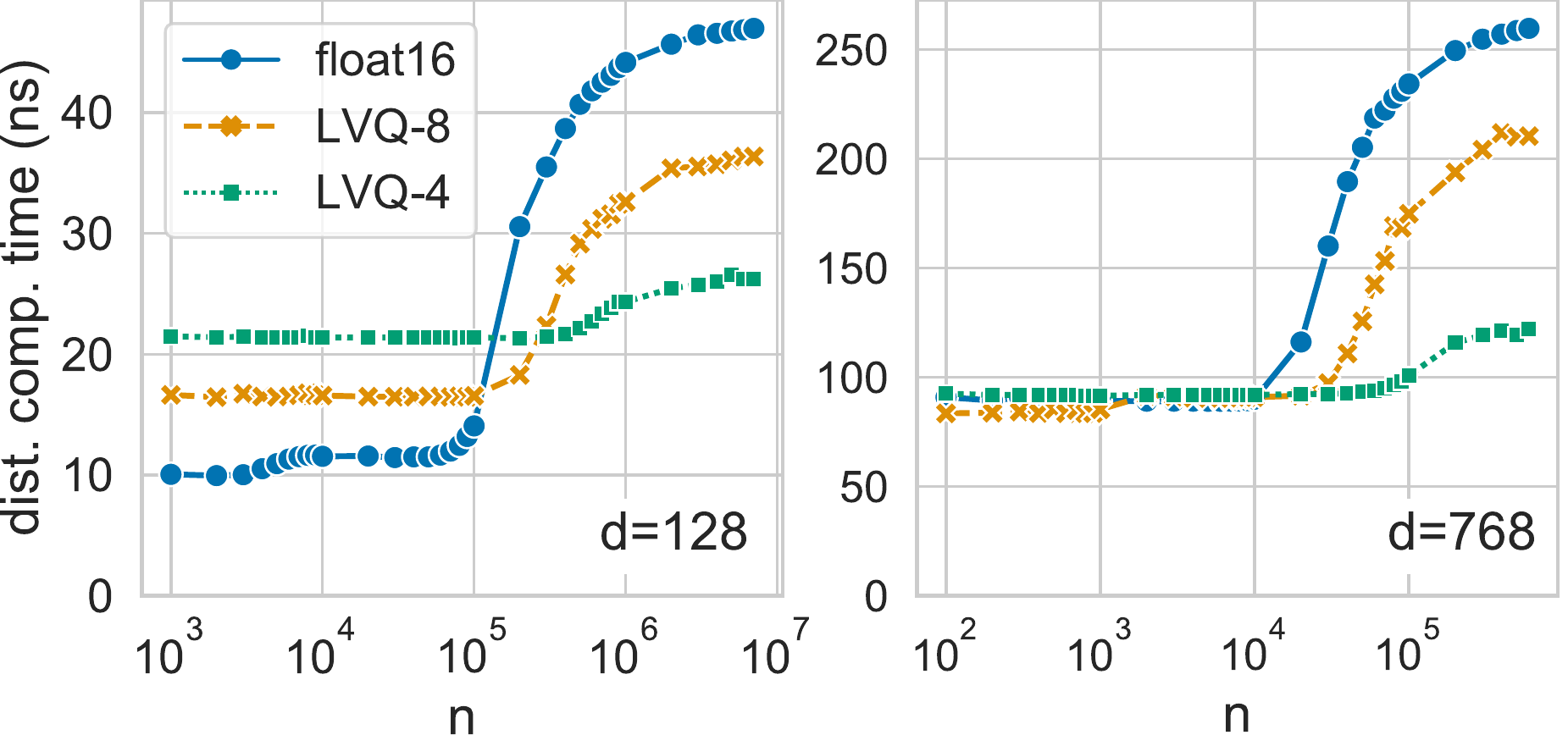}
  \caption{Mean similarity computation times using different encodings for $n$ vectors in $d$ dimensions ($d=128,768$ on the left and right). We use a sequential access pattern to minimize any memory side effects.
  In each case, the curve inflexion marks the point where the vectors do not fit in the L2 cache.
  When vectors fit in cache, distances with float16 values are 2x faster than those with LVQ-4.
  When vectors exceed the cache size, LVQ quickly becomes 2.12x faster than float16 due to the reduced number of fetched cache lines.}
  \label{fig:distance_computation_timing}
\end{figure}

A notable optimization is the ability to set the dimensionality at compile time (static) versus at runtime (dynamic). As the dimensionality of a dataset is fixed once and for all, setting it statically presents no detrimental aspects and it improves the compiler's ability to unroll loops in the similarity function kernel more extensively. We observe up to a 32\% performance speedup when using static versus dynamic dimensionality.
All in all, this implementation achieves \textasciitilde2x faster computation performance, as shown in \cref{fig:distance_computation_timing}.

\textbf{Advanced prefetching.}
Prefetching involves proactively moving data that will be accessed soon into the CPU cache.
This can either be done automatically in hardware, or manually through software instructions.
If done well, it can improve application throughput by lowering memory latency when data is finally accessed and by overlapping computation with memory access.
Due to the random data access pattern in graph-based search, hardware prefetchers are ineffective.
This leaves software prefetching.

When computing distances between a query and all neighbors of a vertex, we introduce two tunable parameters for prefetching data vectors:
\emph{prefetch-offset} that controls the lookahead offset in the list of vectors to prefetch, and \emph{prefetch-step} that sets the number of vectors to prefetch in each iteration.

 In \cref{fig:prefetching}, we observe that our scheme provides up to a 2x speedup over the no-prefetch case (baseline) where both parameters are zero. 
 A simple prefetch strategy (\emph{prefetch-step}=1) which simply prefetches the immediate next vector shows no performance gain over the baseline.
 However increasing \emph{prefetch-offset} boosts performance (red bars).
 Furthermore, schemes with \emph{prefetch-step} > 1 show similar performance.
 Although there is not a universally best combination, we found that \emph{prefetch-offset}=0 and \emph{prefetch-step}=2 works well in many cases. Last, to account for cache-unaligned dimensionality $d$, we selectively bring additional cache lines with no negative impact on the performance for aligned $d$. 

\textbf{Optimizing graph search.}
We now analyze opportunities for optimizing the overall implementation for \cref{alg:search}.
We use a sorted linear buffer to implement the queue $\queue$, storing whether a node has been explored inline with the node id and distance from the query.
For values of $\searchWin{}$ common in our datasets (a few dozens), we found this to be faster than using a heap-like data structure due to its hardware friendliness.
Common implementations of graph search use an associative data structure to track whether a node has been visited to avoid unnecessary memory accesses and compute.
With our optimized similarity computations and smart prefetcher in hand, keeping a visited set carries a performance regression.

The overall advantage of disabling the visited set depends both on the dimensionality $d$ and the CPU micro-architecture.
For example, with deep-96-100M on a Cascade Lake server CPU, disabling the visited set can improve throughput by 15-20\% whereas on an Ice Lake server, the improvement is capped at 2-3\%.
We observe no performance difference for the DPR-768-10M dataset (\cref{table:datasets}) with its larger dimensionality ($d=768$).

Finally, we parallelize the search across queries with each thread being responsible for a subset of the query batch, running a single-thread search routine for each query in the batch.
This is a common pattern for graph-based similarity search implementations.
Parallelizing the search for an individual query~\cite{peng2022speed-ann}, and understanding whether it presents actual benefits, remains as future work.

\textbf{Memory layout and allocation.}
\label{sec:memory_layout}
Modern computer systems use virtual memory~\cite{jacob1998virtual_memory} to provide process isolation and address space independence.  
Virtual memory addresses used by programs are translated to physical addresses through the use of page tables, a process that is accelerated in hardware using a Translation Look-aside Buffer (TLB) to cache recently used translations.

For billion-scale datasets, a TLB miss for each random access in \cref{alg:search} is nearly certain when using typical 4096~kB pages. Because the probability of the corresponding page table entry being resident in cache is nearly zero, this miss requires another access to main memory, which degrades performance. Using 2~MB or 1~GB huge pages greatly increases the probability that the missed page-table entry is in the cache~\cite{ram2021tlb_hardware}.
To that end, we avoid graph layouts that involve memory indirections (such as CSR or a list of lists), which further decrease the cache hit rate.

Consequently, we use large contiguous block allocations and implement explicit huge-page allocators.
\cref{fig:huge_pages} demonstrates a 20\% and 90\% performance gain by using huge pages at 0.90 recall in deep-96-100M and deep-96-1B, respectively. 

\textbf{System utilization.}
Our algorithm and its implementation achieve high system bandwidth and scale well with the number of cores. Our system's peak read-only memory bandwidth, measured with Intel{\textregistered}'s Memory Latency Checker~\cite{viswanathan2013intel}, is 174GB/s per socket. As shown in \cref{fig:bandwidth}, the average memory bandwidth achieved by our implementation during the search is 160GB/s and 135GB/s utilizing more than 90\% and 78\% of the peak bandwidth for float16-valued and LVQ-8 vectors, respectively. This high bandwidth utilization is the product of the system optimizations described above. 

Next, we present the performance scaling with the number of hardware threads used. Our system has 40 cores per socket with 2-way hyper-threading enabled (maximum 80 threads per socket). \cref{fig:core_scaling} shows that the QPS increases up to 40 threads in both float16 and LVQ-8 data types. In float16, however, the performance scaling slows down between 30 and 40 threads and saturates at 40 threads as the required memory bandwidth approaches the system peak. On the other hand, LVQ-8 reduces the memory bandwidth pressure and scales beyond 40 threads reaching its maximum value at 80 threads, fully utilizing the hyper-threading capabilities. As a result, our technique gains 23.5x and 33x performance over single-thread in float16 and LVQ-8, respectively. In both data types, the performance drops immediately after 40 threads due to hyperthreading. As a result, the threads sharing a core are slower and cap the overall search performance in batch mode. However, increasing the thread usage further leads to higher throughput outweighing the individual thread latency.  

\section{Experimental results}
\label{sec:results}

In this section, we present different empirical results clarifying the following topics.
We first discuss an exhaustive evaluation showing that our optimized graph-based search and LVQ, subsequently denoted as \iss{}, establishes a new SOTA for both small and large scale datasets. We also show that this performance benefits also come with memory savings.
Then, we address the benefits of LVQ over the standard Product Quantization.
Finally, we present an ablation study, where we compare different quantizers under our optimized graph-based search.
In all cases, LVQ-compressed vectors are padded to half cache lines ($p=32$, see \cref{sec:lvq}). We report the best out of 5 runs for each method~\cite{aumuller2020ann}.

\begin{table}
  \setlength{\tabcolsep}{3pt}
  \caption{Evaluated datasets, where $n$ ($n_q$) represent the number of vectors (queries) and $d$ their dimensionality. The space is measured in GiB. We generated DPR-768-10M from~\cite{karpukhin2020dense, raffel2020t5} (reproducibility details in Section ~\ref{appendix:DPR_code} of the supp.~material).}
  \label{tab:freq}
  \resizebox{\linewidth}{!}{
  \begin{tabular}{clS[table-format=3]cccS[exponent-mode=scientific, table-format=1e1, print-zero-exponent=true, 
  print-unity-mantissa=false] S[table-format=3.1]}
    \toprule
    & Dataset & {$d$} & {$n$} & Encoding & Similarity & {$n_q$} & {Space}\\
    \midrule
    
    \multirow{6}{*}{\begin{sideways}Small scale\end{sideways}}
    & gist-960-1M~\cite{irisa}     & 960 & 1M   & float32 & L2     & \num{1e3} & 3.6\\
    & sift-128-1M~\cite{irisa}     & 128 & 1M   & float32 & L2     & \num{1e4} & 0.5 \\
    & deep-96-10M~\cite{deep}     & 96  & 10M  & float32 & cos sim. & \num{1e4} & 3.6 \\  
    & deep-96-1M~\cite{deep}     & 96  & 1M  & float32 & cos sim. & \num{1e4} & 0.4 \\  
    & glove-50-1.2M~\cite{glove}   & 50  & 1.2M & float32 & cos sim. & \num{1e4} & 0.2 \\    
    & glove-25-1.2M~\cite{glove}   & 25  & 1.2M & float32 & cos sim. & \num{1e4} & 0.1 \\    
    \midrule
    
    \multirow{4}{*}{\begin{sideways}Large scale\end{sideways}}
    & DPR-768-10M     & 768 & 10M  & float32 & inner prod. & \num{1e4} & 28.6\\    
    & t2i-200-100M~\cite{text2image}    & 200 & 100M & float32 & inner prod. & \num{1e4} & 74.5 \\    
    & deep-96-100M~\cite{deep}    & 96  & 100M & float32 & cos sim.   & \num{1e4} & 35.8 \\
    & deep-96-1B~\cite{deep}      & 96  & 1B   & float32 & cos sim.  & \num{1e4} & 257.6 \\
  \bottomrule
\end{tabular}}
\label{table:datasets}
\end{table}

\subsection{Datasets}
\label{sec:datasets}

To cover a wide range of use cases, we evaluate our method on standard datasets of diverse dimensionalities ($d=25$ to $d=768$), number of elements ($n=10^6$ to $n=10^9$), data types and metrics (see Table~\ref{table:datasets}). In addition, we introduce a new dataset containing 10 million 768-dimensional embeddings generated with the dense passage retriever (DPR)~\cite{karpukhin2020dense} model. This dataset allows us to benchmark our method in a very high-dimensional setting, that has become ubiquitous in retrieval enhanced deep learning and most tasks that make use of large language models. We use text snippets from the C4 dataset~\cite{raffel2020t5} to generate: 10 million context DPR embeddings (base set) and 10000 question DPR embeddings (query set). We refer the reader to \cref{appendix:DPR_code} ~\cite{aguerrebere2023similarity} for reproducibility details about this dataset. For deep-96-100M and deep-96-1B, as the vectors have norm one, we compute the cosine similarity using Euclidean distance.

\subsection{System setup}
\label{sec:system_setup}

We run our experiments on two 2-socket systems.
Those in \cref{ssec:small_scale_benchmarks} run on 3rd generation Intel{\textregistered} Xeon{\textregistered} 8360Y @2.40GHz CPUs with 36 cores and 256GB DDR4 memory (@2933MT/s) per socket. All other experiments run on 3rd generation Intel{\textregistered} Xeon{\textregistered} Platinum 8380 @2.30GHz CPUs with 40 cores and 1TB DDR4 memory (@3200MT/s) per socket. Both systems run Ubuntu 22.04.\footnote{Performance varies by use, configuration and other factors. Learn more at \url{www.Intel.com/PerformanceIndex}. Performance results are based on testing as of dates shown in configurations and may not reflect all publicly available updates.
No product or component can be absolutely secure. Your costs and results may vary. Intel technologies may require enabled hardware, software or service activation. \textcopyright Intel Corporation.  Intel, the Intel logo, and other Intel marks are trademarks of Intel Corporation or its subsidiaries.  Other names and brands may be claimed as the property of others.}

We ran all experiments in a single socket to avoid introducing performance regressions due to remote NUMA memory accesses.

We use the \textit{hugeadm} Linux utility to preallocate a sufficient number of 1GB huge pages for each algorithm. Our implementation uses huge pages natively to reduce virtual memory overheads (see \cref{sec:memory_layout}). For a fair comparison, we run other methods with system flags enabled to automatically use huge pages for large allocations. 

\subsection{Performance on small scale search}
\label{ssec:small_scale_benchmarks}
We adopt the standard ANN-benchmarks~\cite{aumuller2020ann} protocol and consider small scale datasets of diverse dimensionality ($d$=25, 50, 96, 128, 960) and number of vectors ($n=10^6,\dots,10^7$). See details in Table~\ref{table:datasets}. We compare \iss{} to the SOTA algorithms for each dataset\footnote{We consider the SOTA results for single query mode, as those are the ones reported by ANN-benchmarks~\cite{aumuller2020benchmarking}. Last accessed on Feb. 15 2023.}, as well as to other widely adopted approaches, for two query batch sizes: one query at a time (single query mode) and full batch (total number of queries in the dataset). The evaluated algorithms are: Vamana~\cite{jayaram2019diskann}, HNSWlib~\cite{malkov2018efficient}, FAISS-IVFPQfs~\cite{johnson2019billion}), NGT-qg~\cite{iwasaki2018nearest} and ScaNN~\cite{guo2020accelerating}. NGT-qg is not included in the query batch mode evaluation because the available implementation did not support multi-query processing. Following ANN-benchmarks, we generate Pareto curves of QPS vs. recall for a series of parameter settings. For the graph-based methods (Vamana, HNSWlib and \iss{}), we build graphs with $\maxOutDeg=32, 64, 128$\footnote{This corresponds to $M=16,32,64$ in HNSW parameter notation.}. For IVFPQfs, ScaNN and NGT-qg we use the provided configuration settings~\cite{aumuller2020benchmarking}. For \iss{}, we include various LVQ settings (LVQ-8, LVQ-4x4, LVQ-4x8, and LVQ8x8). As explained in \cref{sec:system_setup}, our implementation explicitly uses huge pages when available. However, system flags to automatically use huge pages did not work in the ANN-benchmarks~\cite{aumuller2020ann} Docker image. Therefore, huge pages were not allocated during these experiments to ensure equal conditions for all methods. When using cosine similarity, we follow the standard approach of normalizing the vectors and running the search using Euclidean distance.

\begin{figure*}
  \centering  

  \includegraphics[width=0.9\textwidth]{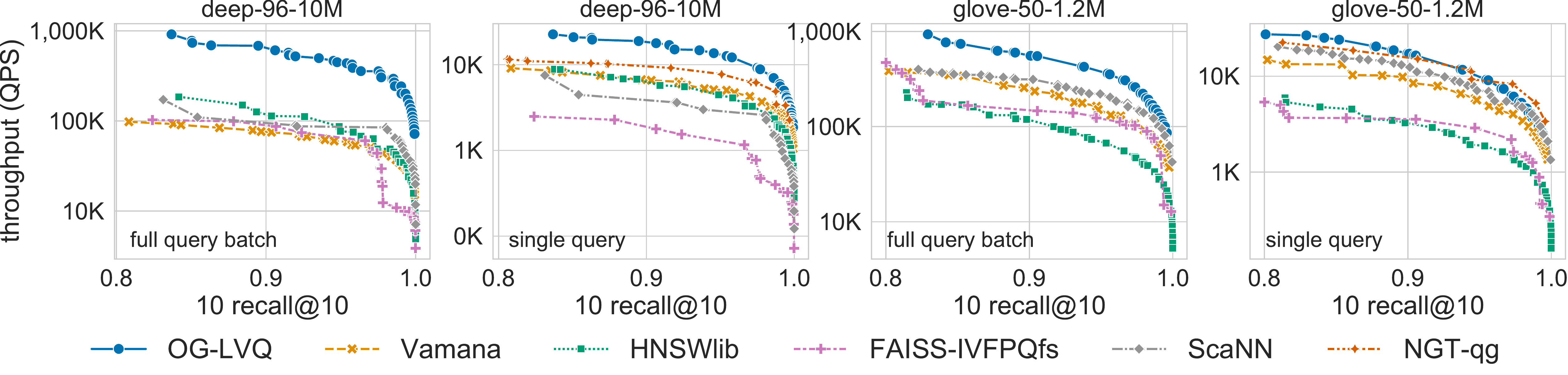}
  \caption{Benchmarking results for small scale datasets (deep-96-10M and glove-50-1.2M). In batch mode (first and third plots), \iss{} prevails by a large margin. In single-query mode (second and fourth plots), \iss{} takes the lead in one and is second on the other one. Numerical comparisons for 10-recall@10 of 0.9 are shown in \cref{tab:small_scale_benchmarks_qps_ratios}.}
  \label{fig:small_scale_benchmarking}
\end{figure*}

In~\cref{tab:small_scale_benchmarks_qps_ratios}, we observe that \iss{} outmatches the competition in full query batch mode, where it supersedes its closest competitor by 1.15x to 5.49x across all datasets with their unique sizes and dimensionalities. In single query mode, \iss{} still wins in 3 out 5 cases and NGT-qg takes the other two. The performance gains of \iss{} are consistent across the entire recall range, as shown in~\cref{fig:small_scale_benchmarking} for the deep-96-10M and glove-50-1.2M datasets (all datasets are included in \cref{fig:small_scale_benchmarking_suppMat} in the supplementary material \cite{aguerrebere2023similarity}). \Cref{fig:small_scale_benchmarking_50recall_suppMat,fig:small_scale_benchmarking_100recall_suppMat} in the supplementary material \cite{aguerrebere2023similarity} show similar results for 50 recall@50 and 100 recall@100, respectively.

\subsection{Performance on large scale search}
\label{ssec:large_scale_benchmarking}
We adopt the big-ann-benchmarks~\cite{simhadri2022results} framework to run our large-scale studies in full batch query mode. For this study, we consider the datasets in \cref{table:datasets} with a large footprint. We compare \iss{} to four widely adopted approaches: Vamana~\cite{jayaram2019diskann}, HNSWlib~\cite{malkov2018efficient}, FAISS-IVFPQfs~\cite{johnson2019billion}, and ScaNN~\cite{guo2020accelerating}. We use the following parameter setting to build Vamana graphs for all the datasets: $\maxOutDeg=128$ (we use $\maxOutDeg=126$ for deep-96-1B), $\alpha=1.2$ and $\alpha=0.95$ for L2 distance and inner product, respectively. For \iss{}, we include various LVQ settings (LVQ-8, LVQ-4x4, LVQ-4x8, and LVQ8x8). For HNSWlib, we build all graphs with a window search size of 200 and $R=128$\footnote{This corresponds to $M=64$ in HNSW parameter notation.}, except deep-96-1B for which we must reduce $R$ to 96 to fit the working set size in 1TB memory. For FAISS-IVFPQfs, as the build-time is long for deep-96-1B, we pre-build an index with $\text{nlist} = 32768$ and $\text{bins}=d/2$. While for t2i-200-100M and DPR-768-10M, indices are built on the fly with combinations of $\text{nlist} = \{512, 1024, 4096, 8192\}$ and $\text{nbins}=\{d/4, d/2, d\}$. To achieve higher recall rates, we enable re-ranking in FAISS-IVFPQfs and sweep $\text{nprobe}=\{1,5,10,50,100,200\}$ and $\text{k for re-ranking} = \{0,10,100,1000\}$, at runtime. For ScaNN, we use the parameters setting recommended by the authors ($\text{n\_leaves}=\sqrt{n}, \text{avq\_threshold}=0.2, \text{dims\_per\_block}=2$), as that was the best among several evaluated settings, and vary the runtime parameters ($\text{leaves\_to\_search}=2-1000, \text{reorder}=20-1000$). Finally, we did not include NGT~\cite{iwasaki2018nearest} in the evaluation as the algorithm designed for large-scale datasets (NGT-QBG) achieves low accuracy saturating at 0.86 recall even for a small 1-million vectors dataset. 

\begin{figure*}
  \centering    
   \includegraphics[width=0.9\textwidth]{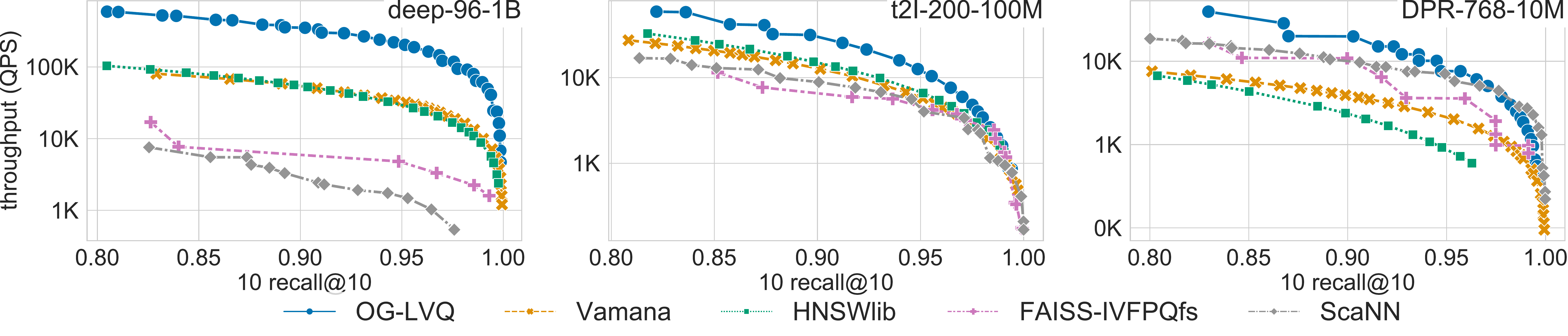}
  \caption{For large scale datasets, the proposed \iss{} outperforms its competitors across all datasets and almost across the entire recall range. For a 10-recall@10 of 0.9, the performance of \iss{} is 1.8x to 6.5x better than the alternatives. Furthermore, notice that, for larger dimensionalities (second and third plots), the second-best method is not the same in both cases. This highlights the versatility of our approach.} 
  \label{fig:large_scale_benchmarking}
\end{figure*}

Figure~\ref{fig:large_scale_benchmarking} shows \iss{}'s significant performance advantage across recall values for deep-96-1B, with a 6.5x higher throughput over the closest competitor for 10-recall@10 of 0.9. For the datasets that use inner product, the advantage is still present for recall values below 0.97 for t2i-200-100M and 0.95 for DPR-768-10M. There, \iss{} achieves 2.0x and 1.8x higher throughput than the closest competitor for 10 recall@10 of 0.9 on t2i-200-100M and DPR-768-10M, respectively. In this case, the \iss{} performance for very high recall values is on par with the alternatives. This phenomenon is not due to the quantization error, as it is also present in the graph search with full-precision vectors (not shown in the figure). Understanding this phenomenon will require further studies.
Similar results are observed for 50 recall@50 and 100 recall@100 in \cref{fig:large_scale_benchmarking_50_100recall_suppMat} in the supplementary material \cite{aguerrebere2023similarity}.

\begin{table*}[t]  
  \caption{The proposed \iss{} shows significant gains in small scale datasets at 0.90 10-recall$@$10, clearly winning 8 out the 10 tested cases (the best throughput in each case is shaded). Note that the alternative with the second-highest throughput is not consistently the same, showing the versatility of \iss{}. For the other schemes, we include their raw throughput and the ratio between ours and theirs. The geometric mean across all datasets highlights the overall superiority of \iss{}. }
  \label{tab:small_scale_benchmarks_qps_ratios}

  \sisetup{detect-weight=true,detect-inline-weight=math}
  
  \resizebox{\linewidth}{!}{\begin{tabular}
  {lS[table-format=7] *{4}{S[table-format=6]@{\hspace{0.5\tabcolsep}}S[table-format=1.2]} S[table-format=5] *{5}{S[table-format=5]@{\hspace{0.5\tabcolsep}}S[table-format=1.2]}}
    \toprule
 & \multicolumn{9}{c}{Full query batch} & \multicolumn{11}{c}{Single query}  \\
    \cmidrule(lr){2-10} \cmidrule(lr){11-21}
 & {OG-LVQ} & \multicolumn{2}{c}{Vamana} & \multicolumn{2}{c}{HNSWlib} & \multicolumn{2}{c}{FAISS-IVFPQfs} & \multicolumn{2}{c}{ScaNN} & {OG-LVQ} & \multicolumn{2}{c}{Vamana} & \multicolumn{2}{c}{HNSWlib} & \multicolumn{2}{c}{FAISS-IVFPQfs} & \multicolumn{2}{c}{ScaNN} & \multicolumn{2}{c}{NGT-qg}\\
\cmidrule(lr){2-2} \cmidrule(lr){3-4} \cmidrule(lr){5-6} \cmidrule(lr){7-8} \cmidrule(lr){9-10} \cmidrule(lr){11-11} \cmidrule(lr){12-13} \cmidrule(lr){14-15} \cmidrule(lr){16-17} \cmidrule(lr){18-19} \cmidrule(lr){20-21}
 & {QPS} & {QPS} & {Ratio} & {QPS} & {Ratio}  & {QPS} & {Ratio}  & {QPS} & Ratio & {QPS} & {QPS} & Ratio & {QPS} & {Ratio} & {QPS} & {Ratio} & {QPS} & {Ratio} & {QPS} & {Ratio} \\
\midrule
deep-96-10M & \cellcolor[gray]{0.9}648122 & 75543 & 8.58 & 118053 & 5.49 & 89550 & 7.24 & 95266 & 6.80 & \cellcolor[gray]{0.9}18440 & 6643 & 2.78 & 6347 & 2.91 & 1895 & 9.73 & 3881 & 4.75 & 9600 & 1.92\\
gist-960-1M & \cellcolor[gray]{0.9}56717 & 14586 & 3.89 & 11313 & 5.01 & 34640 & 1.64 & 16588 & 3.42 & 1857 & 992 & 1.87 & 725 & 2.56 & 827 & 2.24 & 883 & 2.10 & \cellcolor[gray]{0.9}4420 & 0.42\\
glove-25-1.2M & \cellcolor[gray]{0.9}1224266 & 1062091 & 1.15 & 418850 & 2.92 & 703996 & 1.74 & 443237 & 2.76 & \cellcolor[gray]{0.9}34118 & 29382 & 1.16 & 13304 & 2.56 & 7993 & 4.27 & 28230 & 1.21 & 25902 & 1.32\\
glove-50-1.2M & \cellcolor[gray]{0.9}558606 & 246095 & 2.27 & 117585 & 4.75 & 148793 & 3.75 & 313337 & 1.78 & \cellcolor[gray]{0.9}17268 & 9151 & 1.89 & 3237 & 5.33 & 3564 & 4.85 & 12582 & 1.37 & 15676 & 1.10\\
sift-128-1M & \cellcolor[gray]{0.9}852705 & 323014 & 2.64 & 200634 & 4.25 & 464048 & 1.84 & 189196 & 4.51 & 21969 & 13807 & 1.59 & 9100 & 2.41 & 6852 & 3.21 & 4856 & 4.52 & \cellcolor[gray]{0.9}23117 & 0.95\\
\midrule
Geometric mean &&& 2.97 && 4.38 && 2.70 && 3.42 &&& 1.78 && 3.00 && 4.29 && 2.37 && 1.02\\
    \bottomrule
  \end{tabular}}
\end{table*}

\subsection{LVQ: Fast graph search with small footprint}
\label{sec:LVQ_memory_reduction}

We now show that combining a highly optimized graph-based method (see \cref{sec:systems}) with LVQ provides high search performance with a fraction of the memory.
\Cref{fig:search_with_reduced_memfootprint_1B} shows the search throughput as a function of the memory footprint (measured as the maximum resident main memory usage during search) of different algorithms for deep-96-1B at 0.9 10-recall@10 (similar results are shown in \cref{fig:search_with_reduced_memfootprint} in the supplementary material \cite{aguerrebere2023similarity} for deep-96-100M). In the case of the graph-based methods (\iss{}, Vamana, HNSWlib), the memory footprint increases with the graph size given by the maximum number of outbound neighbors $R=\{32, 64, 128\}$. In the case of FAISS-IVFPQfs, the memory footprint remains almost constant for all combinations of the considered parameters (nlist=\{4096, 8192,16384\}, nbins=\{48,96\}, nprobe=\{1,5,10,50,100,200\}, k=\{0,10,100,1000\} for re-ranking), increasing by only 7\% (from 42GiB to 45GiB for deep-96-100M) when using 48 or 96 PQ segments (nbins), respectively. A similar behavior is observed for ScaNN, as it uses the same index design. LVQ-compressed vectors are padded to half cache lines ($p=32$, see \cref{sec:lvq} for details), as it improves performance and has a low impact on the overall memory footprint (e.g., 5\% larger footprint for deep-96-1B with $R$=128).

These results show that \iss{} can use a much smaller graph ($\maxOutDeg{}=32$) and still outperform its competitors: (A) for deep-96-1B by 2.3x, 2.2x, 20.7x, and 43.6x in throughput with 3x, 3.3x, 1.7x, and 1.8x less memory (\cref{fig:search_with_reduced_memfootprint_1B}), and (B) for deep-96-100M by 3.2x, 2.7x, 7.4x, and 11.5x in throughput with 3.1x, 3.3x, 1.8x and 1.9x less memory (\cref{fig:search_with_reduced_memfootprint} in the supplementary material \cite{aguerrebere2023similarity}), with respect to Vamana, HNSWlib, FAISS-IVFPQfs, and ScaNN, respectively.
\iss{}'s superiority in QPS and memory footprint is consistent across all recall values (see \cref{fig:search_with_reduced_memfootprint_recall} in the supplementary material \cite{aguerrebere2023similarity}). \iss{}, with a memory footprint of 24GiB (LVQ-8 and $\maxOutDeg{}=32$), outperforms all its competitors up to recall 0.97. In the high accuracy regime, \iss{} with LVQ-4x8 is on par with the competition with a 2.5x smaller footprint than the best alternative.

\subsection{LVQ versus PQ for exhaustive search}
\label{sec:OPQ_PQ_ISS_comparison}

Product quantization (PQ)~\cite{jegou2010product} is the most popular compression technique for similarity search. PQ is often used at high compression ratios, and is combined with re-ranking using full-precision vectors to achieve a reasonable recall~\cite{johnson2019billion}. Subramanya et al.~\cite{jayaram2019diskann} use PQ in this fashion for graphs stored in SSDs. When working with in-memory indices, we have a tough choice: either keep the full-precision vectors in memory (in addition to the compressed codes) and defeat compression altogether, or discard them and experience a severely degraded accuracy. Note that keeping the full-precision vectors in a less expensive storage (e.g., SSD), is not an option as it would severely degrade search throughput. This limits the usefulness of PQ for in-memory graph-based search.

\begin{figure}
  \centering
     \begin{minipage}[b]{0.46\linewidth}
         \subfloat[All compression ratios.]{%
         \includegraphics[width=\textwidth]{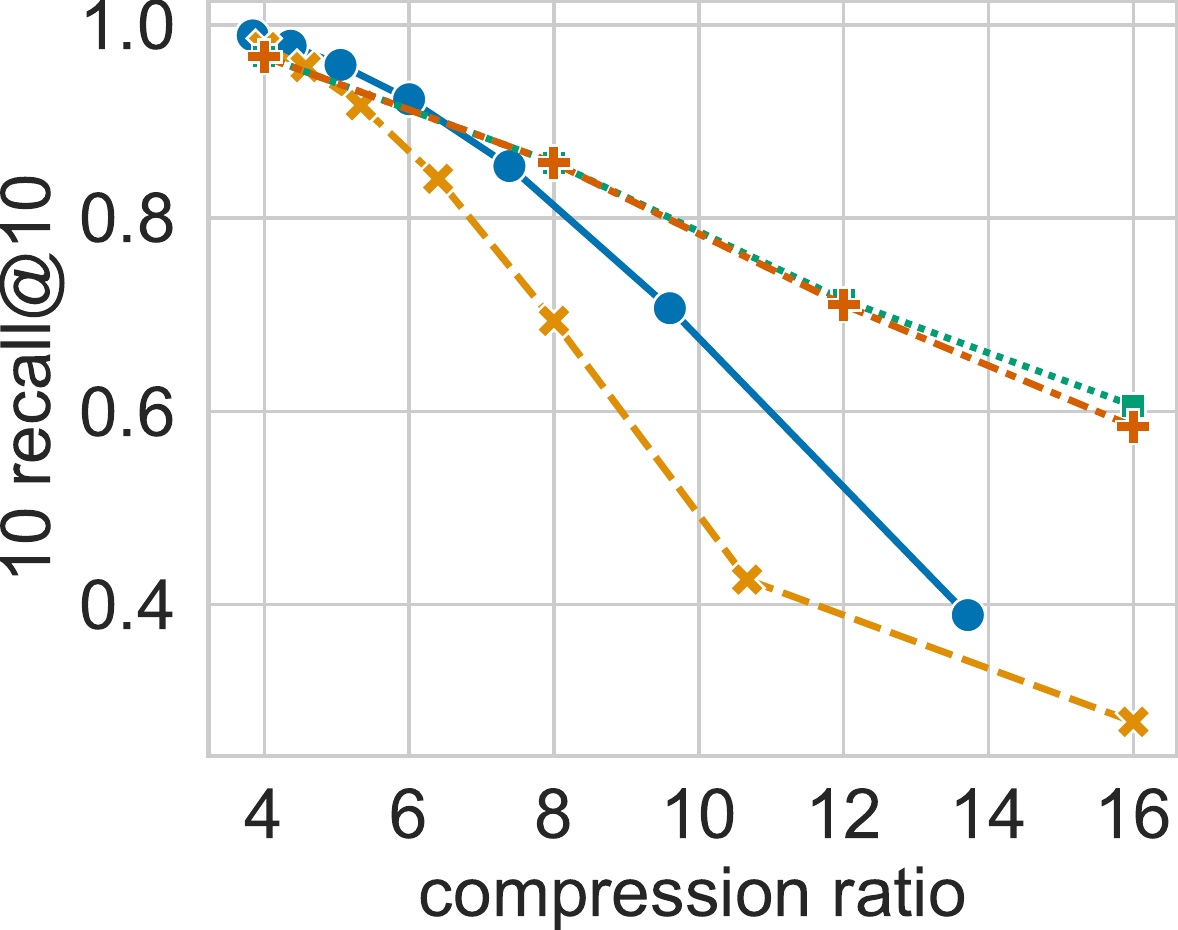}%
         }%
     \end{minipage}%
     \hfill%
     \begin{minipage}[b]{0.46\linewidth}
         \subfloat[Zoom on smaller ratios.]{%
         \includegraphics[width=\textwidth]{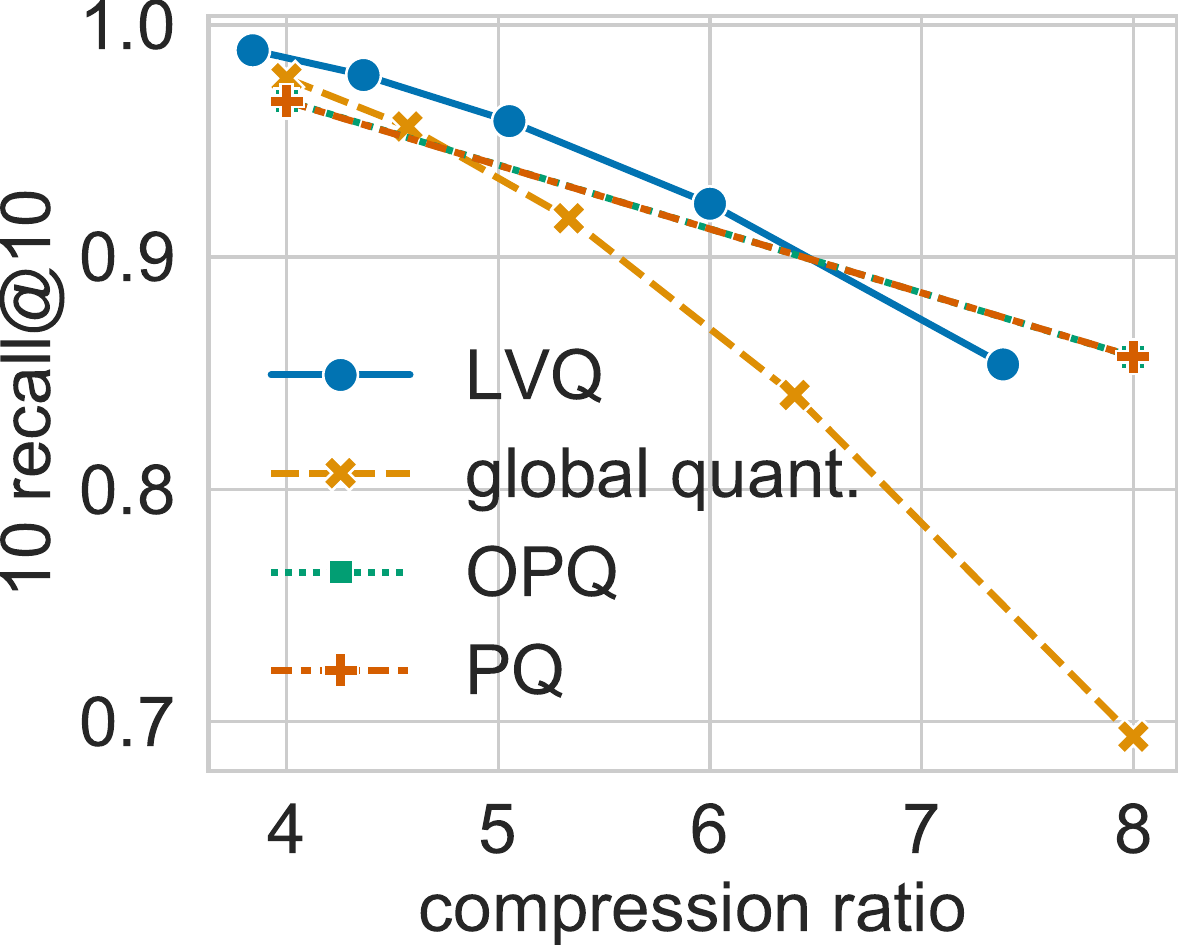}%
        }%
     \end{minipage}%
     
  \caption{Exhaustive search accuracy with different vector compression approaches for the deep-96-1M dataset. At lower compression ratios (<6x), see detail on the right, where PQ~\cite{jegou2010product} and OPQ~\cite{ge2013optimized} may not require re-ranking, LVQ achieves higher accuracy than both of them (their curves overlap). LVQ comes with the additional advantage of having much faster similarity calculations. At higher compression ratios, re-ranking with full-precision vectors is required to reach a reasonable accuracy, defeating the original purpose of compression. The compression ratio for LVQ is defined in \cref{eq:compression_ratio}, with the same formula for PQ and OPQ, defining their footprint as their number of segments (using 256 centroids per segment).}
  \label{fig:OPQ_PQ_ISS_comparison}
\end{figure}

Figure~\ref{fig:OPQ_PQ_ISS_comparison} shows the recall achieved by running an exhaustive search with vectors compressed using PQ, OPQ~\cite{ge2013optimized} (a PQ variant), LVQ and global quantization for the deep-96-1M dataset (a similar behavior is observed for other datasets). PQ and OPQ perform better for smaller footprints. This occurs because only 1 to 3 bits can be allocated in LVQ to each value, presenting an overly coarse encoding. On the other hand, PQ and OPQ can exploit correlations across different dimensions, making a better use of the available bits. However, the achieved recall (below 0.7) is not acceptable in modern applications, requiring re-ranking and thus limiting the usefulness of PQ as stated above. At higher footprints, where re-ranking can be avoided, our vector quantization achieves higher accuracy, while introducing almost no overhead for distance computations.

\subsection{An ablation study comparing quantizers}
\label{ssec:compression_ablation_study}

\begin{figure}
  \centering  
  \includegraphics[width=0.8\linewidth]  
  {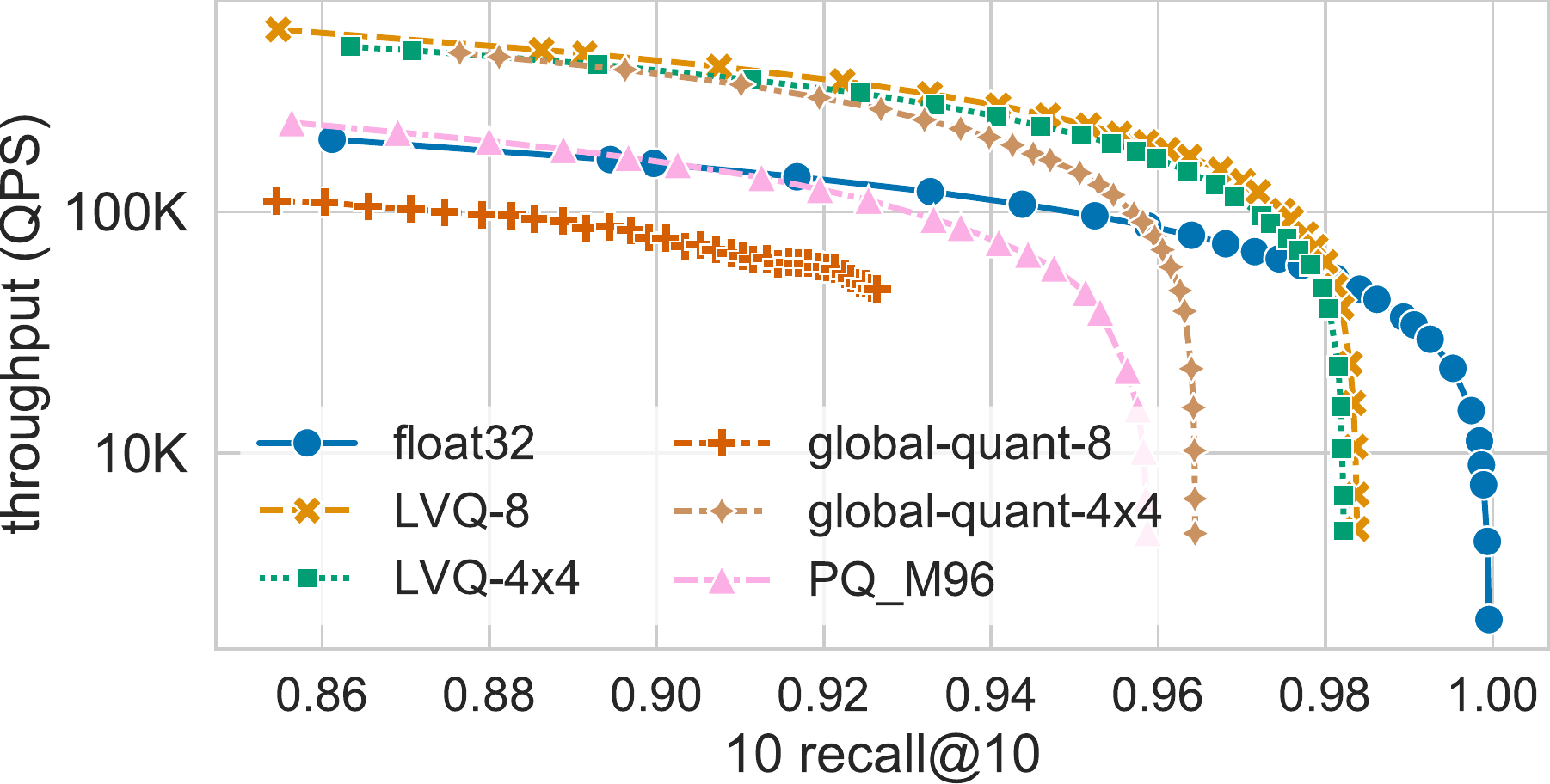}

\caption{LVQ outcompetes the alternatives for vector compression for deep-96-100M (graph with out-degree $\maxOutDeg=128$). Both variants of LVQ are superior to global quantization with non-vector specific parameters (see \cref{sec:lvq}). We use the standard full-precision (i.e., float32-valued) vectors and the ubiquitous Product Quantization (PQ)~\cite{jegou2010product} with 96 segments as our baselines (only this number of segments does not warrant a re-ranking with full-precision vectors at the end of the search). LVQ is clearly superior up to the very high recall of 0.98 (higher recall is achieved with LVQ-4x8, with a slightly higher footprint, as shown in \cref{fig:compression_ablation_variousDatasets}). \iss{} with LVQ-8 has a 5\% larger memory footprint (its vectors are padded to half cache lines) than global quantization with 8 bits and PQ.}
  \label{fig:compression_ablation_deep100M}
\end{figure}

We now analyze the search performance advantage that stems from LVQ and compare it to PQ. To assess the quality of vector compression for graph-search, we integrate PQ into our optimized graph-based search and compare the search performance of both compression techniques under the same implementation.
We set the number of PQ segments to 96, with 8 bits per segment, as that is the only setting that achieves high enough search recall without the need of re-ranking (see \cref{sec:OPQ_PQ_ISS_comparison}).
We also evaluate one and two-level compression schemes using global scalar quantization.
For LVQ, we consider three settings: LVQ-8 (one-level with 8 bits), LVQ-4x4 (two-levels with 4 bits each) and LVQ-4x8 (two-levels with 4 and 8 bits). We also evaluate one and two-level compression schemes using global scalar quantization.
Finally, we also tried the quantization method by Ko et al.~\cite{ko2021low}, using the suggested parameter settings. This scheme saturates at 0.85 10-recall@10 for deep-96-100M, never reaching our standard of 0.9. At 0.80 10-recall@10, LVQ-8 is 56\% faster. We thus omit it from further comparisons. 

LVQ achieves higher throughput and higher recall than global scalar quantization. As shown in \cref{fig:compression_ablation_deep100M} for the deep-96-100M dataset, the maximum recall achieved by global quantization is 0.96 whereas LVQ goes over 0.98. Note that, with a larger memory footprint, LVQ-4x8 reaches higher accuracy as shown in \cref{fig:compression_ablation_variousDatasets} for deep-96-1B. A similar behavior is observed for other datasets. Moreover, the storage overhead of using local constants is small for most datasets (e.g., 4\% for deep-96-100M). These two aspects confirm the advantage of LVQ over global quantization. 

For deep-96-100M, LVQ-4x4 performs slightly worse than LVQ-8, showing that the residual encoding is not the best option in this case (\cref{fig:compression_ablation_deep100M}). However, as shown in~\cref{fig:compression_ablation_variousDatasets} and~\cref{tab:compression_results}, this depends on the dataset. As expected, using two levels has an advantage for higher dimensional datasets as seen for DPR-768-10M. Depending on the dataset, LVQ gives \iss{} a QPS boost ranging from 2.6x to 4.7x, vector storage reductions of up to 3.8x and total memory-footprint reduction (considering the space occupied by the graph and the vectors) of up to 2.7x.~\Cref{tab:compression_results} includes float16 encoding as well, confirming the large advantage of LVQ over this compression.
For smaller and larger dimensionalities, one or two-level LVQ takes the lead, respectively.

In \cref{fig:compression_ablation_deep100M}, LVQ-8 outperforms PQ at all recall values with 5.2x more QPS at 0.9 recall for similar compression ratios (4x for PQ vs. 3.84 for LVQ-8). See \cref{sec:related_work} for a detailed explanation.

\begin{figure}
  \centering
  \includegraphics[width=\linewidth]{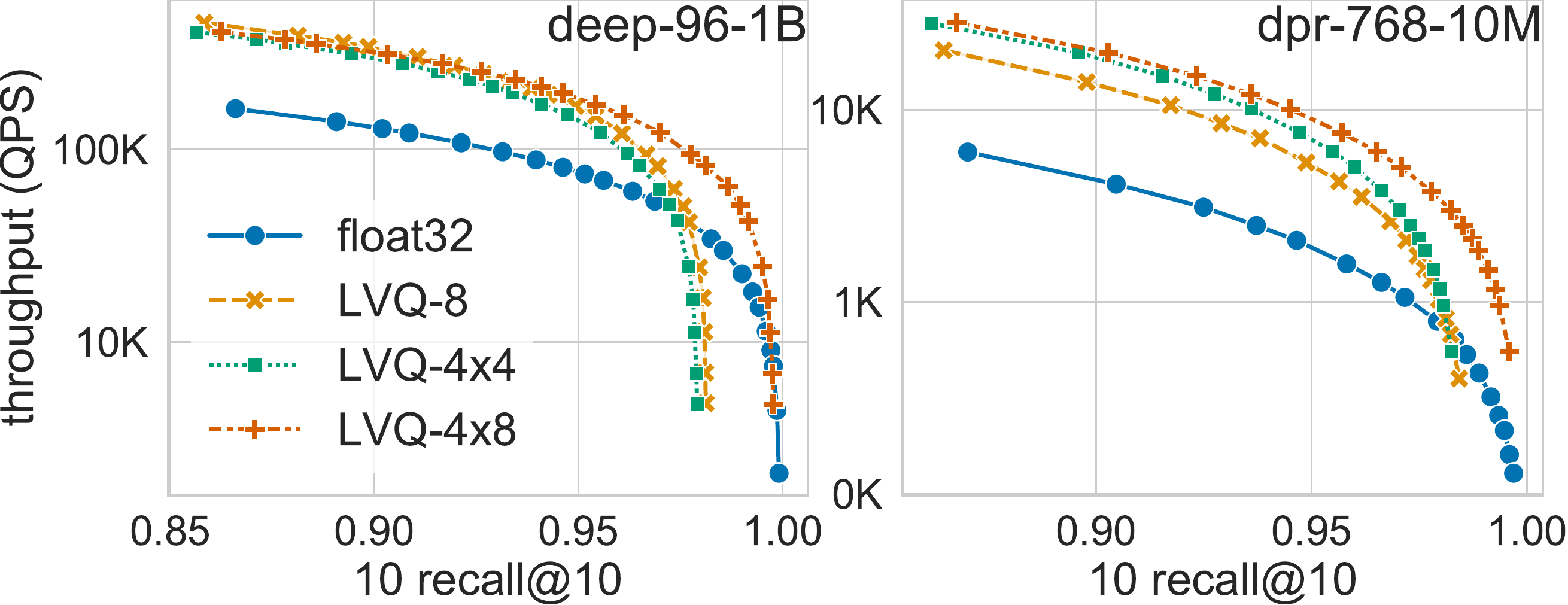}
  \caption{The proposed LVQ increases the performance of graph-baseed search when compared to standard full-precision vectors for deep-96-1B (left) and DPR-768-10M (right). The choice of one vs. two-level LVQ depends on the dimensionality of the dataset. For smaller dimensionalities ($d=96$), where there is not such a bandwidth pressure, LVQ-8, with its faster compute prevails, see \cref{sec:OPQ_PQ_ISS_comparison}. For larger dimensionalities ($d=768$), the additional bandwidth reduction vastly compensates for the extra compute.}
  \label{fig:compression_ablation_variousDatasets}
\end{figure}

\begin{table}[t]
  \caption{The performance of graph search benefits significantly from LVQ. The use of LVQ easily beats the standard baseline of using full-precision (i.e., float32-valued) vectors and even float16-valued vectors, which have recently shown to provide competition-winning performances~\cite{simhadri2022results}. The memory-footprint ratio (MR) measures the space occupied by the graph ($\maxOutDeg=128$) and the float32-valued vectors relative to the space occupied by the graph and the LVQ-compressed vectors. For larger dimensionalities ($d=768$), LVQ highly reduces the memory requirements achieving a large MR, and the additional bandwidth reduction from LVQ-4$\mathsf{x}$4 and LVQ-4$\mathsf{x}$8 provides a meaningful performance boost over LVQ-8. The compression ratio (CR) for LVQ is defined in \cref{eq:compression_ratio}. LVQ achieves up to 3.8x vector compression (for improved performance, LVQ-compressed vectors are padded to half-cache lines). The largest QPS, CR, and MR improvements in each case are shaded.}
  \label{tab:compression_results}
  \resizebox{\linewidth}{!}{
  \begin{tabular}{cccccccccc}
    \toprule
 & \multicolumn{3}{c}{deep-96-1B} & \multicolumn{3}{c}{t2i-200-100M} & \multicolumn{3}{c}{DPR-768-10M} \\
 \cmidrule(lr){2-4} \cmidrule(lr){5-7} \cmidrule(lr){8-10}
 w.r.t. float32 & QPS & CR & MR & QPS & CR & MR & QPS & CR & MR \\ 
\midrule
float16 & 2.1x & 2.0x & 1.3x & 1.9x & 2.0x & 1.4x & 1.7x & 2.0x & 1.8x\\
LVQ-8 & \cellcolor[gray]{0.9}2.6x & 3.0x & \cellcolor[gray]{0.9}1.4x & 2.9x & \cellcolor[gray]{0.9}3.6x & \cellcolor[gray]{0.9}1.8x & 3.1x & \cellcolor[gray]{0.9}3.8x & \cellcolor[gray]{0.9}2.7x \\
LVQ-4$\mathsf{x}$4 & 2.3x & \cellcolor[gray]{0.9}3.4x & \cellcolor[gray]{0.9}1.4x & 2.2x & 3.5x & \cellcolor[gray]{0.9}1.8x & 4.3x & \cellcolor[gray]{0.9}3.8x & \cellcolor[gray]{0.9}2.7x \\
LVQ-4x8 & 2.5x & 2.4x & 1.3x & \cellcolor[gray]{0.9}3.1x & 2.4x & 1.6x & \cellcolor[gray]{0.9}4.7x & 2.6x & 2.1x \\
    \bottomrule
  \end{tabular}}
\end{table}

\section{Related Work}
\label{sec:related_work}

The literature on similarity search is vast~\cite{li_approximate_2020,shimomura_survey_2021}. Research on the topic evolves quickly, trying to keep up with ever-increasing requirements: more data with larger dimensionality, higher speeds, and a high recall. Trees~\cite{silpa-anan_optimised_2008,muja_scalable_2014,cayton_fast_2008} suffer from the curse of dimensionality. Hashing~\cite{indyk1998approximate,jafari_survey_2021} and learning-to-hash~\cite{wang_survey_2018} techniques often struggle to simultaneously achieve high accuracy and high speeds. Graph-based methods~\cite{arya1993approximate,malkov2018efficient,jayaram2019diskann,fu_fast_2019,iwasaki_optimization_2018,ChenW18} offer a better latency-accuracy trade-off than other types of algorithms~\cite{wang_graph_survey_2021}.

 Product Quantization (PQ)~\cite{jegou2010product} and other related methods~\cite{ge2013optimized,babenko_additive_2014,zhang2014composite,andre_cache_2015,matsui_paper_2018,guo2020accelerating,wang_deltapq_2020,johnson_billion-scale_2021,andre2021quicker_adc,ko2021low} were introduced to handle large datasets in settings with limited memory capacity~\cite[e.g.,][]{jayaram2019diskann}. However, when used for high-throughput graph-search, these quantizers do not enable extremely fast similarity computations in a predominantly random memory access pattern.

For inverted indices~\cite{johnson2019billion}, the setup for which PQ was designed, the similarity between partitions of the query and each corresponding centroid is generally precomputed to create a look-up table of partial similarities.
The computation of the similarity between vectors essentially becomes a set of indexed gather and accumulate operations on this table, which are generally quite slow~\cite{pase2019gather}. This is exacerbated with an increased dataset dimensionality: the lookup table does not fit in L1 cache, which slows down the gather operation.
In no small part, the success of recent PQ-based methods~\cite{guo2020accelerating} can be attributed to Quicker ADC~\cite{andre2021quicker_adc}, with its optimized lookup operations using AVX shuffle and blend instructions to compute the distance between a query and multiple dataset elements simultaneously. This is enabled by storing these elements in a transposed fashion. This transposition and Quicker ADC by extension are not compatible with the random memory access pattern
we see in graph-based similarity search.

Scalar quantization is used for low-precision inference in neural networks~\cite{Dai2021VSQuantPS} to compress the parameter tensors, quantizing each SIMD-sized vector within a tensor individually. They also propose compressing the quantization parameters themselves instead of the residuals as in LVQ. These techniques have not been used for similarity search.

There is existing art in two-level quantization. In~\cite{jayaram2019diskann} the index resides in an SSD instead of in main memory, which enables the use of full-precision vectors. However, SSD indices cannot achieve the performance of their in memory counterparts. In~\cite{xu2022residual}, PQ is used for both levels, inheriting the aformentioned issues with PQ. Additionally, both levels are jointly optimized using PQ, which is prohibitive at billion scale.

Dimensionality reduction~\cite{ailon2006approximate,zhang_connecting_2022} is an appealing alternative for vector compression that provides an orthogonal reduction to LVQ: they can be combined for stacked gains.

\section{Conclusions}
\label{sec:conclusion}

We presented new techniques for creating faster and smaller indices for similarity search. We introduced a novel vector compression method, Locally-adaptive Vector Quantization (LVQ), that simultaneously reduces memory footprint and improves search performance, with minimal impact on search accuracy. LVQ is designed to work optimally in conjunction with graph-based indices, reducing their effective bandwidth while enabling random-access-friendly fast similarity computations. LVQ, combined with key optimizations for graph-based indices in modern datacenter systems, establishes the new state of the art in terms of performance and memory footprint, outcompeting the second-best alternatives for billion scale datasets:
(1) in the low-memory regime, by up to 20.7x in throughput with up to a 3x memory footprint reduction, and
(2) in the high-throughput regime by 5.8x with 1.4x less memory.

For future work, we plan on studying the impact of LVQ on dynamic similarity search, dimensionality reduction~\cite{zhang_connecting_2022} as a preprocessing step, and intra-query parallelism~\cite{peng2022speed-ann}.

\begin{acks}
We would like to thank Kianoosh Zandifar for his invaluable help with our computer systems, and Jawad Khan and Sourabh Dongaonkar for their contributions to an earlier version of our work.
\end{acks}

\clearpage
\onecolumn
\begin{multicols}{2}
\bibliographystyle{ACM-Reference-Format}
\bibliography{refs}
\end{multicols}

\clearpage

\appendix
\onecolumn
\setcounter{page}{1}

\begin{center}
\Huge{Supplementary material\\Similarity search in the blink of an eye with compressed indices}

\vspace{2em}

\end{center}

\section{Locally-adaptive Vector Quantization}
\label{sec:vector_components_distro_supMat}

\begin{figure}[hb]
  \centering
  \includegraphics[width=0.3\linewidth]{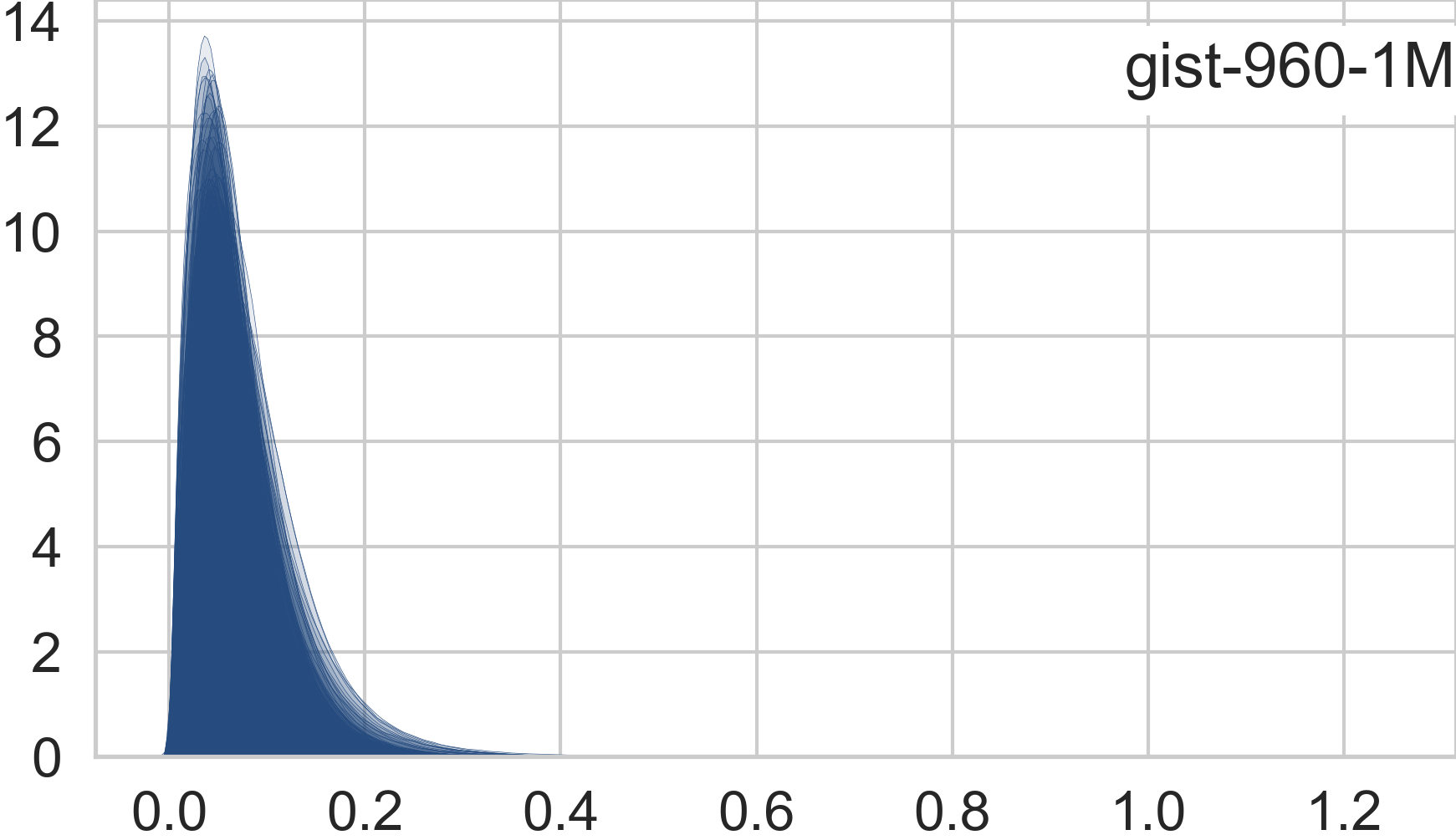}
  \includegraphics[width=0.3\linewidth]{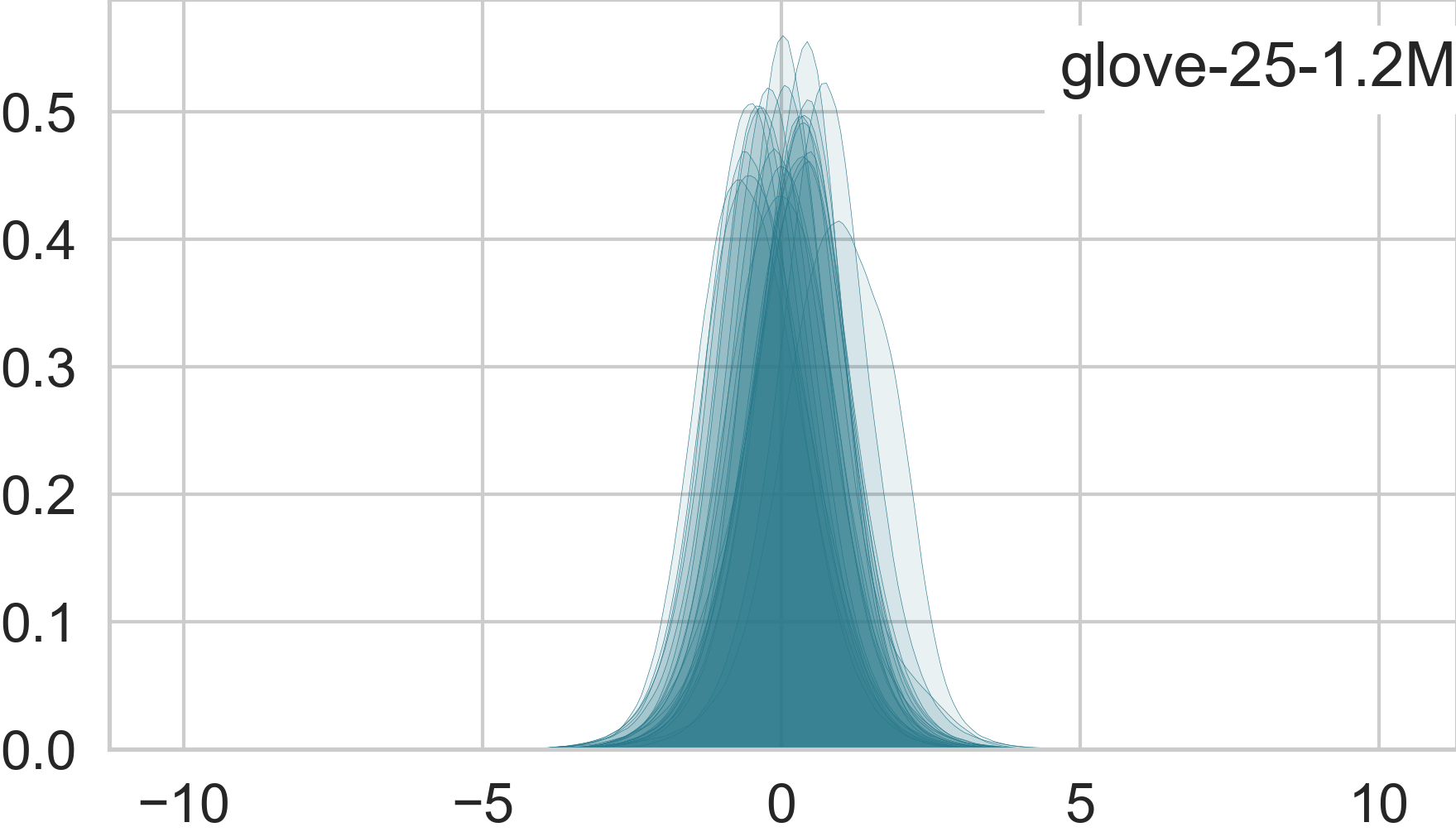}  
  \caption{(Continuation of \cref{fig:vector_components_distro}) Empirical distributions of vector values in individual dimensions for datasets gist-960-1M and glove-25-1.2M. These are almost identical to each other, mainly differing in their mean and with a slight variance change. After de-meaning, the resulting values become highly amenable to quantization, as the empirical distributions will not contain regions within the dynamic range with either very high or very small density.}
  \label{fig:vector_components_distro_supMat}
\end{figure}

\subsection{Experimental results with large variance differences}
\label{supp:large_variance_differences}
In standard datasets the different dimensions follow very similar distributions, once we remove their mean. This is why a de-meaning step is incorporated in the definition of LVQ. In pathological cases, these distributions may be significantly different potentially affecting the reconstruction accuracy of LVQ.

The dimensions with the largest variances are the ones determining the quantization step (as the upper and lower bounds are more likely to come from them) and therefore the reconstruction accuracy for the whole vector. How this will impact search accuracy, however, is not direct. Under such scenario, the dimensions with orders of magnitude larger variance would be much more relevant in the distance between the query and the vectors. This would essentially restrict the vectors comparison to the distance between the dimensions with large variance (which have smaller reconstruction error), rendering the dimensions with smaller variance (which have larger reconstruction error) meaningless in the comparison.

\begin{figure*}
  \centering  

  \begin{minipage}{0.05\textwidth}
  \rotatebox{90}{deep-96-10M-modified}
  \end{minipage}\hspace{-1.5em}
  \begin{minipage}{0.9\textwidth}
    \centering         
    \includegraphics[width=\textwidth]{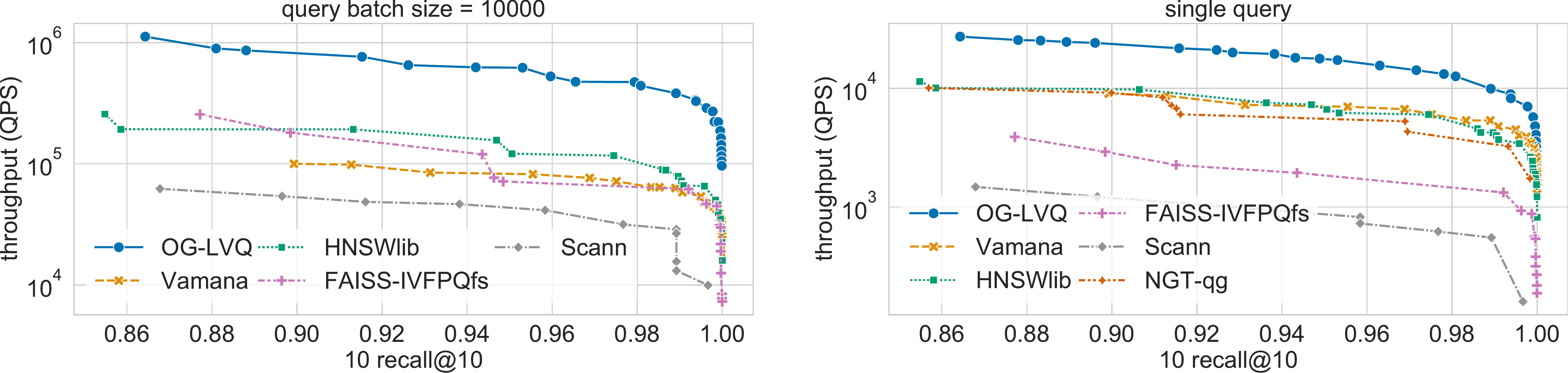}
  \end{minipage}  
  \vspace{10pt}

  \begin{minipage}{0.05\textwidth}
  \rotatebox{90}{gist-960-1M-modified}
  \end{minipage}\hspace{-1.5em}
  \begin{minipage}{0.9\textwidth}
    \centering         
    \includegraphics[width=\textwidth]{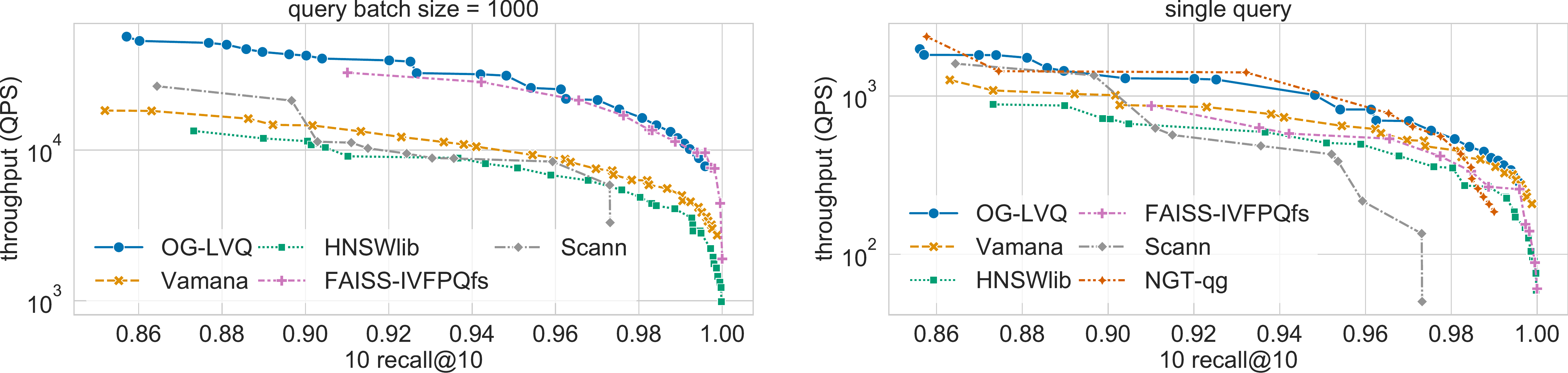}
  \end{minipage}     
  
  \begin{minipage}{0.05\textwidth}
  \rotatebox{90}{random-96-1M}
  \end{minipage}\hspace{-1.5em}
  \begin{minipage}{0.9\textwidth}
    \centering         
    \includegraphics[width=\textwidth]{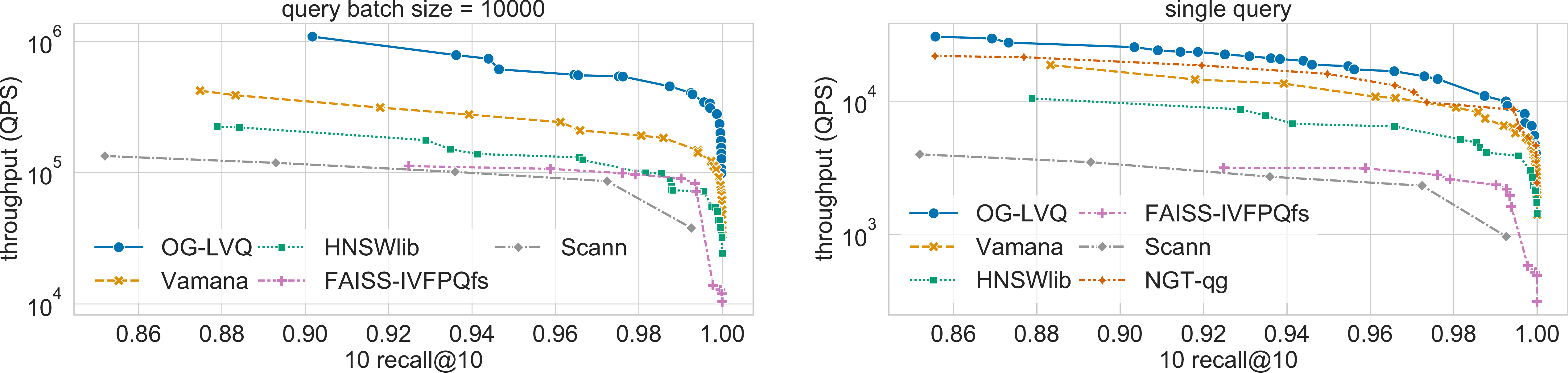}
  \end{minipage}  
  \vspace{10pt}  
  
  \caption{Benchmarking results for datasets with significantly different variances across dimensions. Each row corresponds to a different dataset and each column to a different query mode. \iss{} outperforms or is on par with other methods in all cases, showing the robustness of the LVQ compression even when dimensions have orders of magnitude different variances. }
  \label{fig:small_scale_benchmarking_diverseVariance}
\end{figure*}

As none of the standard datasets used to evaluate similarity search algorithms exhibit this pathological behavior, to evaluate the performance of \iss{} in this scenario we modify two of the standard datasets (deep-image-96-angular and gist-960-euclidean) by scaling the variances of 20\% of their dimensions by a factor between 10 and 100, and we also generate a new dataset with 1 million 96-dimensional vectors with Gaussian distributed values, where 20\% of the dimensions have standard deviations between 10 and 100 and the remaining 80\% have standard deviations between 0.1 and 1.0. In the case of the deep-image-96-angular dataset, we use the euclidean distance instead of the cosine similarity (as was used for the original dataset) because, after scaling, the vectors are no longer normalized to norm one. \cref{fig:small_scale_benchmarking_diverseVariance} shows that \iss{} is very competitive even in this pathological scenario.

We use the following code used to generate these modified datasets:\\

\begin{lstlisting}[language=python]
import numpy as np
import random

np.random.seed(0)
random.seed(0)

# Function to generate a new random dataset with variable variance per dimension
def generate_dataset_variable_variance(n, nq, d, low_scale=10, high_scale=100, perc_diff_var=0.1):
    print('Generating random dataset')
    scale_large = np.random.uniform(low=low_scale, high=high_scale, size=int(d * perc_diff_var))
    scale_small = np.random.uniform(low=0.1, high=1.0, size=d - len(scale_large))
    scale = np.concatenate([scale_small, scale_large])
    X = np.random.normal(scale=scale, size=[n, d])
    Q = np.random.normal(scale=scale, size=[nq, d])
    return X.astype('float32'), Q.astype('float32')

# Function to modify the variance of a randomly chosen subset of the dimensions of a given dataset (base vectors and query set)
def modify_dataset_variance(X, Q, perc_diff_var=0.1, low_factor=10, high_factor=100):
    print('Modifying dataset')
    d = X.shape[1]
    chosen_dims = random.sample(list(range(d)), int(d * perc_diff_var))
    scale_factor = np.random.uniform(low=low_factor, high=high_factor, size=int(d * perc_diff_var))
    X_modif = X.copy()
    X_modif[:, chosen_dims] = np.multiply(X_modif[:, chosen_dims], scale_factor)
    Q_modif = Q.copy()
    Q_modif[:, chosen_dims] = np.multiply(Q_modif[:, chosen_dims], scale_factor)
    return X_modif.astype('float32'), Q_modif.astype('float32'), chosen_dims, scale_factor


\end{lstlisting}

\section{Indexing with compressed vectors}
\label{proof:compute_mu_sigma}

\begin{figure}[hb]
    \centering
    \includegraphics[width=0.5\linewidth]{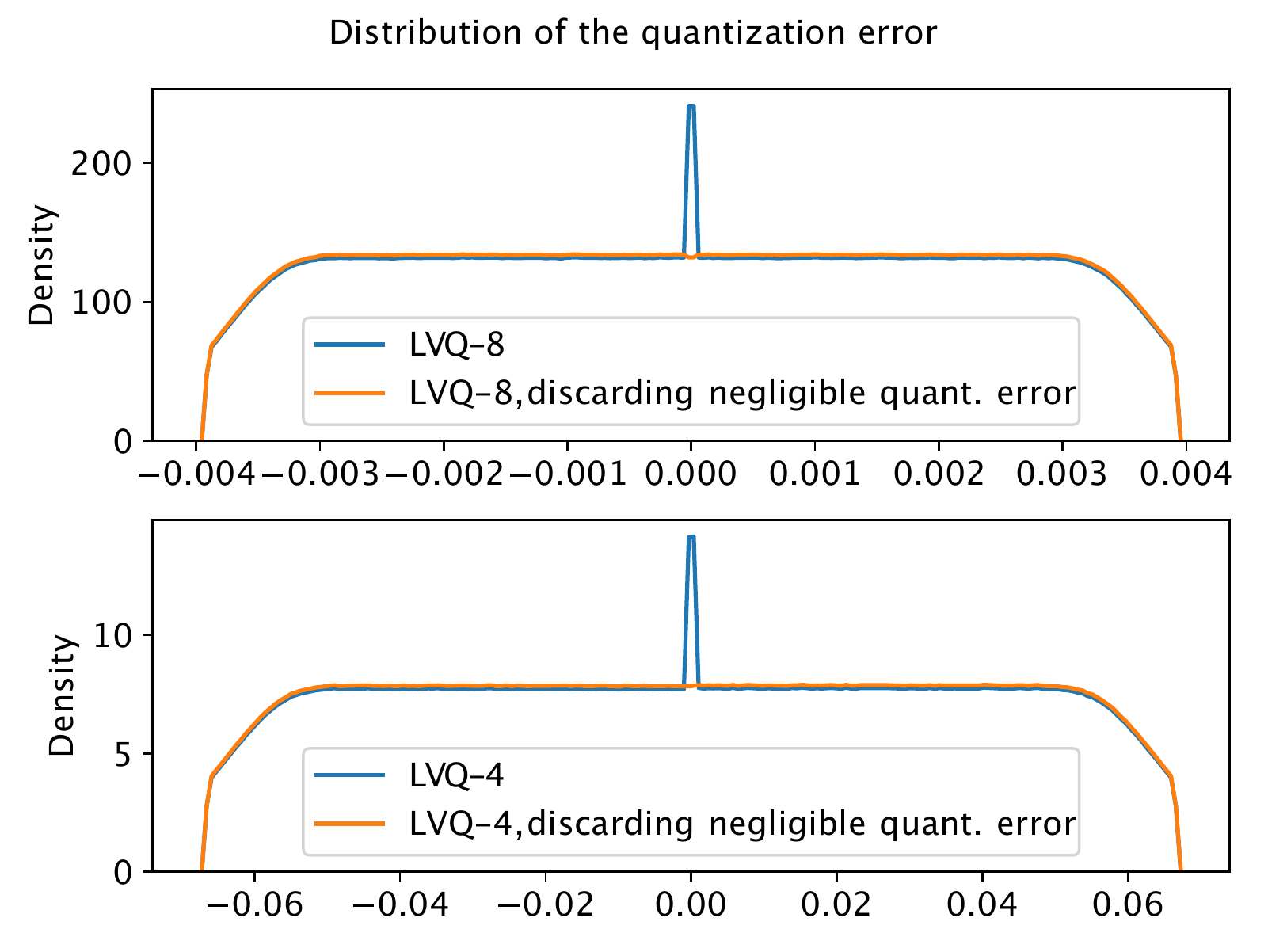}
    \caption{The uniform distribution approximates faithfully the empirical distribution of the LVQ quantization error for a subset of 1M vectors from deep-96-10M. The exception is the center peak, which corresponds to the values at the extremes of the vector (min.~and max.) that have zero quantization error. After removing those entries, that can be perfectly reconstructed anyway, the center peak disappears.}
    \label{fig:uniform_quantization_error}
\end{figure}

\proof[Proof of \cref{prop:pruning_rule_euclidean_equivalence}]{
We compute $\hat{\vect{a}}$ and $\hat{b}$ from $Q(\vect{x})$ and $Q(\vect{x}^*)$ using \cref{eq:sign_ax_b}.
Let $\vect{z}_v = \vect{v} - Q(\vect{v}), \forall \vect{v} \in \Real^d$.
We have
\begin{align}
    \sign(\transpose{\hat{\vect{a}}} Q(\vect{x}') - \hat{b})
    &= \sign \left( \frac{\transpose{(Q(\xv)-Q(\xv^*))}}{ \| Q(\xv)-Q(\xv^*) \|} Q(\xv') - \frac{\|Q(\xv)\|^2 - \|Q(\xv^*\|^2}{2 \| Q(\xv)-Q(\xv^*) \|} \right) \\
    &= \sign \left( (\xv - \xv^*)^\top\xv' - \frac{1}{2} (\xv^\top\xv - \xv^{*T}\xv^*) - E \right) \\
    &= \sign((\transpose{\vect{a}} \vect{x}' - b)\|\vect{x} - \vect{x}^*\| - \error),
\end{align}
where
\begin{equation}
    E = (\vect{z}_x - \vect{z}_{x^*})^\top\xv' + (\xv - \xv^*)^\top \vect{z}_{x'} - \vect{z}_x^\top \vect{z}_{x'} + \vect{z}_{x^*}^\top \vect{z}_{x'} + \frac{1}{2} ( \vect{z}_x^\top \vect{z}_x -2\xv^\top \vect{z}_x - \vect{z}_{x^*}^\top \vect{z}_{x^*} + 2\xv^{*T} \vect{z}_{x^*}) .
    \label{eq:error_term_realization}
\end{equation}

When \cref{eq:pruning_rule_euclidean_equivalence} holds and from the definition of $\vect{a}$ and $\vect{b}$ in \cref{eq:sign_ax_b}, we finally have
\begin{align}
    \sign(\transpose{\hat{\vect{a}}} Q(\vect{x}') - \hat{b})
    &= \sign((\transpose{\vect{a}} \vect{x}' - b)\|\vect{x} - \vect{x}^*\| - \error) \label{eq:sign_approximation} \\
    &= \sign((\transpose{\vect{a}} \vect{x}' - b)\|\vect{x} - \vect{x}^*\|) \\
    &= \sign(\transpose{\vect{a}} \vect{x}' - b) .
    \label{eq:sign_approximation2}    
\end{align}
\qed
}

\proof[Proof of \cref{prop:pruning_rule_euclidean_equivalence_error}]{
Let us consider random variable $\Z_x=[Z_{x_1},\dots,Z_{x_d}]$ where the \textit{i.i.d} scalar random variables $Z_{x_i} \sim U \left[ -\tfrac{\Delta_x}{2},\tfrac{\Delta_x}{2} \right]$ $i=1,\dots,d$ model the quantization error $Z_{x_i}=X_i-Q(X_i)$ for component $i$ of vector $\vect{x}$. Similarly we consider $\Z_{x'}$ and 
$\Z_{x^*}$, with $Z_{x'_i} \sim U \left[ -\tfrac{\Delta_{x'}}{2},\tfrac{\Delta_{x'}}{2} \right]$ and $Z_{x^*_i} \sim U \left[ -\tfrac{\Delta_{x^*}}{2},\tfrac{\Delta_{x^*}}{2} \right]$, for the quantization errors of vectors $\xv'$ and $\xv^*$, respectively. Vectors $\xv$, $\xv'$ and $\xv^*$ are assumed given and deterministic. 

With a slight abuse of notation, we rewrite $E$ in \cref{eq:error_term_realization} as a random variable, obtaining
\begin{equation}
    E = (\Z_x - \Z_{x^*})^\top\xv' + (\xv - \xv^*)^\top\Z_{x'} - \Z_x^\top\Z_{x'} + \Z_{x^*}^\top\Z_{x'} + \frac{1}{2} ( \Z_x^\top\Z_x -2\xv^\top \Z_x - \Z_{x^*}^\top\Z_{x^*} + 2\xv^{*T} \Z_{x^*}) .
    \label{eq:error_term}
\end{equation}
Each term in $E$ is the sample mean of $d$ \textit{i.i.d} random variables (up to a multiplicative factor $1/d$). Therefore, by the central limit theorem, it converges to a Gaussian distributed random variable. Hence, $E$ is a linear combination of Gaussian random variables and it is therefore also Gaussian distributed. 
Let us now compute its expected value $\mu_E$ and variance $\sigma_E$. By re-organizing the terms in \cref{eq:error_term} we have,

\begin{equation}
    \mu_E = \mathbb{E}(\underbrace{\Z_x^\top(\xv' - \xv)}_{\text{t}_1}) - \mathbb{E}( \underbrace{\Z_{x^*}^\top(\xv' - \xv^*)}_{\text{t}_2}) + \frac{1}{2}\mathbb{E}(\underbrace{\Z_x^\top\Z_x}_{\text{t}_3}) - \frac{1}{2}\mathbb{E}(\underbrace{\Z_{x^*}^\top\Z_{x^*}}_{\text{t}_4}) + \mathbb{E}( \underbrace{(\xv - \xv^*)^\top\Z_{x'}}_{\text{t}_5}) - \mathbb{E}( \underbrace{\Z_x^\top \Z_{x'}}_{\text{t}_6}) + \mathbb{E}(\underbrace{\Z_{x^*}^\top\Z_{x'}}_{\text{t}_7}).
\label{eq:mu_E_compute}
\end{equation}

Because $\mathbb{E}(Z_{x_i})=\mathbb{E}(Z_{x'_i})=\mathbb{E}(Z_{x^*_i})=0$  $\forall i=1,\dots,d$ and $\Z_x$, $\Z_{x^*}$ and $\Z_{x'}$ are mutually independent we have, 
\begin{align}
    \mu_E &= \frac{1}{2} \mathbb{E}( \Z_x^\top\Z_x ) - \frac{1}{2} \mathbb{E}( \Z_{x^*}^\top\Z_{x^*}) \\
          &= \sum_{i=1}^d Z_x^2 + \sum_{i=0}^d Z_{x^*}^2 \\
          &= \frac{d}{24} (\Delta_x^2 - \Delta_{x^*}^2).
\end{align}

To simplify the notation in the computation of $\sigma_E^2$, we refer to the terms $\text{t}_{j=1,\dots,7}$ indicated in \cref{eq:mu_E_compute}. From \cref{eq:error_term}, $\sigma_E^2$ is given by,
\begin{align}
    \sigma_E^2 = &\sum_{j=1}^7 \var{\text{t}_j} + \cov(\text{t}_1,\text{t}_3) + \cov(\text{t}_2,\text{t}_4) + 2\cov(\text{t}_2,\text{t}_7) + \cov(\text{t}_3,\text{t}_6) +  \cov(\text{t}_4,\text{t}_7) + 2\cov(\text{t}_5,\text{t}_6) + 2\cov(\text{t}_5,\text{t}_7) + 2\cov(\text{t}_6,\text{t}_7).
\end{align}

Because $\mathbb{E}(Z_{x_i})=0$, $Z_{x_i}$ and $Z_{x_j}$ are independent for $i \neq j$, and $Z_{x_i} \sim U \left[ -\tfrac{\Delta_x}{2},\tfrac{\Delta_x}{2} \right]$, $i=1,\dots,d$ we have,
\begin{align}
    \cov(\text{t}_1,\text{t}_3)&= \mathbb{E}(\text{t}_1\text{t}_3) - \mathbb{E}(\text{t}_1)\mathbb{E}(\text{t}_3) \\
                               &= \mathbb{E}(\Z_x^\top(\xv' - \xv)\Z_x^\top\Z_x) \\
                               &= \sum_{i=1}^d (x'_i - x_i) \mathbb{E}(Z_{x_i}^3) \\
                               &= \sum_{i=1}^d (x'_i - x_i) \frac{1}{4\Delta_x} \left [ \left ( \frac{\Delta_x}{2} \right )^4 - \left ( \frac{\Delta_x}{2} \right )^4 \right ]\\
                               &= 0.
\end{align}
In the same way, we find that $\cov(\text{t}_2,\text{t}_4)=0$.

Because $\mathbb{E}(Z_{x_i})=0$, $Z_{x_i}$ and $Z_{x_j}$ are independent for $i \neq j$, and $Z_{x_i}$ and $Z_{x'_i}$ are independent we have,
\begin{align}
    \cov(\text{t}_1,\text{t}_6)&= \mathbb{E}(\text{t}_1\text{t}_6) - \mathbb{E}(\text{t}_1)\mathbb{E}(\text{t}_6) \\
                               &= \mathbb{E}(\Z_x^\top(\xv' - \xv)\Z_x^\top \Z_{x'}) \\
                               &= \sum_{i=1}^d (x'_i - x_i) \mathbb{E}(Z_{x_i}^2)\mathbb{E}(Z_{x'_i}) \\
                               &= 0 .
\end{align}
In the same way, we find that $\cov(\text{t}_2,\text{t}_7)=0$.

Because $\mathbb{E}(Z_{x_i})=0$ and $Z_{x_i}$ and $Z_{x'_i}$ are independent we have,
\begin{align}
    \cov(\text{t}_3,\text{t}_6)&= \mathbb{E}(\text{t}_3\text{t}_6) - \mathbb{E}(\text{t}_3)\mathbb{E}(\text{t}_6) \\
                               &= \mathbb{E}(\Z_x^\top\Z_x^\top\Z_x^\top\Z_{x'}) \\
                               &= \sum_{i=1}^d \mathbb{E}(Z_{x'_i}Z_{x_i}^3) \\                    
                               &= 0.
\end{align}
In the same way, we find that $\cov(\text{t}_4,\text{t}_7)=0$.

Because $\mathbb{E}(Z_{x_i})=0$, $Z_{x_i}$ and $Z_{x'_i}$ are independent we have,
\begin{align}
    \cov(\text{t}_5,\text{t}_6)&= \mathbb{E}(\text{t}_5\text{t}_6) - \mathbb{E}(\text{t}_5)\mathbb{E}(\text{t}_6) \\
                               &= \mathbb{E}((\xv-\xv^*)^\top\Z_{x'}\Z_{x}^\top\Z_{x'}) \\
                               &= \sum_{i=1}^d \mathbb{E}((x_i-x^*_i)Z_{x'_i}^2 Z_{x_i}) \\
                               &= 0.
\end{align}
In the same way, we find that $\cov(\text{t}_5,\text{t}_7)=0$.

Because $\mathbb{E}(Z_{x_i})=0$, and $Z_{x_i}$, $Z_{x^*_i}$ and $Z_{x'_i}$ are independent we have,
\begin{align}
    \cov(\text{t}_6,\text{t}_7)&= \mathbb{E}(\text{t}_6\text{t}_7) - \mathbb{E}(\text{t}_6)\mathbb{E}(\text{t}_7) \\                               
                               &= \sum_{i=1}^d \mathbb{E}(Z_{x_i}Z_{x^*_i}Z_{x'_i}^2) \\                
                               &= 0.
\end{align}

Moreover,
\begin{equation}
    \var{\text{t}_1} = \|\xv' - \xv \|^2 \frac{\Delta_{x}^2}{12}, \quad \var{\text{t}_2} = \| \xv' -\xv^* \|^2 \frac{\Delta_{x^*}^2}{12}\quad \text{and} \quad \var{\text{t}_5} = \|\xv - \xv^*\|^2 \frac{\Delta_{x'}^2}{12}.
\end{equation}

Because $Z_{x_i}$ and $Z_{x_j}$ are independent for $i \neq j$, we have 
\begin{align}
    \var{\text{t}_3} &= \sum_{i=1}^d \var{Z_{x_i}^2} \\
                     &= \sum_{i=1}^d \mathbb{E}(Z_{x_i}^4) - \mathbb{E}(Z_{x_i}^2)^2 \\
                     &= \frac{d\Delta_{x}^4}{720}.
\end{align}
Likewise, we find that $\var{\text{t}_4}=\tfrac{d\Delta_{x^*}^4}{720}$. 

Because $Z_{x_i}$ is independent of $Z_{x'_i}$, and $Z_{x_i}Z_{x'_i}$ is independent of $Z_{x_j}Z_{x'_j}$ for $j \neq i$, we have
\begin{align}
    \var{\text{t}_6} &= \sum_{i=1}^d \var{Z_{x_i}Z_{x'_i}} \\
                     &= \sum_{i=1}^d \mathbb{E}(Z_{x_i}^2)\mathbb{E}(Z_{x'_i}^2) - \mathbb{E}(Z_{x_i})^2\mathbb{E}(Z_{x'_i})^2\\
                     &= \frac{d}{144}\Delta_{x}^2\Delta_{x'}^2.
\end{align}
Likewise, we find that $\var{\text{t}_7}=\tfrac{d}{144}\Delta_{x^*}^2\Delta_{x'}^2$. Finally,
\begin{equation}
    \sigma_E^2 = \frac{\Delta_x^2}{12} \| \xv' -\xv \|^2 + \frac{\Delta_{x^*}^2}{12} \| \xv' -\xv^* \|^2 + \frac{\Delta_{x'}^2}{12} \| \xv -\xv^* \|^2 + \frac{d}{720}(\Delta_x^4+\Delta_{x^*}^4) + \frac{d}{144}\Delta_{x'}^2(\Delta_x^2+\Delta_{x^*}^2).
\end{equation}
\qed
}

\section{DPR Dataset Generation Details}
\label{appendix:DPR_code}
We developed the code to generate the DPR-768-10M dataset and have open-sourced it at \url{https://github.com/IntelLabs/DPR-dataset-generator.git}. The code allows to specify the number of vectors in the dataset.
To generate the 10 million base vector embeddings we used the version of the C4 dataset prepared by AllenAI and hosted at \emph{https://huggingface.co/datasets/allenai/c4}. We used the data split \emph{en/training} to generate the 10 million base vectors (files \emph{c4-train.00000-of-01024.json.gz} and \emph{c4-train.00001-of-01024}) and \emph{en/validation} to generate the 10000 query set (file \emph{c4-validation.00000-of-00008.json.gz}). We set the parameters as follows: \texttt{doc\_stride}=32, \texttt{max\_length}=64. The ground-truth for the dataset is generated using inner product, as it is the metric suggested in~\cite{karpukhin2020dense}.

\section{Additional experimental results}
\label{appendix:additional_experimental_results}
\begin{figure*}
  \centering  

  \begin{minipage}{0.05\textwidth}
  \rotatebox{90}{sift-128-1M}
  \end{minipage}\hspace{-1.5em}
  \begin{minipage}{0.9\textwidth}
    \centering         
    \includegraphics[width=\textwidth]{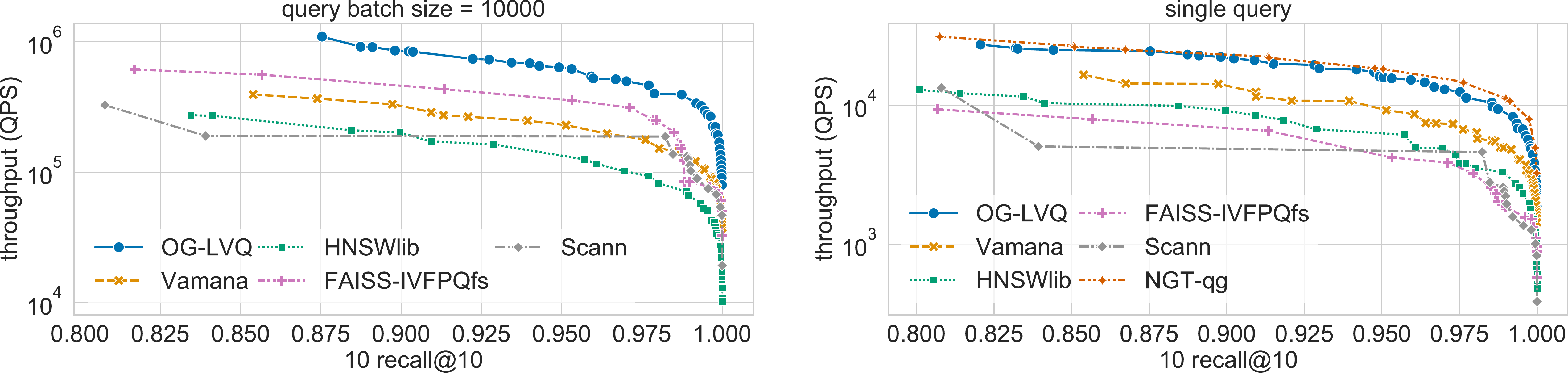}
  \end{minipage}  
  \vspace{10pt}

  \begin{minipage}{0.05\textwidth}
  \rotatebox{90}{gist-960-1M}
  \end{minipage}\hspace{-1.5em}
  \begin{minipage}{0.9\textwidth}
    \centering         
    \includegraphics[width=\textwidth]{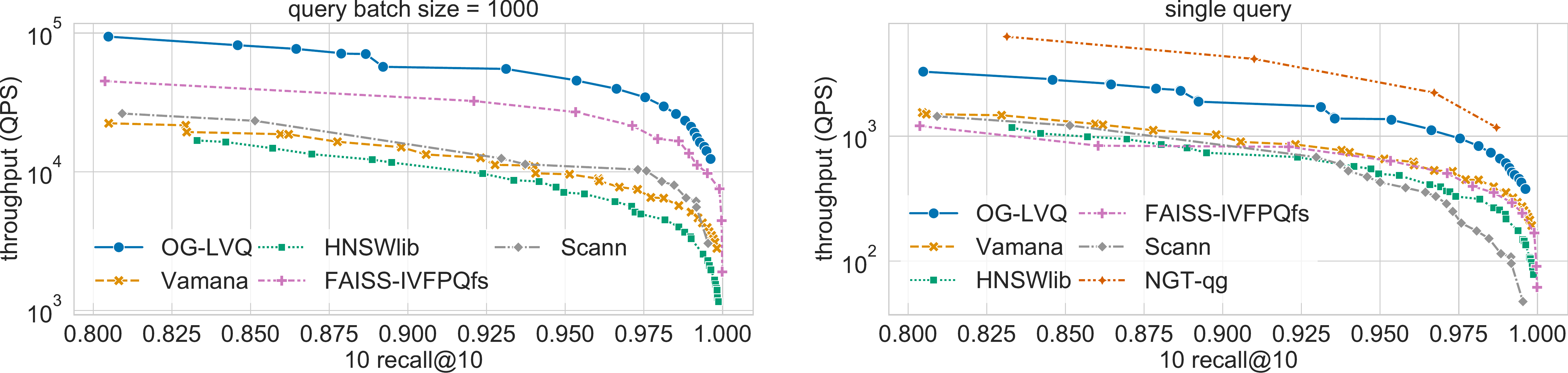}
  \end{minipage}     
  
  \begin{minipage}{0.05\textwidth}
  \rotatebox{90}{glove-25-1.2M}
  \end{minipage}\hspace{-1.5em}
  \begin{minipage}{0.9\textwidth}
    \centering         
    \includegraphics[width=\textwidth]{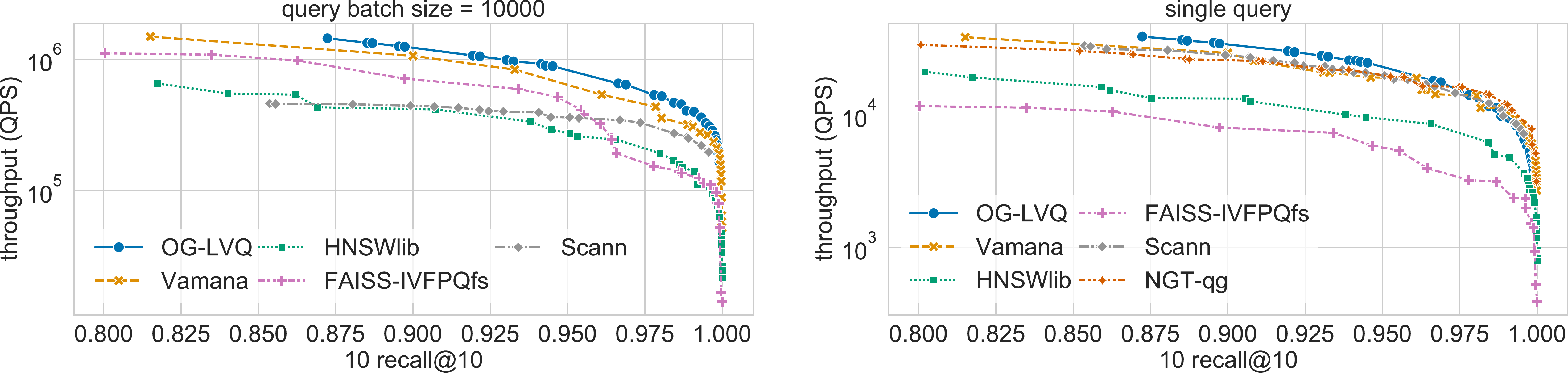}
  \end{minipage}  
  \vspace{10pt}  
  
  \begin{minipage}{0.05\textwidth}
  \rotatebox{90}{glove-50-1.2M}
  \end{minipage}\hspace{-1.5em}
  \begin{minipage}{0.9\textwidth}
    \centering         
    \includegraphics[width=\textwidth]{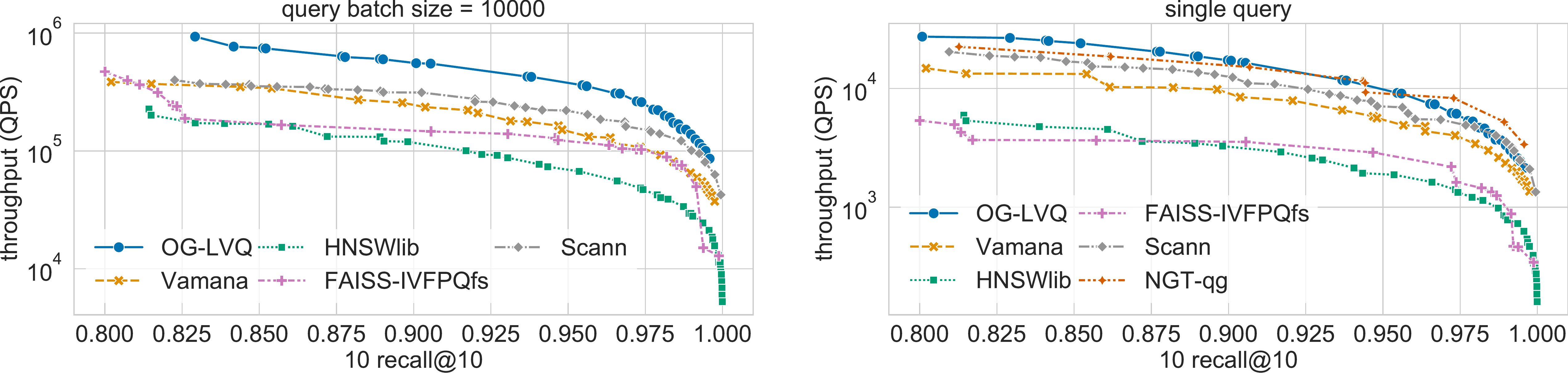}
  \end{minipage}        
  \vspace{10pt}

  \begin{minipage}{0.05\textwidth}
  \rotatebox{90}{deep-96-10M}
  \end{minipage}\hspace{-1.5em}
  \begin{minipage}{0.9\textwidth}
    \centering         
    \includegraphics[width=\textwidth]{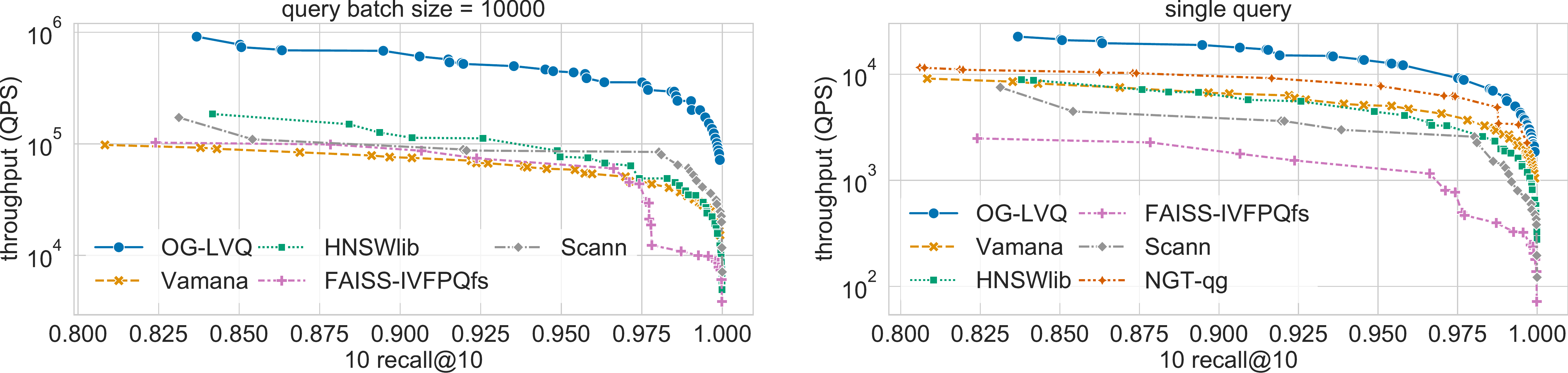}
  \end{minipage}  
  \caption{Benchmarking results for small scale datasets. Each row corresponds to a different dataset and each column to a different query mode. \iss{} establishes a new SOTA for all datasets, revealing its versatility and suitability for very diverse use cases. }
  \label{fig:small_scale_benchmarking_suppMat}
\end{figure*}

\begin{figure*}
  \centering  

  \begin{minipage}{0.05\textwidth}
  \rotatebox{90}{gist-960-1M}
  \end{minipage}\hspace{-1.5em}
  \begin{minipage}{0.9\textwidth}
    \centering         
    \includegraphics[width=\textwidth]{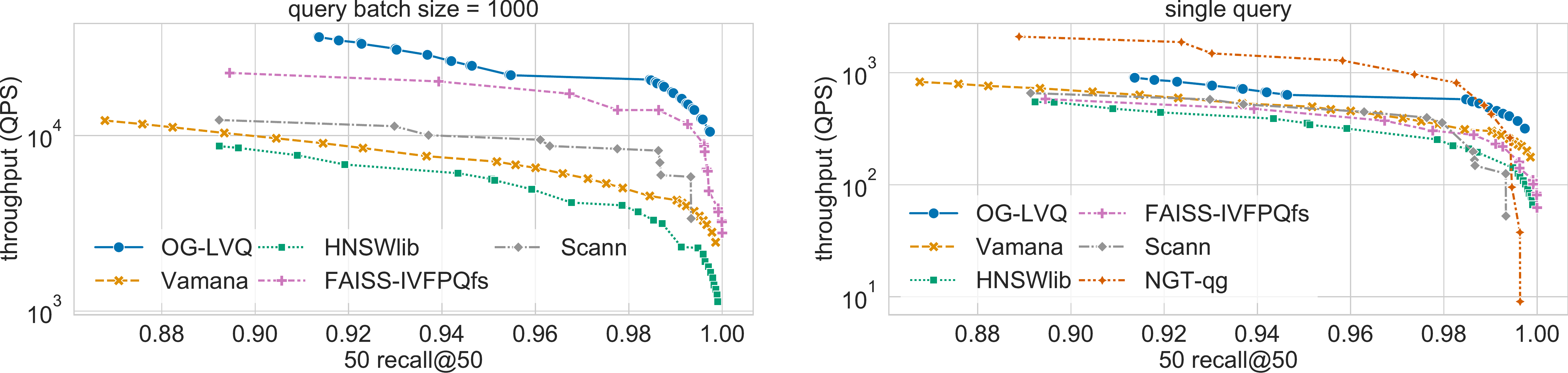}
  \end{minipage}     
  
  \begin{minipage}{0.05\textwidth}
  \rotatebox{90}{deep-96-10M}
  \end{minipage}\hspace{-1.5em}
  \begin{minipage}{0.9\textwidth}
    \centering         
    \includegraphics[width=\textwidth]{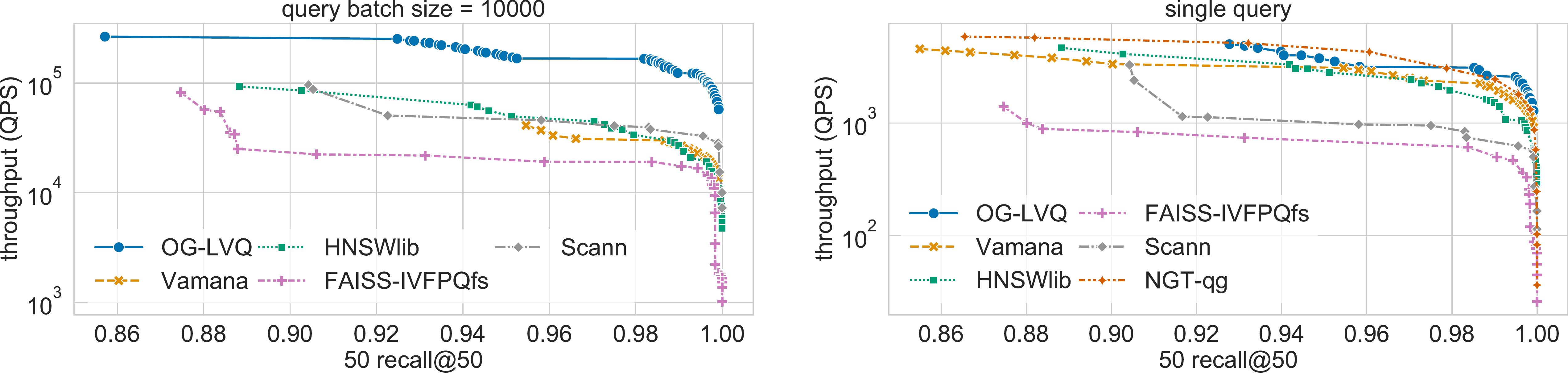}
  \end{minipage}  
  \caption{Benchmarking results for small scale datasets for a search accuracy of 50 recall@50. Each row corresponds to a different dataset and each column to a different query mode. Results are consistent with the ones reported for 10 recall@10. }
  \label{fig:small_scale_benchmarking_50recall_suppMat}
\end{figure*}

\begin{figure*}
  \centering  

  \begin{minipage}{0.05\textwidth}
  \rotatebox{90}{gist-960-1M}
  \end{minipage}\hspace{-1.5em}
  \begin{minipage}{0.9\textwidth}
    \centering         
    \includegraphics[width=\textwidth]{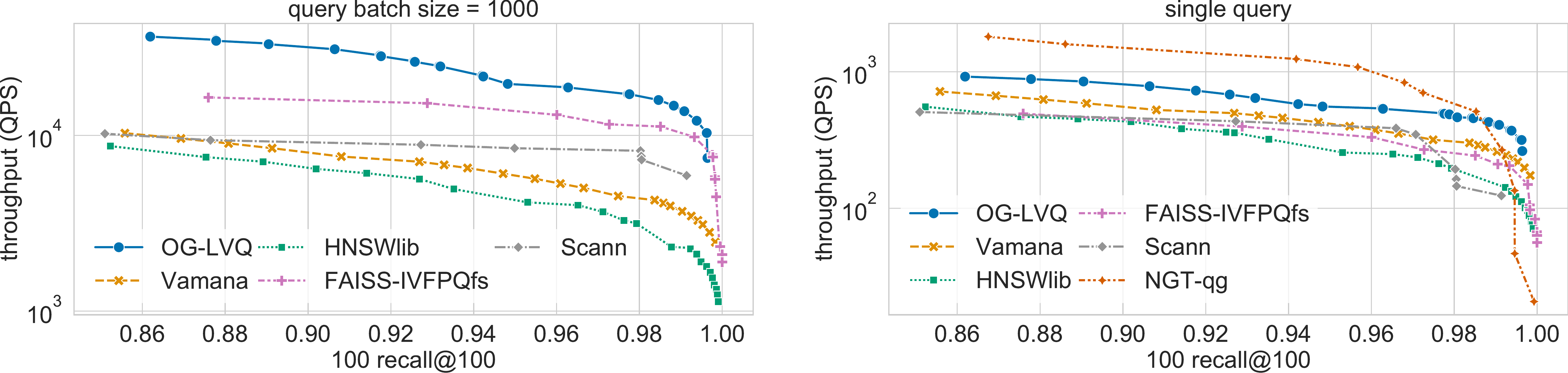}
  \end{minipage}     
  
  \begin{minipage}{0.05\textwidth}
  \rotatebox{90}{deep-96-10M}
  \end{minipage}\hspace{-1.5em}
  \begin{minipage}{0.9\textwidth}
    \centering         
    \includegraphics[width=\textwidth]{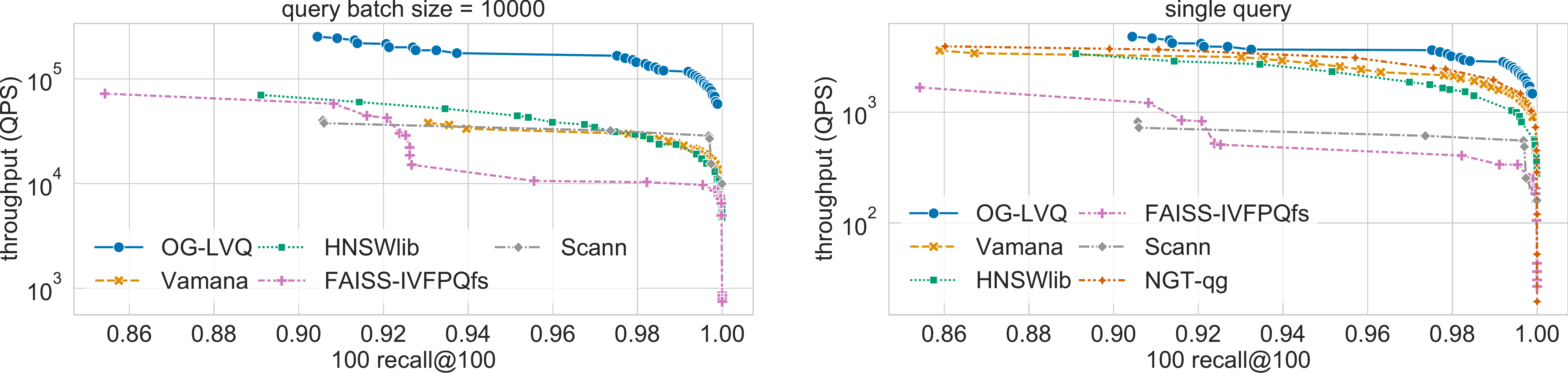}
  \end{minipage}  
  \caption{Benchmarking results for small scale datasets for a search accuracy of 100 recall@100. Each row corresponds to a different dataset and each column to a different query mode. Results are consistent with the ones reported for 10 recall@10.}
  \label{fig:small_scale_benchmarking_100recall_suppMat}
\end{figure*}

\begin{figure*}
  \centering  

  \begin{minipage}{0.9\textwidth}
    \centering         
    \includegraphics[width=\textwidth]{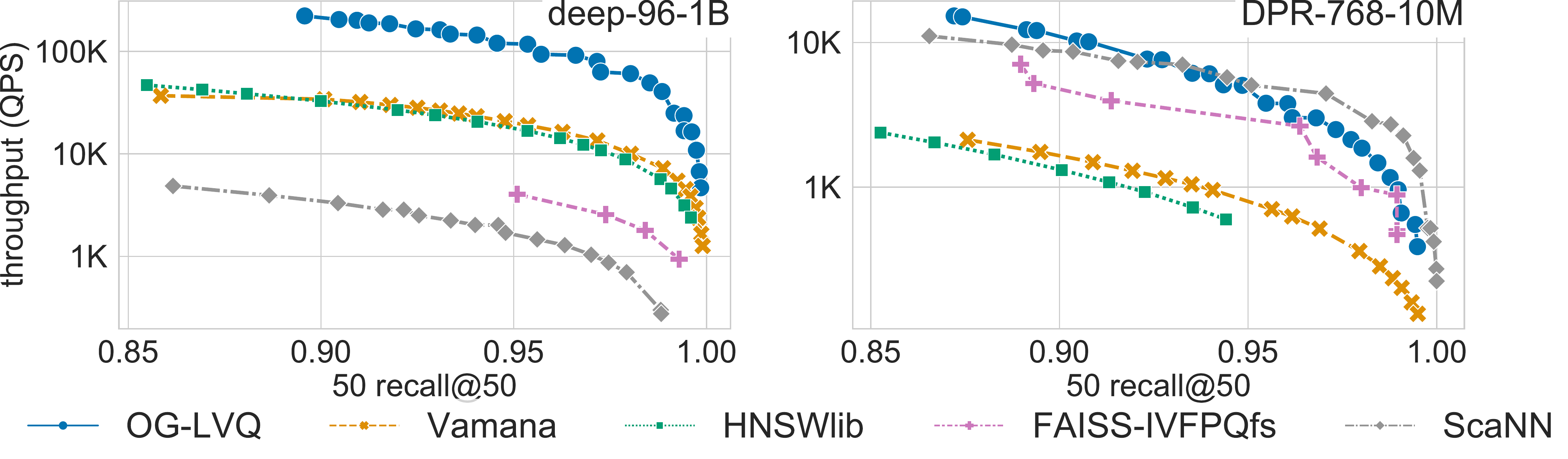}
  \end{minipage}     
  
  \begin{minipage}{0.9\textwidth}
    \centering         
    \includegraphics[width=\textwidth]{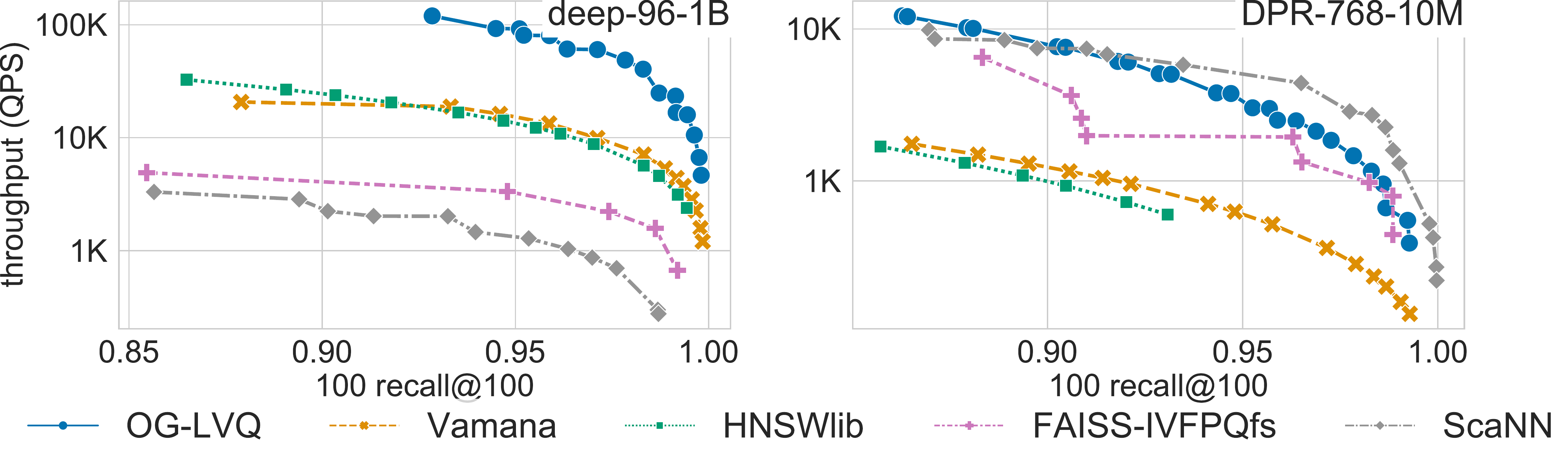}
  \end{minipage}  
  \caption{Benchmarking results for large scale datasets for a search accuracy of 50 recall@50 and 100 recall@100. Each row corresponds to a different recall regime and each column to a different dataset. Results are consistent with the ones reported for 10 recall@10. }
  \label{fig:large_scale_benchmarking_50_100recall_suppMat}
\end{figure*}

\cref{fig:small_scale_benchmarking_suppMat} shows the complete set of results for small scale datasets for 10 recall@10. 
Moreover, in order to evaluate other accuracy regimes, we report the results for 50 recall@50 and 100 recall@100 for two of the small scale datasets (deep-96010M and gist-960-1M representative of medium and high dimensionality datasets) in Figures \ref{fig:small_scale_benchmarking_50recall_suppMat} and \ref{fig:small_scale_benchmarking_100recall_suppMat} respectively, as well as for two of the large scale datasets (deep-96-1B and DPR-768-10M representative of medium and high dimensionality datasets) in \cref{fig:large_scale_benchmarking_50_100recall_suppMat}. 

Results for 50 recall@50 and 100 recall@100 are consistent with what we observe for 10 recall@10. The main difference is for the DPR-768-10M dataset, for which \iss{} is on par with ScaNN up to recall 0.90, but the latter takes the lead for increased accuracy regimes.

\subsection{Fast search and a small footprint with LVQ}
\begin{figure}
  \centering
    \begin{minipage}[b]{0.49\textwidth}
        \subfloat[Results for deep-96-100M with a 10-recall$@$10 of 0.9]{
            \includegraphics[width=\linewidth]{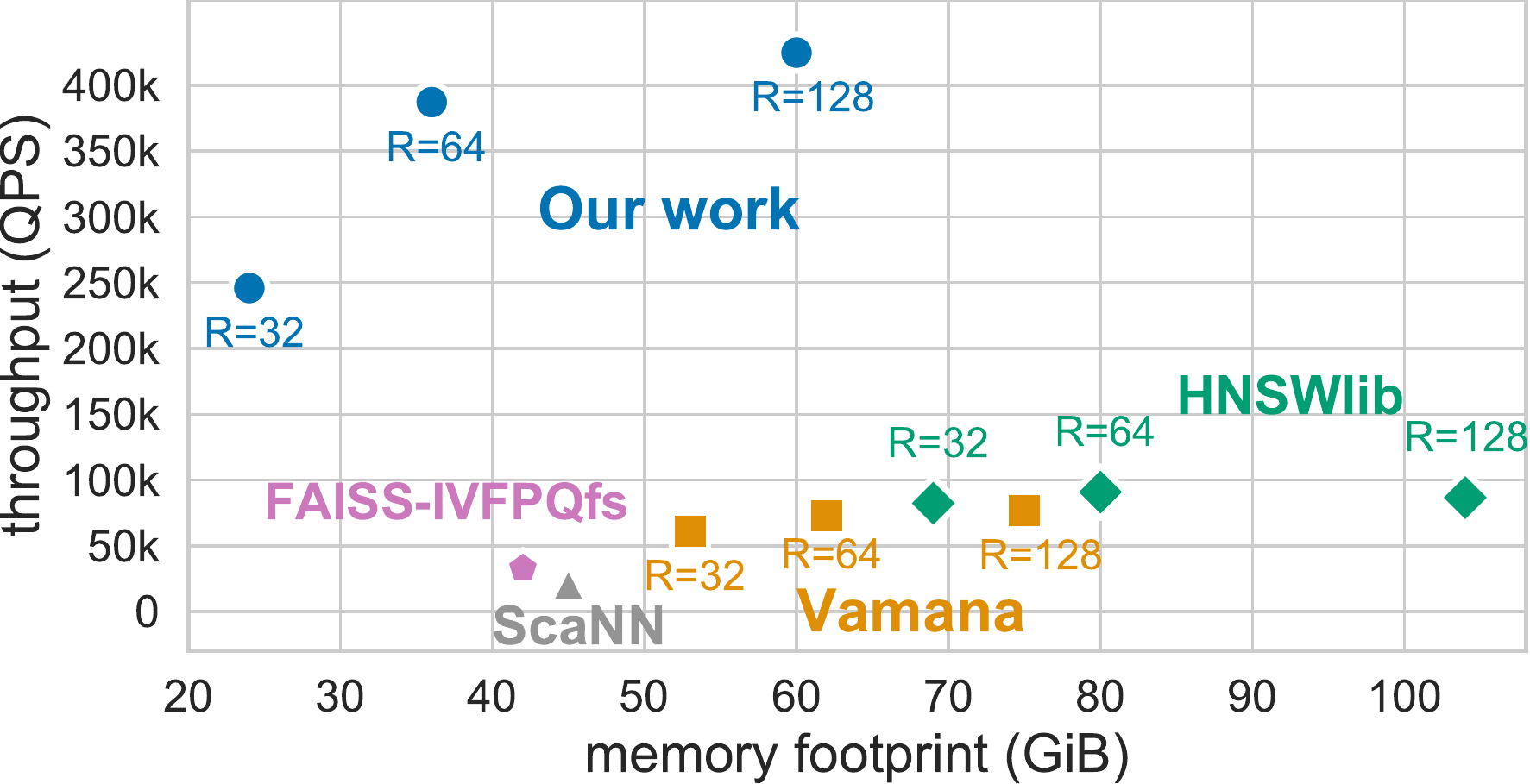}
            \label{fig:search_with_reduced_memfootprint}
        }
     \end{minipage}%
    \begin{minipage}[b]{0.47\textwidth}
        \subfloat[Results for deep-96-100M for all recall values]{
            \includegraphics[width=\linewidth]{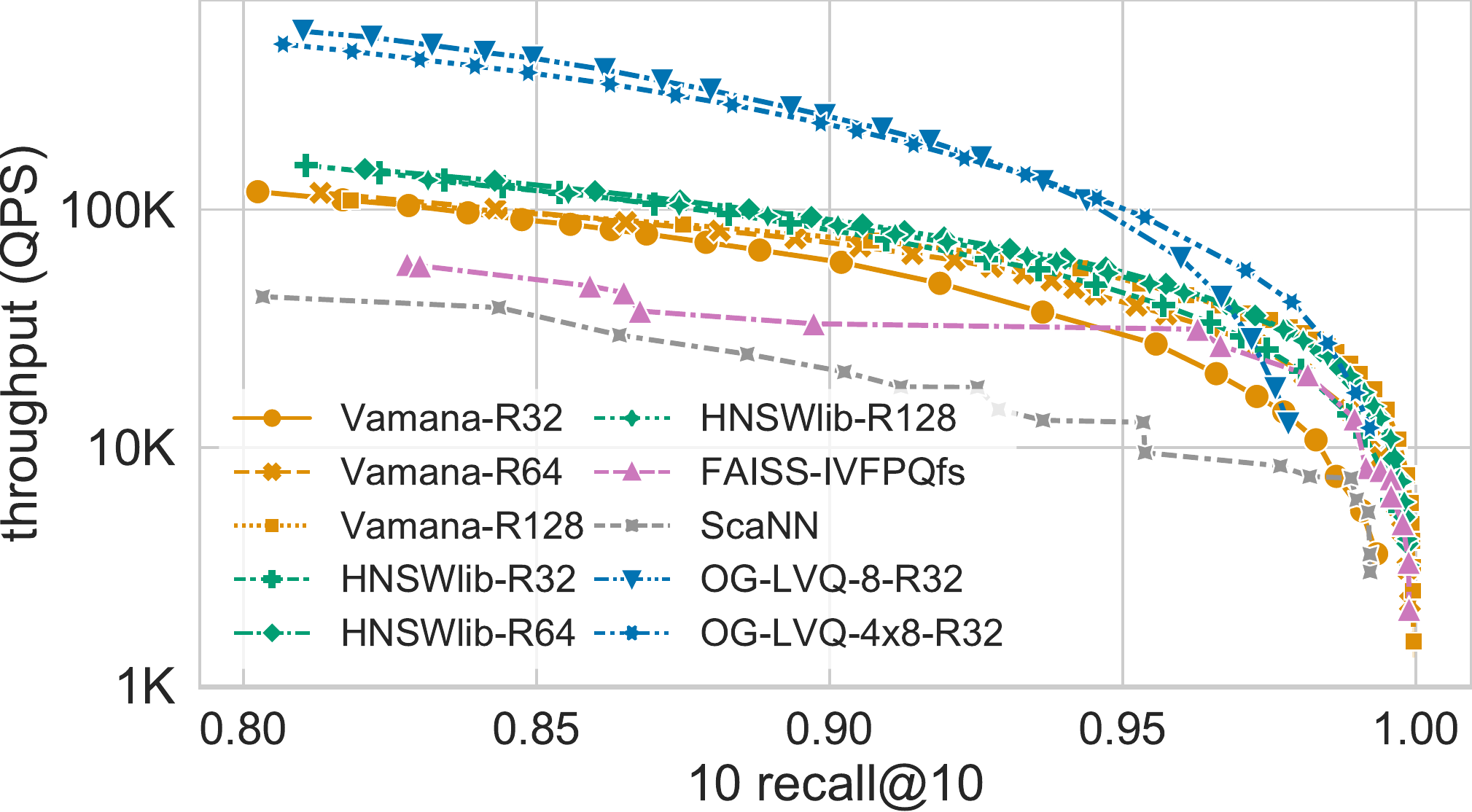}
            \label{fig:search_with_reduced_memfootprint_recall}
        }
     \end{minipage}%
   \caption{
   \protect\subref{fig:search_with_reduced_memfootprint} LVQ, combined with our system optimizations, enables high-throughput and high-accuracy search with a small memory footprint (results with a 10-recall$@$10 of 0.9 for deep-96-100M). For graph-based methods, the memory footprint is a function of the graph out-degree $\maxOutDeg{}$ and the vector footprint.
   Our low-memory configuration (LVQ-8 and $\maxOutDeg{}=32$) outperforms Vamana, HNSWlib, FAISS-IVFPQfs, and ScaNN, by 3.2x, 2.7x, 7.4x, and 11.5 with 3.1x, 3.3x, 1.8x, and use 1.9x less memory, respectively.
   Our highest-throughput configuration (LVQ-8 and $\maxOutDeg{}=128$) outperforms the second-highest by 4.7x with 1.3x less memory.
   \protect\subref{fig:search_with_reduced_memfootprint_recall} The proposed \iss{}, for LVQ-8 and maximum out-degree $\maxOutDeg{}=32$ , exhibits a higher throughput than its competitors across the entire recall range up to 0.97 for deep-96-100M. For higher recall values, \iss{} with LVQ-4x8 is on par with the competition with a 2.5x smaller footprint than the best alternative.
   }
\end{figure}

Figures \ref{fig:search_with_reduced_memfootprint} and \ref{fig:search_with_reduced_memfootprint_recall} show the search
throughput as a function of the memory footprint (measured as
the maximum resident main memory usage while doing the query
search) of different algorithms for deep-96-100M.

\end{document}